\documentclass[runningheads]{llncs}
\usepackage[T1]{fontenc}
\usepackage{graphicx}
\usepackage{hyperref}
\hypersetup{
    colorlinks=true,
    linkcolor=blue,
    citecolor=green,
    filecolor=black,
    urlcolor=black,
}

\usepackage{amsfonts}
\usepackage{bbm}
\usepackage{pgfplots} 
\usepackage{adjustbox}
\usepackage{comment}
\usepackage[caption=false]{subfig} 
\usepackage{nicematrix}
\usepackage{amsopn}
\usepackage{booktabs} 
\usepackage{amsmath}
\usepackage{algorithm}
\usepackage{algpseudocode}
\usepackage{enumitem}

\DeclareMathOperator{\relu}{ReLU}
\DeclareMathOperator{\sigmoid}{Sigmoid}

\begin{document}
\title{
From Compass and Ruler to Convolution and Nonlinearity:
On the Surprising Difficulty of Understanding a Simple CNN Solving a Simple Geometric Estimation Task
}
%
%
\author{Thomas Dag\`es
\and
Michael Lindenbaum
\and
Alfred M. Bruckstein
}
%
\authorrunning{T. Dag\`es et al.}
%
\institute{Department of Computer Science, Technion Israel Institute of Technology, Haifa, Israel\\
\email{thomas.dages@cs.technion.ac.il}
}
%

\titlerunning{From Compass and Ruler to Convolution and Nonlinearity}

\maketitle              
\begin{abstract}

    Neural networks are omnipresent, but remain poorly understood. Their increasing complexity and use in critical systems raises the important challenge to full interpretability. 
    We propose to address a simple
    well-posed learning problem: estimating the radius of a centred 
    pulse in a one-dimensional signal or of a centred disk in two-dimensional images using a simple convolutional neural network.
    Surprisingly, understanding what trained networks have learned is difficult and, to some extent, counter-intuitive.
    However, an in-depth theoretical analysis in the one-dimensional case allows us to comprehend constraints due to the chosen architecture, the role of each filter and of the nonlinear activation function, and every single value taken by the weights of the model.
    Two fundamental concepts of neural networks arise: the importance of invariance and of the shape of the nonlinear activation functions%
    .

\end{abstract}

\section{Introduction}

Deep neural networks have become a universal tool of choice for solving complex machine learning tasks with outstanding performance \cite{lecun2015deep,goodfellow2016deep} in all fields of signal processing. Furthermore, increasing network complexity, e.g. their depth, is known to benefit performance when they are properly trained, justifying their use of sometimes hundreds of millions of parameters \cite{he2016deep}.

Unfortunately, increasing complexity also lessens network interpretability. What is the network doing? What did it extract from the data? How and why did it learn this concept, and why not that other concept? These questions and many more have become impossible to answer directly.
As Humans can only understand explanations using a handful of concepts, 
the hundred million tweaked parameter networks are unfathomable, even though each weight is known, regardless of any intuition of architectural components prior to training.

Rather than considering them as black boxes, a growing research community, also motivated by advances in legislation \cite{goodman2017european}, seeks to open the box and try to understand what is going on in neural networks \cite{adadi2018peeking}. Understanding in networks can be viewed from three perspectives \cite{fan2021interpretability}: understanding what a trained network has learned \cite{zeiler2014visualizing} (post hoc), incorporating understanding concepts directly in the design and training of networks \cite{chen2016infogan} (ad hoc), or uncovering general mysteries such as the surprising ability to not overfit in overparametrised models \cite{belkin2019reconciling}.

However, understanding methods are usually designed in contexts where not only the network, but also the general problem or data are not well understood, leading to extra confusion. In this paper, we call such problems ill-posed, in contrast to the well-posed ones where everything is fully understood: the data, the target function and its relationship to the data.

In this paper, we revisit neural network understanding by returning to a well-posed problem, where everything is simple, fully controlled, and understood. To do so, we look towards geometry, which provides a framework combining strongly understood mathematical concepts with highly intuitive visual ones. We mathematically study a simple CNN for estimating the width of a one-dimensional pulse or the radius of a disk in an image. We analytically design all the network's weights and show that they are consistent with the learned ones. To the best of our knowledge, fully designing each weight has not been done before (even for the simplest networks). Our contributions are threefold:
\begin{enumerate}[label=(\alph*)]
    \item We show that even the seemingly simplest problems are harder than expected and require a network of certain complexity to succeed. 
    \item We mathematically analyse  a simple CNN and provide expressions for its full design (i.e. all weights), which explain the empirically learned network.
    \item Important concepts are re-discovered by our analysis, especially invariance and the shape of pointwise nonlinearities, such as the number of plateaux. 
\end{enumerate}



This paper is accompanied by 
an Appendix (denoted App.),
where further information and discussions are pushed to.

\section{Exercise on simple one-dimensional pulse functions}

We focus on a simple mathematically well-defined task and its learning-based solution. Although trivial-looking, interesting and surprising insights can be found. The goal is to fully understand the learned neural network, including its weights.

\subsection{Task and data}

We study the estimation of the (half-)width of centred one-dimensional rectangular pulse signals defined on the unit interval $\Omega=[0,1]$. For consistency with the later discussed two-dimensional problem, we use the term radius rather than half-width of the pulse. The clean data thus consists in randomly generated one-dimensional pulse signals $f_\theta^{CL}(x)\in[0,1]$ sampled $D$ times such that:
\begin{equation}
\label{eq: step clean}
    f_\theta^{CL}(x) = 
    \begin{cases}
        b & \text{if } |x-x_m| > r\\
        f & \text{if } |x-x_m| \le r,
    \end{cases}
\end{equation}
where $\theta = (r,b,f)$ are the random intrinsic signal parameters consisting in the radius of the pulse and the background and foreground intensities, with minimum contrast $\delta$ between them $|f-b|>\delta>0$, and $x_m  = \tfrac{1}{2}$ the centre of the domain and of all the pulses.  Signals with $f>b$ have positive polarity. The clean data is blurred with a Gaussian convolution filter $g_{\sigma_g}$ of standard deviation $\sigma_g$ and then contaminated with additive i.i.d. Gaussian noise $n\sim\mathcal{N}(0,\sigma_n^2 I_D)$:
\begin{equation}
    f_\theta = g_{\sigma_g}*f_\theta^{CL} + n.
\end{equation}

See App. \ref{subsec: appendix generating 1d dataset} for more details, especially for the distribution of $\theta$.

\subsection{Neural model}
\label{subsec: neural model}

\begin{figure}[ht]
    \centering
    \includegraphics[width=0.5\textwidth]{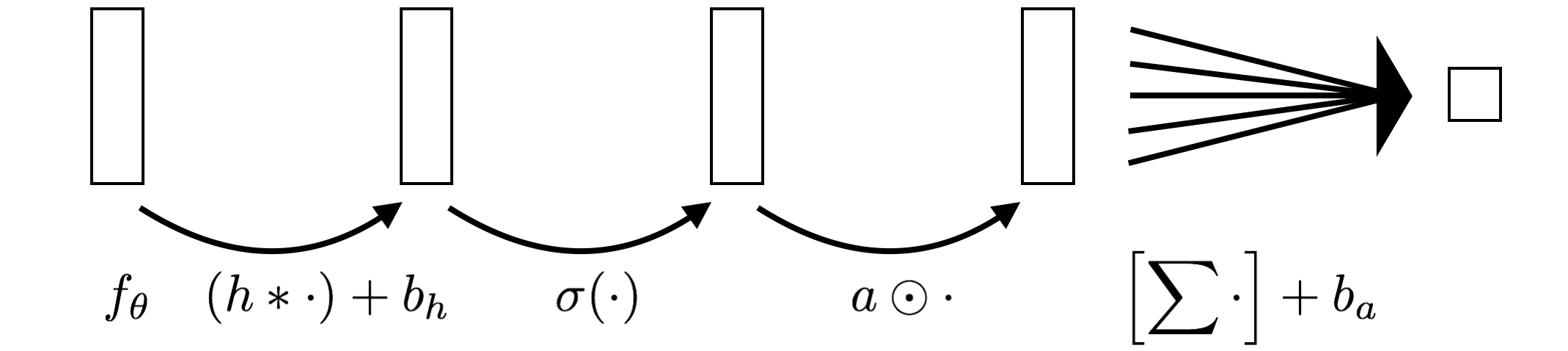}%
\caption{Neural network structure chosen for radius estimation.}
\label{fig: MAIN cnn struct step}
\end{figure}

The model $\mathcal{R}$ estimates the radius $r$ of the signal $f_\theta$. We constrain its architecture to the simplest convolutional neural network (CNN), having only one hidden layer: it has a single convolution\footnote{Following common practice in the literature, $h$ is actually a correlation operator.} layer $h$ with $C=1$ channel with additive bias $b_h$, followed by a $\sigma=\relu$ pointwise nonlinearity and finally a fully connected layer $a$ with additive bias $b_a$ (see Figure \ref{fig: MAIN cnn struct step}). For interpretability, the convolution filter has a small support to perform simple and local operations only. With abuse of notation, $h$ (resp. $a$) represents both the convolution (resp. linear) operation and the filter (resp. weight map) of the operation. For more details on the network architecture see App. \ref{subsec: appendix 1d network architecture}. Thus:
\begin{equation}
\label{eq: estimation discrete 1d}
    \mathcal{R}(f_\theta) = a\sigma(h*f_\theta + b_h) + b_a = \left[\sum\limits_{i\in\{1,\hdots,D\}} a_{i} \sigma(h*f_\theta + b_h)_{i}\right] + b_a.
\end{equation}



The literature nowadays mostly works on deep neural networks, with many more layers, more channels per layers, and more complex operations, e.g. skip connections, or optimisation tricks, e.g. normalisation or pooling. In fact, practical networks are overparametrised, which might be a key reason in the success of neural methods \cite{belkin2019reconciling,nakkiran2021deep}. However, our network is clearly underparametrised to allow interpretability. Thus, our findings give interesting insights in neural models but are far from explaining the secrets of deep learning in real applications.

The network is trained with a gradient descent-like optimisation to minimise the estimation's mean squared error (MSE) (see App. \ref{subsec: appendix 1d training procedure} for more details).

\begin{figure}[!ht]
    \centering
    \subfloat[]{\label{fig: MAIN step cnn relu weights standard context h R+-}
         \includegraphics[width=0.25\textwidth]{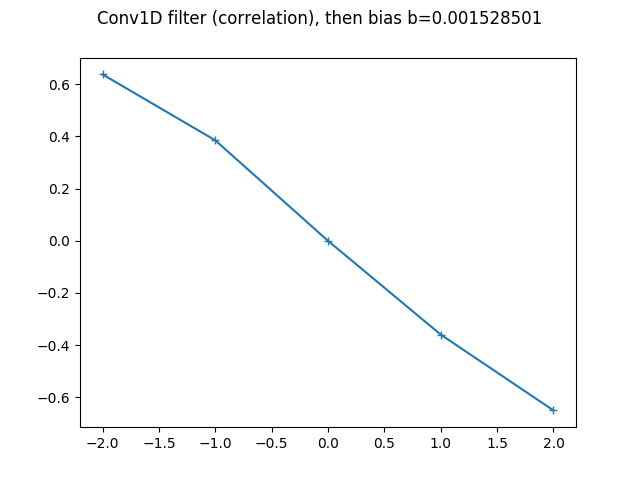}}
    \hspace{-1.5em}
    \subfloat[]{\label{fig: MAIN step cnn relu weights standard context a R+-}
         \includegraphics[width=0.25\textwidth]{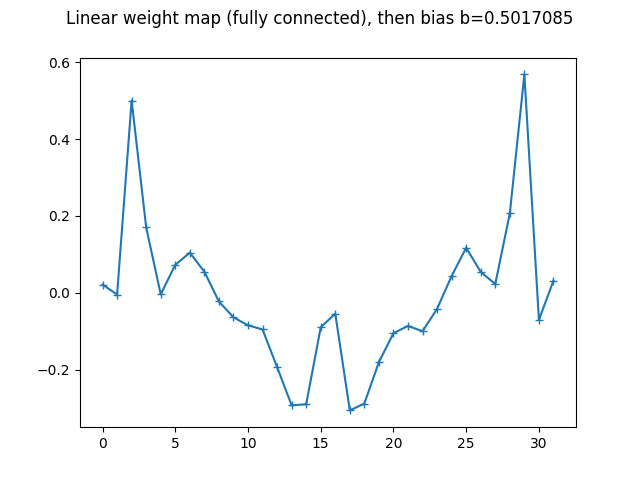}}
    \hfill
    \subfloat[]{\label{fig: MAIN step cnn relu weights standard context h R+}
         \includegraphics[width=0.25\textwidth]{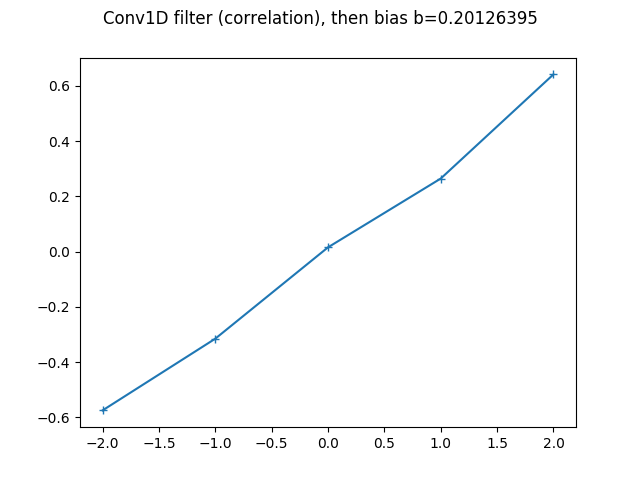}}
    \hspace{-1.5em}
    \subfloat[]{\label{fig: MAIN step cnn relu weights standard context a R+}
         \includegraphics[width=0.25\textwidth]{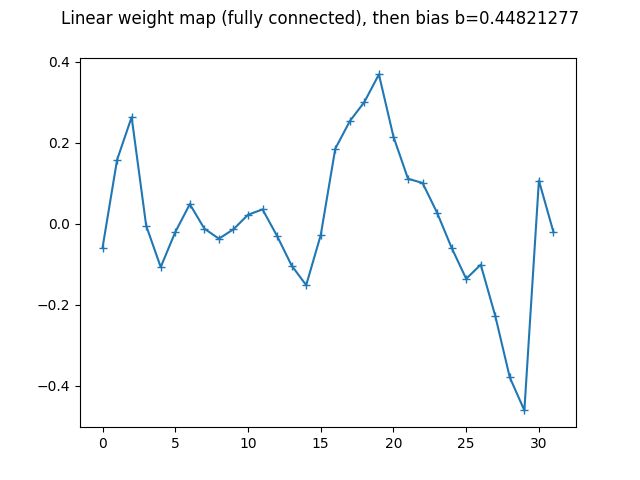}}
    \hfill\\
\caption[]{Learned CNN weights. 
The correlation filter (\ref{fig: MAIN step cnn relu weights standard context h R+-}) and fully connected layer (\ref{fig: MAIN step cnn relu weights standard context a R+-}) of the $\mathcal{R}_\pm$ network trained on both polarity data. The same for (\ref{fig: MAIN step cnn relu weights standard context h R+}) and (\ref{fig: MAIN step cnn relu weights standard context a R+}) but for the $\mathcal{R}_+$ network trained on positive-only polarity data.}
\label{fig: MAIN step cnn relu weights standard context}
\end{figure}

We present the learned network weights in Figure \ref{fig: MAIN step cnn relu weights standard context}%
\footnote{In App. \ref{subsec: appendix 1d varying D no noise no blur} we present learned networks in the noiseless and non blurry case and also intermediate representations of several instances at each step of the networks.}%
, for networks trained either on datasets with positive-only ($\mathcal{R}_+$) or both polarities present ($\mathcal{R}_\pm$). At first glance, we surprisingly do not trivially understand what they mean and how the networks are basing their estimation. The convolution filter could be seen as a derivative, but what about its additive bias or the fully connected layer?



A close performance analysis of the networks (see App. \ref{subsec: appendix 1d on quality of converged networks}) shows that only $\mathcal{R}_+$ estimates correctly the radius, up to noise, but unexpectedly not $\mathcal{R}_\pm$ even though the problem looks so simple. We will explain the limited expressiveness of $\mathcal{R}_\pm$ (and the learned weights of $\mathcal{R}_+$) using analytical arguments.

\subsection{What would a neural engineer do?}

 To better understand the learned models, we suggest to analyse the problem from their perspective. Should an engineer fully design a neural network (including its weights), he would trade his regular tools, such as compasses and rulers, to those authorised by the network, mainly convolutions and predetermined nonlinearities. 
One can design a succession of common operations to solve the problem. However, some operations cannot be  carried out by neurons. 
In principle, each operation may be approximated by its own network, as implied by the universal approximation property \cite{cybenko1989approximation,hornik1991approximation}. Unfortunately, this approach is not compatible with a single simple CNN with small convolution support and a fairly small depth.   
%
This means that we need to find another way to design a neural solution. 
 
For clarity, our discussion below uses several simplifications. For mathematical ease, our analysis is in the continuum, i.e., $D\to\infty$, although highly oscillating functions (in small areas the size of a pixel) are excluded to compare with finite resolution signals. For easing the analysis and maximising understanding, we assume that the data is approximately clean $f_\theta\approx f_\theta^{CL}$, i.e., $\sigma_b\approx\sigma_n\approx0$.

\subsubsection{Exemplar exercise}

Recall that the goal is to manually design the weights of a convolutional neural network, while respecting its constrained architecture, that correctly estimates the radius of a centred pulse.

Intermediate representations are denoted as follows: $f_h = h * f_\theta$, $f_{hb} = f_h + b_h$, $f_\sigma = \sigma(f_{hb})$, $f_{a\odot} = a\odot f_\sigma$, $f_{\int a} = \int_\Omega f_{a\odot}(x)dx$, and $\mathcal{R}(f_\theta) = f_{\int a} + b_a$. Note that the discrete sum in Equation (\ref{eq: estimation discrete 1d}) become an integral in the continuum.

We focus on the positive polarity sub-problem $\mathcal{R}_+$, i.e., $f>b$ for all signals. To estimate $r$ using a CNN with a convolution with small support, it seems natural to choose an unbiased derivative filter\footnote{The filter $h$ is unbiased if $\int_\Omega h = 0$.} whose non zero responses are located at the edges of the pulse. In fact, a detailed analysis proves that this is the only correct choice (see App. \ref{subsec: 1d extensions unbiased diff conv operators}). Denote $\alpha$ its gain and $\Delta$ the size of its small support. Without loss of generality, assume that $f_h(x)$ is positive when $f_\theta$ is locally increasing at $x$\footnote{In other words, $h(-\tfrac{\Delta}{2})<0$ and $h(\tfrac{\Delta}{2})>0$ for the correlation filter $h$.}: $h$ is then said to be a positive unbiased derivative filter.

Given a positive polarity signal $f_\theta$, the output of the convolution is (approximately) $0$ everywhere except at the borders of the pulse: $f_h$ has two narrow bell curves, called peaks, of height $\pm\alpha\tfrac{f-b}{2}$ and width $2\Delta$. For simplicity, we assume that $\Delta$ is small enough for the bell curves to approximately be small pulses.

If $b_h \le 0$, after the $\relu$ activation, only the left peak of $f_{\sigma}$ remains non zero.  Its thresholded height is affine in the signal intensity difference, which is not desirable, and so is $\mathcal{R}(f_\theta)$ or any other non-constant affine function of it.


For $b_h> 0$, $f_{hb}(x) \approx b_h$ everywhere except at the peaks which are translated upwards by $b_h$. If $b_h\ge\tfrac{\alpha}{2}$, then $f_{hb}(x)\ge 0$ for all $x$, and $\sigma$ acts as the identity on all signals. Thus, for any $x$, $f_\sigma(x)$ as well as the final estimation are incorrectly either  constant or affinely  depending on the intensity difference $f-b$. In fact, if $\tfrac{\alpha\delta}{2}\le b_h \le \tfrac{\alpha}{2}$, the same reasoning applies on the authorised low-contrast signals. Therefore, we must have $0< b_h < \tfrac{\alpha\delta}{2}$, e.g. $b_h=  \tfrac{\alpha\delta}{4}$.

With our chosen $h$ and $b_h$, $f_{hb}$ is always positive everywhere, except on the right downward peak. Thus $f_\sigma$ is identical to $f_{hb}$ except on this right peak, which has been thresholded to $0$ with constant height $b_h$ regardless of $f$ and $b$ and is now called a ``drop''.  Note that regardless of $r$, since we only focus on positive polarity signals, $f_\sigma$ is invariant to the intensities $f$ and $b$ in $(\tfrac{1}{2}, 1]$,  but not in the other non overlapping interval $[0,\tfrac{1}{2})$.

We now design the weight map $a$. The goal is to integrate the weighted activated signal $f_{a\odot}$ and obtain the radius up to a final additive constant. If we do not discard the left peak by weighting it to $0$, then the estimation unacceptably explicitly depends on the intensities in an affine non trivial way. Since the peak could be located anywhere in $[0,\tfrac{1}{2}]$, we need to choose $a_{\big| [0,\tfrac{1}{2})}\equiv 0$. The radius must then be affinely inferred solely by the drop in $(\tfrac{1}{2}, 1]$. Since it has constant height, an affine weight map $a_{\big|[0,\tfrac{1}{2}]}$ can locate it. Let us derive this result more formally. Denote $P_l = [\tfrac{1}{2} - r - \Delta, \tfrac{1}{2} - r + \Delta]$ and $P_r = [\tfrac{1}{2} + r - \Delta, \tfrac{1}{2} + r + \Delta]$ the domains of the left and right peaks. The estimation is:
\begin{align}
    \mathcal{R}(f_\theta) &\approx b_h \int_{\Omega} a(x)dx + \alpha\tfrac{f-b}{2}\int_{P_l} a(x)dx - b_h\int_{P_r} a(x)dx +b_a. \label{eq: step relu pos polarity unbiased pred still has f-b}
\end{align}

Only the second term in Equation (\ref{eq: step relu pos polarity unbiased pred still has f-b}) depends on the intensities, thus to make the $r$ estimator,  $\mathcal{R}(f_\theta)$, independent from $f-b$, we must ensure that:
\begin{equation}
    \label{eq: step relu pos polarity int left peak is 0}
    \int_{P_l}a(x)dx = \int_{\frac{1}{2}-r-\Delta}^{\frac{1}{2}-r+\Delta} a(x)dx = 0.
\end{equation}

For moderately small $\Delta$, and recalling that highly oscillating behaviours of $a$ are not allowed,  we have  $\int_{P_l} a(x)dx \approx 2  \Delta \cdot a({\scriptstyle \frac{1}{2}}\!-\!r)$. Equation (\ref{eq: step relu pos polarity int left peak is 0}) then becomes $a({\scriptstyle \frac{1}{2}}\!-\!r)=0$ for any $r$, and thus:
\begin{equation}
    a_{\big|[0,\frac{1}{2}]}\equiv 0.
\end{equation}

Then, 
similarly approximating $\int_{P_r}a(x)dx$, Equation (\ref{eq: step relu pos polarity unbiased pred still has f-b}) becomes:
\begin{equation}
    \label{eq: setp relu pos polarity int taylor}
    \mathcal{R}(f_\theta)  \approx b_h \int_{\Omega} a(x)dx - b_h \cdot 2 \Delta\cdot a({\scriptstyle \frac{1}{2}}\!+\!r) +b_a.
\end{equation}

By requiring  $\mathcal{R}(f_\theta)=r$, as desired, and rearranging terms, we find that the coefficient $a(x)$ at $x=\tfrac{1}{2}+r$ must satisfy:
\begin{equation}
    \label{eq: step relu pos polarity a right domain}
    a({\scriptstyle \frac{1}{2}}\!+\!r) 
    =  \frac{1}{ 2\Delta}\int_{\Omega}a(x)dx + \frac{b_a}{2 \Delta b_h} - \frac{r}{2 \Delta b_h}.
\end{equation}

Thus, $a$ is affinely decreasing in $[\tfrac{1}{2},1]$ with slope $-\tfrac{1}{2\Delta b_h}$, e.g. $-\tfrac{4}{2 \alpha \Delta\delta}$ for $b_h=\tfrac{\alpha\delta}{4}$. Note that Equation (\ref{eq: step relu pos polarity a right domain}) only specifies $a(x)$ in  $[\tfrac{1}{2},1]$ up to an offset.
If we also ask for minimal $L^2$ norm of $a(x)$ (for regularisation), then the minimal norm is obtained by centering $a_{\big|[\tfrac{1}{2},1]}$ by taking $a(\tfrac{3}{4})=0$. This implies that $\int_{[\tfrac{1}{2},1]}a = \int_\Omega a = 0$. Using Equation (\ref{eq: step relu pos polarity a right domain}) with $r=\tfrac{1}{4}$ then implies that the final bias is the mean radius over the entire dataset:
\begin{equation}
    b_a = \frac{1}{4},
\end{equation}
which in turn means that we need to choose a weight map of the form:
\begin{equation}
    a({\scriptstyle \frac{1}{2}}\!+ \!r) = \frac{1}{4}-\frac{r}{2 \Delta b_h}.
\end{equation}

\begin{figure}[ht]
    \centering
    \includegraphics[width=0.8\textwidth]{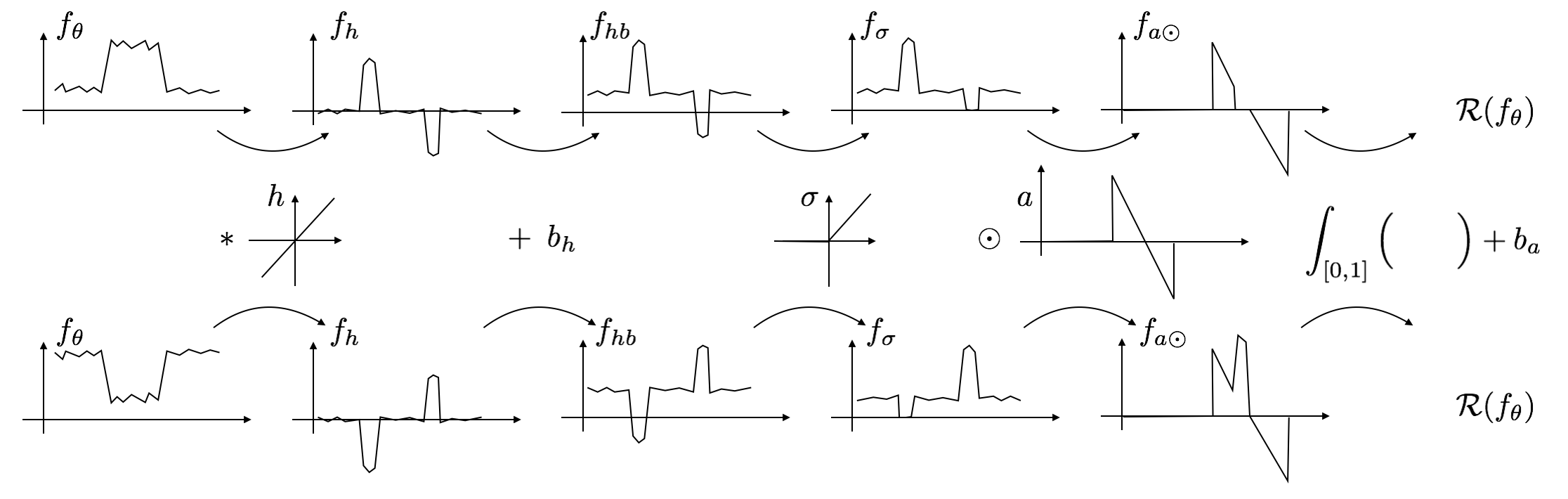}
    \caption[]{Designed $\relu$ network for estimating the radius on positive polarity data applied to a positive polarity (top) and a negative polarity (bottom) signal. The network cannot estimate correctly the radius in the latter case.}
    \label{fig: MAIN theory step relu both pol}
\end{figure}

Figure \ref{fig: MAIN theory step relu both pol} (top) illustrates the derived network and the associated intermediate signals for estimating the radius on positive polarity signals. We summarise what we found in a proposition, and we refer to the previous text for all the ``mild assumptions'' under which it holds.


\begin{proposition}
    \label{prop: step prop relu pos polarity only choice}
    Given a single-channel and single-depth CNN working with $\relu$ activation designed to estimate the radius of centred pulse signals with positive polarity, if its correlation filter $h$ is a positive unbiased derivative filter,
    then, to succeed under ``mild assumptions'', the convolution bias $b_h$ must be strictly positive but small and there must exist scalar constants $\beta_1$ and $\beta_2>0$ such that the fully connected layer's weight map $a(x)$ is of the following form:
    $$ a(x) = 
    \begin{cases} 0 & \text{if } 0\le x<\frac{1}{2}\\ \beta_1 - \beta_2 (x-\frac{1}{2}) &\text{if } \frac{1}{2} \le x \le 1 \end{cases}.
    $$
\end{proposition}

The weights designed by the proposed analysis closely resemble the empirically learned one from Section \ref{subsec: neural model} and those from App. \ref{subsec: appendix 1d varying D no noise no blur} which were trained without noise or Gaussian blur with increasing resolution $D$, thus removing most of the mystery around them. Nevertheless, differences between the theory and practice are clearly visible, and they are mostly due to the assumptions that we used that do not necessarily hold in reality.

\subsubsection{Extensions}   

We can extend our line of reasoning to similarly derive other results. 
We present here the main ideas (details are in the Appendix).

\paragraph{Changing the convolution filter} If we replace the derivative filter by a local filter of higher order (e.g. second derivative), then we will not be able to find weights to correctly estimate the radius in all cases, as upwards and downwards peaks of affine height in the intensity difference will exist at both the left and right border of the pulse, making it impossible to isolate a meaningful invariant part of the activated signal (see App. \ref{subsec: 1d extensions unbiased diff conv operators}).


\paragraph{Handling both polarities} So far, we only focused on the subcase $\mathcal{R}_+$ of $\mathcal{R}_\pm$. As there was only one choice of correct weights for the network in $\mathcal{R}_+$, we must work with it for $\mathcal{R}_\pm$. But then, $f_\sigma$ associated to $f_{(r,f,b)}$ (negative polarity) is horizontally flipped around $x_m$ compared to $f_\sigma$ associated to $f_{(r,b,f)}$ (positive polarity) (see Figure \ref{fig: MAIN theory step relu both pol}). Thus, the invariant part of the signal lies in $[0, x_m]$ and is zeroed out by $a$, whereas the right peak is now undesirably upwards and affinely depending on the intensities, and so will too the final estimation. Our trivial network strucutre is thus not expressive enough in the general case, which we summarise in the following proposition.


\begin{proposition}
    \label{prop: step prop relu edge filter does not work for both pola}
    Given a single-channel and single-depth CNN working with $\relu$ activation designed to estimate the radius of centred pulse signals with any polarity, if the correlation filter $h$ is a derivative filter,
    then, under ``mild assumptions'', the network cannot correctly estimate the radius in all cases. In other words, the network cannot correctly learn the concept of a radius.
\end{proposition}

In fact, this result extends to higher order local filters as they fail on $\mathcal{R}_+$.

\begin{figure}[ht]
    \centering
    \includegraphics[width=\textwidth]{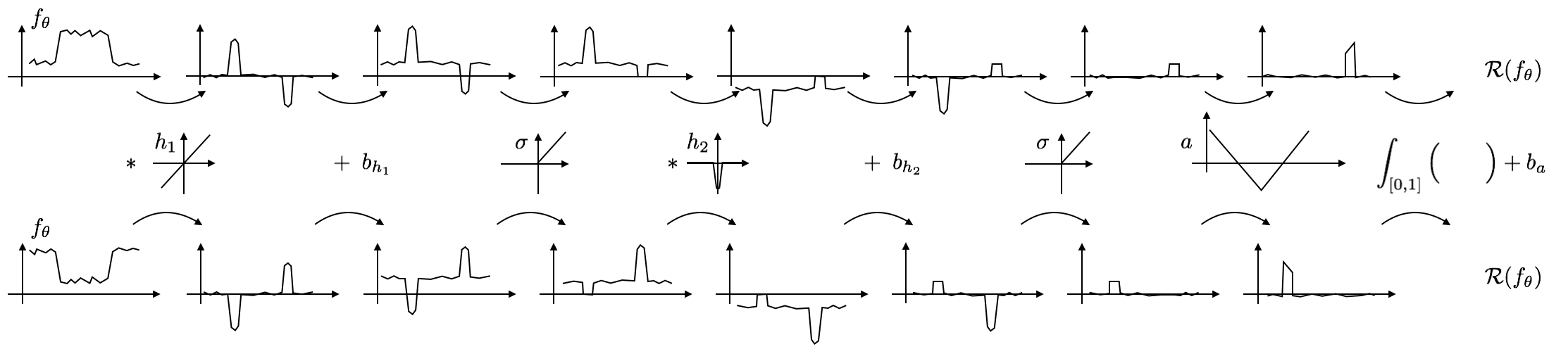}
    \caption[]{Designed $\relu$ network with two convolution layers for correctly estimating $r$ on both polarities applied to a positive (top) and negative (bottom) polarity signal.}
    \label{fig: MAIN theory step relu depth 2 both pol}
\end{figure}

\paragraph{Changing the architecture: going deeper}  Adding another convolution layer resolves the expressiveness issue for $\mathcal{R}_\pm$, as it allows two sequential thresholds to cut off both peaks. For instance, take the positive derivative filter with its small $b_h>0$ as previously for the first layer. The first $\relu$ thresholds the downward peak. We then flip the signal by convolving it with a minus Dirac delta function and translate the output upwards by $b_h$ again (see Figure \ref{fig: MAIN theory step relu depth 2 both pol}). The second $\relu$ then cuts off entirely the new downward peak, creating a signal globally invariant to $f$ and $b$ that is approximately $0$ everywhere except for a constant height $b_h$ peak on one edge of the pulse. A $V$-shaped $a$ horizontally symmetric around $x_m$ can then correctly locate it. This may not be the only correct 2-layer network (see App. \ref{subsec: 1d extensions going deeper}).

\paragraph{Changing the architecture: adding more channels} Adding more channels helps, to some extent, for handling the case with both polarities present in the dataset, however its analysis seems complex, see empirical evidence for the two-dimensional case considered below in App. \ref{subsec: appendix circle more conv channels}.

\subsubsection{Modified exercise: from $\relu$ to $\sigmoid$} 

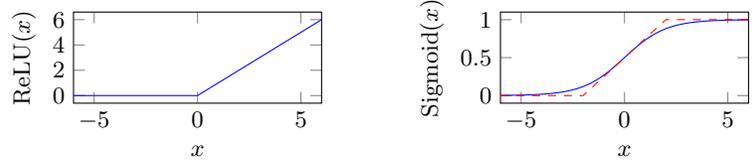
\begin{figure}[!ht]
    \vspace{-2em}  
    \centering
    \subfloat{
        \begin{tikzpicture}
              \begin{axis}[ 
                    xlabel=$x$,
                    ylabel={$\relu(x)$},
                    width=0.4\textwidth, height=1.1in,
                    no marks,
                    ylabel near ticks,
                    xlabel near ticks,
                    xmin=-6, xmax=6,
                    domain=-6:6
                ] 
                \addplot {max(x,0)}; 
              \end{axis}
        \end{tikzpicture}
    }
    \quad\quad\quad
    \subfloat{
        \begin{tikzpicture}[
            declare function={
                func(\x)= (\x < -2) * (0) + and(\x >= -2, \x < 2) * (1/4*\x + 1/2) + (\x >= 2) * (1) ;
            }]
          \begin{axis}[ 
            xlabel=$x$,
            ylabel={$\sigmoid(x)$},
            ylabel near ticks,
            xlabel near ticks,
            width=0.4\textwidth, height=1.1in,
            no marks,
            xmin=-6, xmax=6,
            domain=-6:6,
            samples=100
          ]
            \addplot {1/(1+e^(-x))}; 
            \addplot [red, dashed] {func(x)};
          \end{axis}
        \end{tikzpicture}
    }
\caption[]{Activation functions with in dashed red a piecewise affine $\sigmoid$ ($\tau=2$) showing its three regimes: upper and lower thresholding and intermediate linearity.}
\label{fig: step activation functions}
\end{figure}

To overcome the limited expressiveness, we can also change our neural tools. 
Using the universal approximation theorem \cite{cybenko1989approximation,hornik1991approximation} requires to switch the local convolution to a wide and multi-channel fully connected layer, but we would lose in interpretability. 
Instead, we change the nonlinearity. To threshold both above and below, we take $\sigma = \sigmoid$. Although traditionally $\sigmoid(x) = \tfrac{1}{1+e^{-x}}$, we consider a piecewise linear approximation which is flat and $0$ (resp. $1$) for negative (resp. positive) numbers with magnitude greater than $\tau$
and linear in the middle range (see Figure \ref{fig: step activation functions}).

\begin{figure}[ht]
    \centering
    \includegraphics[width=0.8\textwidth]{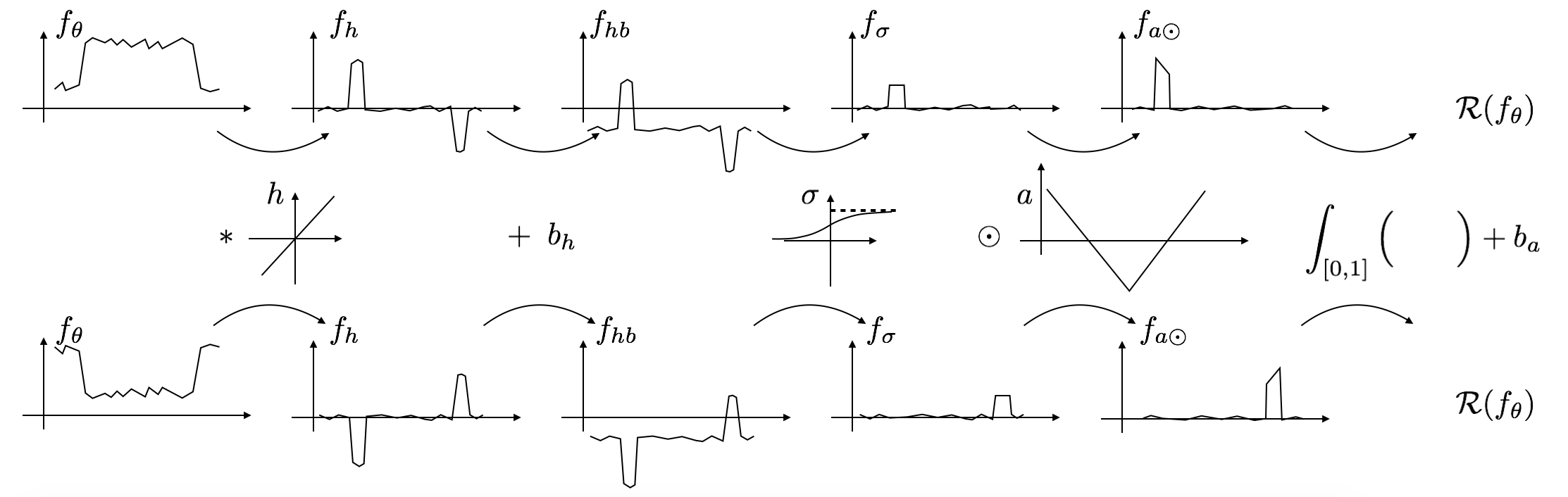}
\caption[]{Designed $\sigmoid$ network for accurately estimating $r$ on both polarities applied to a positive (top) and a negative (bottom) polarity signal.}
\label{fig: MAIN theory step sigmoid filter unbiased both pol}
\end{figure}

Unlike the $\relu$ case, we can design a one-layer $\sigmoid$ network for estimating the radius when both polarities are possible (see App. \ref{subsec: 1d sigmoid}). The idea is to increase the gain $\alpha$ of the local derivative filter\footnote{The reasoning can be easily generalised to higher order local filters.} to push the positive peak into a flat domain of the $\sigmoid$ and at the same time use a large bias $b_h=-\tau$ to push the rest to the other flat domain (see Figure \ref{fig: MAIN theory step sigmoid filter unbiased both pol}). This symmetry-breaking step is necessary because if both peaks remain, then the affine function cannot handle the change of polarity. Then, only the upward peak of $f_{hb}$ is not zeroed out by the $\sigmoid$ and is thresholded to a constant height, and its position is measurable with a $V$-shaped $a$ as done previously.


The advantage of the $\sigmoid$ network is the extra flexibility provided by its additional flat region compared to the $\relu$. Indeed, since the data is bounded, we could theoretically rescale the filters so that both nonlinearities behave similarly: either thresholding to $0$ or the identity for positive inputs. This suggests that nonlinearities with increasing number of flat regions or plateaux provide more flexible tools for learning appropriate invariant concepts, and one could question whether choosing a nonlinearity with many more (e.g. a sine nonlinearity as in \cite{sitzmann2019siren}) plateaux would not be better altogether than the simplistic $\relu$ for neural networks in general\footnote{Although $\relu$'s simplicity allows fast computations and contributes better to overcoming vanishing and exploding gradients in deep networks with its $0-1$ gradient.}.



\section{From one-dimensional signals to images}

The previous exercise can be generalised to higher dimensions to analyse a simple yet natural geometric learning problem. In two dimensions, the signals are greyscale images defined on the unit square $\Omega = [0,1]^2$. The task is to estimate the radius of a centred disk. The clean data consists in randomly generated two-dimensional disk signals $f_\theta^{CL}(x)\in[0,1]$ sampled $D\times D$ times such that:
\begin{equation}
    f_\theta^{CL}(x) = 
    \begin{cases}
        b & \text{if } \lVert x-x_m\rVert_2 > r\\
        f & \text{if } \lVert x-x_m\rVert_2 \le r,
    \end{cases}
\end{equation}
where the random intrinsic parameters $\theta = (r,b,f)$ have the same meaning and are generated as in the one-dimensional case with minimum contrast $\delta$, and the centre of the domain is now $x_m = (\tfrac{1}{2}, \tfrac{1}{2})$. The clean data is blurred with a Gaussian filter $g_{\sigma_g}$ of standard deviation $\sigma_g$ and then contaminated with additive i.i.d. Gaussian noise of standard deviation $\sigma_n$.

The neural model is once again a single layer and single channel CNN, where the convolution is now two-dimensional, with $\sigma=\relu$ nonlinearity.


Although seemingly trivial and artificial, the presented task has real applications. For instance, in Astrophysics, some researchers work on almost centred images of the solar disks, that closely resemble the data presented here: images of a perfect bright disk of uniform intensities over a uniform dark background. In fact, \cite{zhu2020sun} used for radius estimation a modification of VGG \cite{simonyan2014vgg}, which is an extremely complex deep network with 10 convolution layers and 3 fully connected ones, illustrating the difficulty of the task. Unfortunately, their well-performing network is far too complex to be understandable, which is not our goal.

\begin{figure}[htbp]
    \centering
    \subfloat[]{\label{fig: MAIN circle cnn relu weights c 1 h R+-}
         \includegraphics[width=0.15\textwidth]{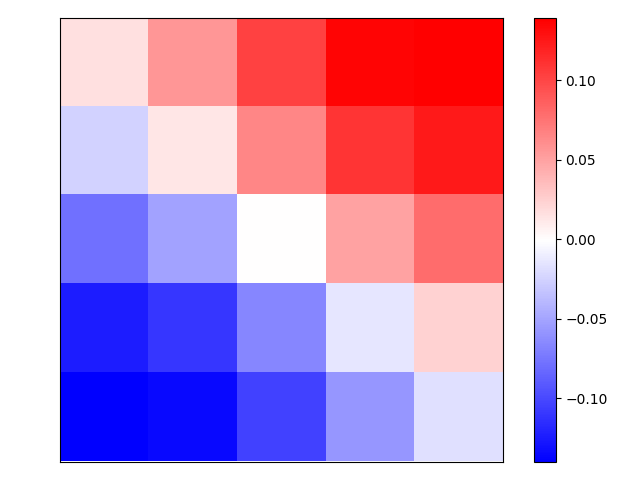}
    }
    \!\!\!\!\!\!\!\!
    \subfloat[]{\label{fig: MAIN circle cnn relu weights c 1 a R+- cut overlap dashed}
        \includegraphics[width=0.15\textwidth]{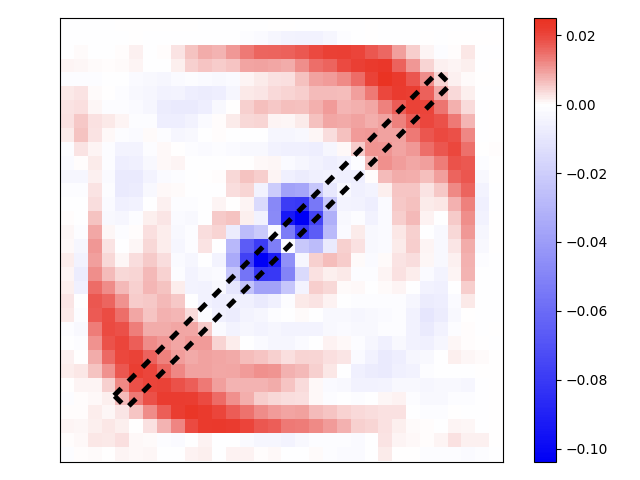}
    }
    \!\!\!\!\!
    \subfloat[]{\label{fig: MAIN circle cnn relu weights c 1 h R+- cut only}
        \includegraphics[width=0.158\textwidth]{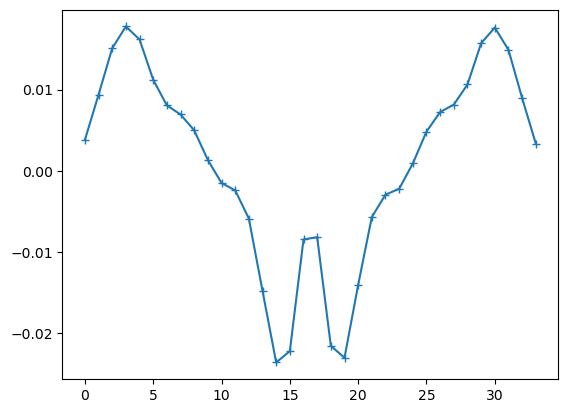}
    }
    \hspace{1em}
    \subfloat[]{\label{fig: MAIN circle cnn relu weights c 1 h R+}
        \includegraphics[width=0.15\textwidth]{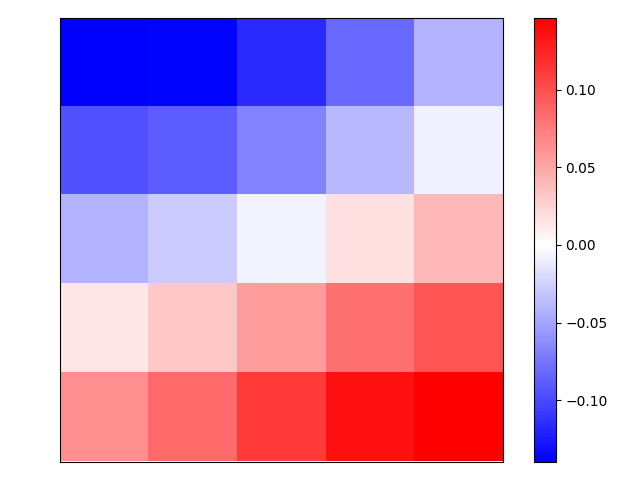}
    }
    \!\!\!\!\!\!\!\!
    \subfloat[]{\label{fig: MAIN circle cnn relu weights c 1 a R+ cut overlap dashed}
        \includegraphics[width=0.15\textwidth]{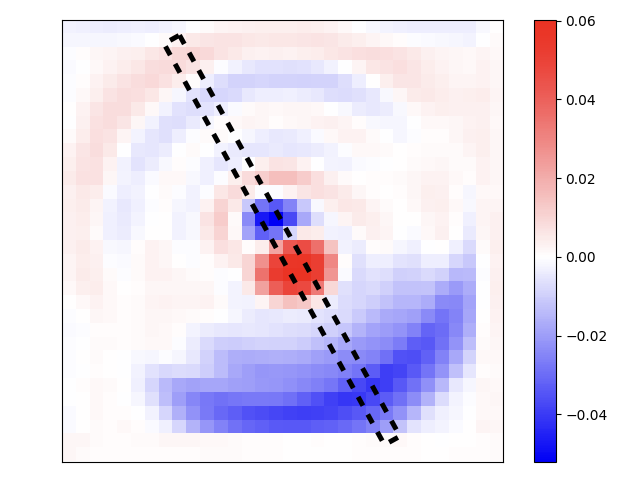}
    }
    \!\!\!\!\!
    \subfloat[]{\label{fig: MAIN circle cnn relu weights c 1 h R+ cut only}
        \includegraphics[width=0.158\textwidth]{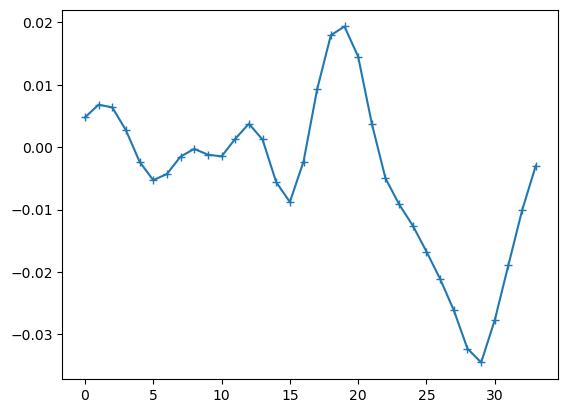}
    }
\caption[]{Learned CNN weights.
The correlation filter (\ref{fig: MAIN circle cnn relu weights c 1 h R+-}) and fully connected layer  (\ref{fig: MAIN circle cnn relu weights c 1 a R+- cut overlap dashed}) of the $\mathcal{R}_\pm$ network trained on both polarity data. The same for the $\mathcal{R}_+$ network trained on positive-only polarity data in (\ref{fig: MAIN circle cnn relu weights c 1 h R+})-(\ref{fig: MAIN circle cnn relu weights c 1 a R+ cut overlap dashed}). 
Colours: white-0, blue-negative, red-positive.
A cut along the edge filter's direction  (\ref{fig: MAIN circle cnn relu weights c 1 h R+-}) 
is super-imposed on the fully connected layer (\ref{fig: MAIN circle cnn relu weights c 1 a R+- cut overlap dashed}) as a dashed rectangle with associated profile in (\ref{fig: MAIN circle cnn relu weights c 1 h R+- cut only}).  The analogous profile for the $\mathcal{R}_+$ network (\ref{fig: MAIN circle cnn relu weights c 1 h R+})-(\ref{fig: MAIN circle cnn relu weights c 1 a R+ cut overlap dashed}) is shown in (\ref{fig: MAIN circle cnn relu weights c 1 h R+ cut only}). The profiles, especially (\ref{fig: MAIN circle cnn relu weights c 1 h R+ cut only}), resemble the corresponding fully connected layers of the one-dimensional case (Figure \ref{fig: MAIN step cnn relu weights standard context}).
}
\label{fig: MAIN circle cnn relu weights c 1}
\end{figure}

We train our model similarly to the one-dimensional case (see App. \ref{subsec: appendix circle training procedure}). We present the learned network weights in Figure \ref{fig: MAIN circle cnn relu weights c 1} for networks trained on datasets with either positive-only ($\mathcal{R}_+$) or both polarities present ($\mathcal{R}_\pm$)\footnote{To help understanding, we present intermediate representations at each step of the networks on several instances in App. \ref{subsec: appendix circle intermediate representations}.}.


Once again, a close analysis of the performance of the networks (see App. \ref{subsec: circle on quality of converged networks}) shows that only $\mathcal{R}_+$ estimates correctly, up to noise, the radius of the signals, but not $\mathcal{R}_\pm$. This discrepancy demonstrates again the limitation of the $\relu$ nonlinearity not being flexible enough.

Clearly, both $h$ correspond to edge filters, which are the direct generalisation in two-dimensions of derivatives, but both $a$ maps are not trivially understood at first glance. However, their weights along a diagonal cut following the direction of the edge filters resemble those of the one-dimensional case. 
The rest of the weights seem to result from a blurring process along the tangential direction, i.e. orthogonal to the radial direction.


A question naturally arises: what radius concept did the successful network $\mathcal{R}_+$ learn? Is it based on well-known definitions, or is it fundamentally new? We understand a radius from many perspectives since Antiquity: it is the length of the perimeter ($r = \tfrac{\mathcal{P}}{2\pi}$), the squared root of the area ($r = \sqrt{\tfrac{\mathcal{A}}{\pi}}$), or half the maximum distance between two points on the circle. Our answer to those questions is that $\mathcal{R}_+$ combines two approaches. First, the radius is found along a scan in a fixed direction passing through the centre as we have studied in the one-dimensional case. Second, this noisy estimate is filtered by generalising this approach to a cone of angle roughly $\tfrac{\pi}{4}$ around the chosen direction. 
Within this cone, the weights of the fully connected layer are approximately constant for each distance to the centre. It is thus producing a number proportional 
to a quarter of the perimeter, with proportionality given by the weights. The network has thus learned the radius as a combination of the one-dimensional radius concept with the one stemming from the perimeter.



Some aspects of our problems deserve more investigation. As an example, the impact of domain discretisation or radius quantisation (especially in the one-dimensional case) is not fully understood. Furthermore, scaling our analysis to more complex problems, especially when changing the architecture, is difficult. In particular, we point out the complexity of the multi-channel architecture extension and dedicate a lengthy discussion on it in App. \ref{subsec: appendix circle more conv channels}.

\section{Conclusion}

We defined and studied a well-posed seemingly trivial geometric  estimation problem using small neural networks. Our goal was to fully understand everything. We thus analysed theoretically and empirically not only the networks' architecture but also every single weight including the biases. Surprisingly, it was not as easy as expected, and it was at first glance unclear what the networks had learned. During our analysis, two fundamental concepts of neural networks naturally emerged. The first is the importance of invariance. The second is the importance of flat regions of pointwise nonlinear activation functions that provide thresholding tools to create invariance. We found that increasing the number of flat regions increases the expressiveness of the (small) networks by providing them more flexibility to perform complex operations in a single layer. While it is clearly unreasonable to imagine fully understanding every weight of a complex neural model in complex real-world applications, the lessons taken from our exercise can be used to improve our understanding of commonly used deep neural networks.


%
%
%
\bibliographystyle{splncs04}
\bibliography{references}

\appendix

\section{Details for one-dimensional pulse function exercise}

\subsection{Generating one-dimensional pulse datasets}
\label{subsec: appendix generating 1d dataset}

We call reference data the data generated using a reference choice of hyperparameters. Unless mentioned otherwise, the values of the hyperparameters presented here are those used in the experiments.

The reference data is $D=32$-dimensional. The domain position $x$ in Equation (\ref{eq: step clean}) refers to the centre of the pixel, with the first and last pixels at positions $x=0$ and $x=1$. To avoid boundary issues, the radii are chosen in $[\tfrac{\epsilon_r}{2}, \tfrac{1-\epsilon_r}{2}]$, where $\epsilon_r$ is typically equal to $\tfrac{1}{10}$. Note that although $r$ is sampled continuously, it generates a finite set of less than $\tfrac{D}{2}$ possible binary masks for the position of the pulse due to the crude radius quantisation in Equation \ref{eq: step clean}. The intensities $f$ and $b$ are chosen in $[0,1]$. The minimum contrast between them is chosen to be $\delta = \tfrac{50}{255}\approx 0.2$, which empirically allows clear visual separation of the background and foreground intensities. The choice for $r$, $f$, and $b$ is almost uniform and independent. Since we force a minimal contrast $\delta$ between the intensities, $f$ and $b$ are not mathematically independent: $b$ is chosen independently of $r$ and uniformly in $[0,1]$ whereas $f$ is then chosen conditionally to $b$ uniformly in the rest of the unit interval with the $\delta$ ball around $b$: $[0,1]\setminus]b-\delta, b+\delta[$\footnote{If we furthermore request only positive polarity data, then we also remove $[0,b]$ from the unit interval for uniformly sampling $f$.}. Mathematically, if we denote $R$, $F$, and $B$ the random variables with realisation $r$, $f$, and $b$, and if we call $F^+$ and $B^+$ the intensity random variables in the case of positive polarity only data, we have:
\begin{align}
    &B \sim \mathcal{U}([0,1]),\\
    &F \mid B=b \sim \mathcal{U} ([0,1]\setminus [b-\delta,b+\delta]),\\
    &B^+ \sim \mathcal{U}([0,1]),\\
    &F^+ \mid B^+=b \sim \mathcal{U} ([b+\delta,1]).
\end{align}

In the reference data, the Gaussian blur level is $\sigma_g = \tfrac{1}{D}$, and the noise level is $\sigma_n = \tfrac{10}{255}\approx 0.04$.

\subsection{Network details}

\subsubsection{Network architecture details}
\label{subsec: appendix 1d network architecture}

We chose to work with convolution rather than an arbitrary linear transformation using a fully connected layer to improve interpretability. While theoretically better, as convolutions are a sub-type of linear transforms, forcing convolution constrains a particular structure on the linear transformation that is simple and more easily interpretable: a (local) sharing weight scheme, with extra interpretable properties such as shift-equivariance or multiscale behaviours with cascades of them. In real world applications, choosing convolutions can also help for convergence in complex models, as fully connected layers can be particularly expensive in terms of weights and get more easily stuck in local minima, but we do not share this motivation in this work.

The convolution filter has a support of only $5$ entries to force locality and encourage interpretability\footnote{This means a support of interval length $\tfrac{5}{D-1}$, which is approximately equal to 0.16 in the reference case.}. A smaller filter of support size $3$ entries would not have much flexibility and a filter with more entries might be too complex to interpret. Note that rather than fixing an interval length we chose a number of entries for the support, meaning that when increasing the resolution $D$ the convolution gets more and more local. This effect is desirable in order to compare with the theory in the continuum, where the convolution is assumed to be a differential operation and thus with infinitely small support. Furthermore, the convolution uses the $0$ rather than circular padding strategy as it is standard in the field.

The chosen non-linearity $\sigma$ is $\relu$, the universally used activation function in the literature.  Other non linearities exist, but they have largely been overshadowed by the $\relu$ due to its computational advantages such as fast forward and gradient computation along with a binary gradient limiting vanishing and exploding gradients in deep networks. Unfortunately, the interpretability of the activation functions are rarely considered when choosing them. In this work, interpretability primes over computational considerations, and the $\relu$ is perhaps one of the simplest and best understood non linear operation: it is clear that $\relu$ is a lower thresholding operation.

\subsubsection{Training procedure}
\label{subsec: appendix 1d training procedure}

Training was done using the ADAM optimiser with default optimiser parameters and a learning rate of $\eta=0.005$ and batch size $32$. We add $L^2$ regularisation penalties with small Lagrangian coefficients for the convolution and fully connected layers but not for the biases. Training is done using $N=10000$ training samples. The validation and test sets also have the same size. Since the networks are fairly small, convergence happens rather fast, but we keep on training for a long duration to make sure that no strange phenomenon occurs.

\subsubsection{Varying the resolution in the ideal case}
\label{subsec: appendix 1d varying D no noise no blur}

\begin{figure}[!ht]
    \centering
    \subfloat{
         \includegraphics[width=0.18\textwidth]{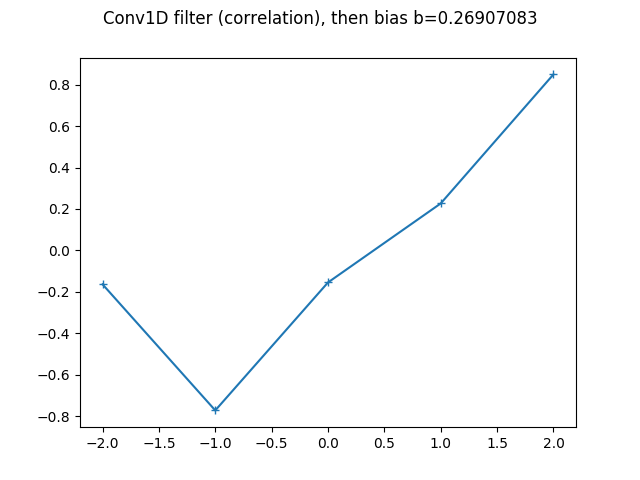}}
    \hfill
    \subfloat{
         \includegraphics[width=0.18\textwidth]{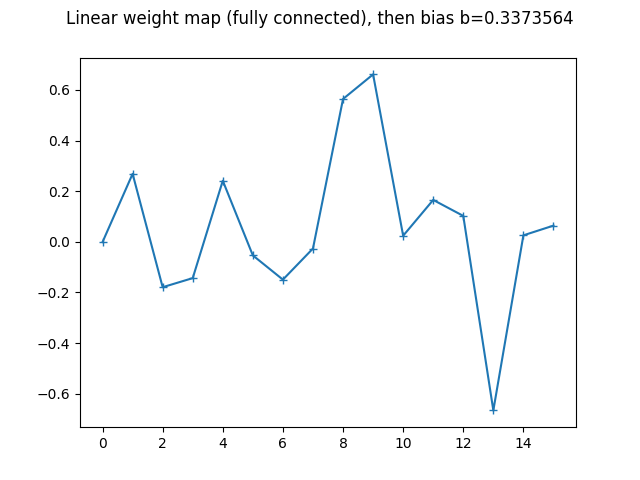}}
    \hfill
    \subfloat{
         \includegraphics[width=0.18\textwidth]{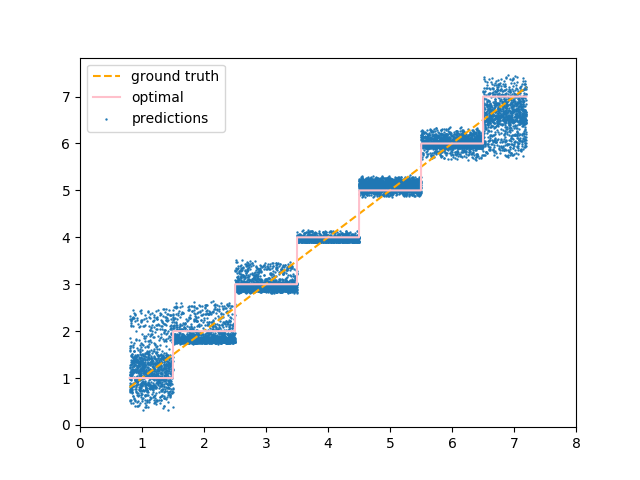}}
    \hfill\\
    \subfloat{
         \includegraphics[width=0.18\textwidth]{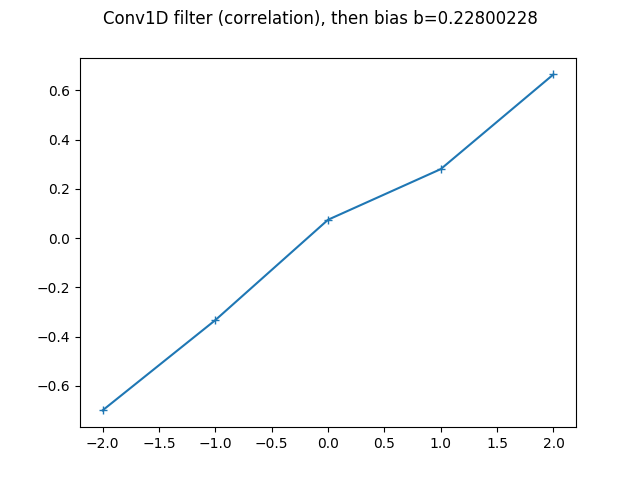}}
    \hfill
    \subfloat{
         \includegraphics[width=0.18\textwidth]{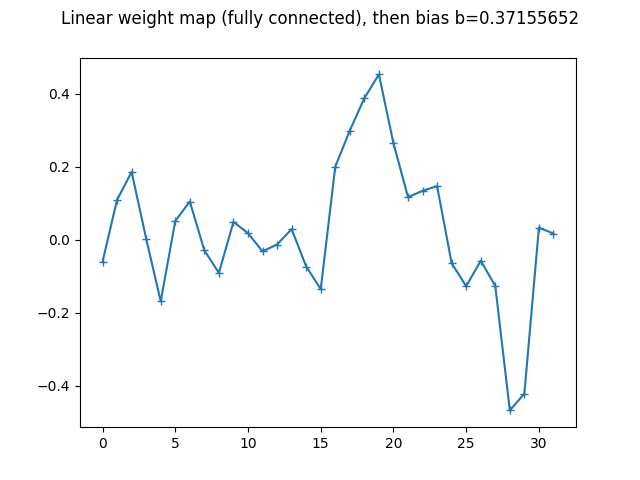}}
    \hfill
    \subfloat{
         \includegraphics[width=0.18\textwidth]{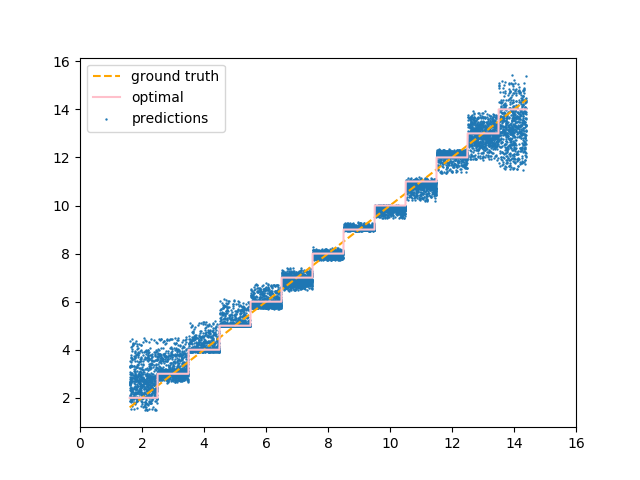}}
    \hfill\\
    \subfloat{
         \includegraphics[width=0.18\textwidth]{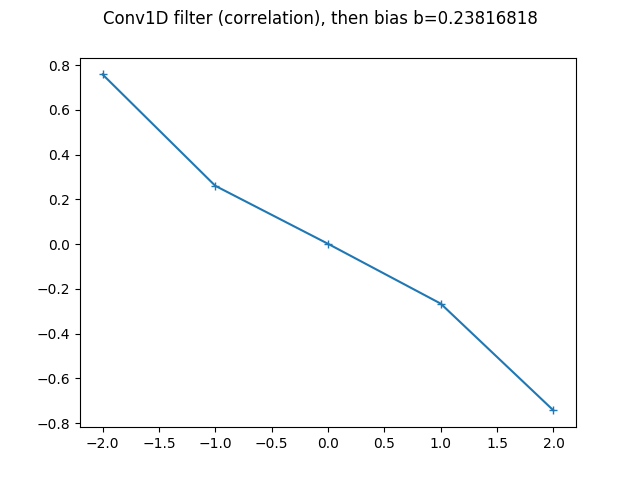}}
    \hfill
    \subfloat{
         \includegraphics[width=0.18\textwidth]{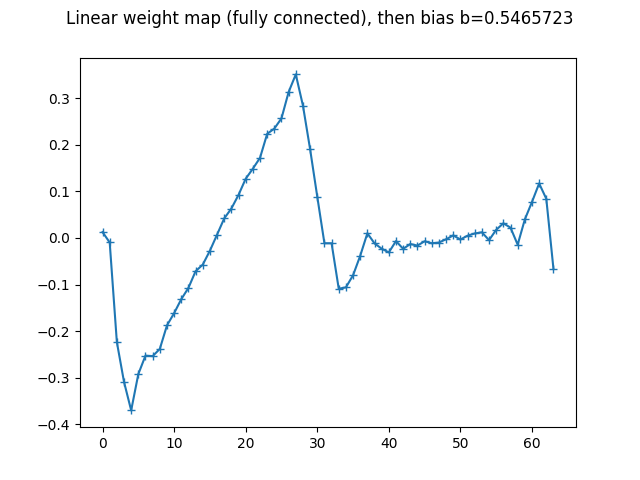}}
    \hfill
    \subfloat{
         \includegraphics[width=0.18\textwidth]{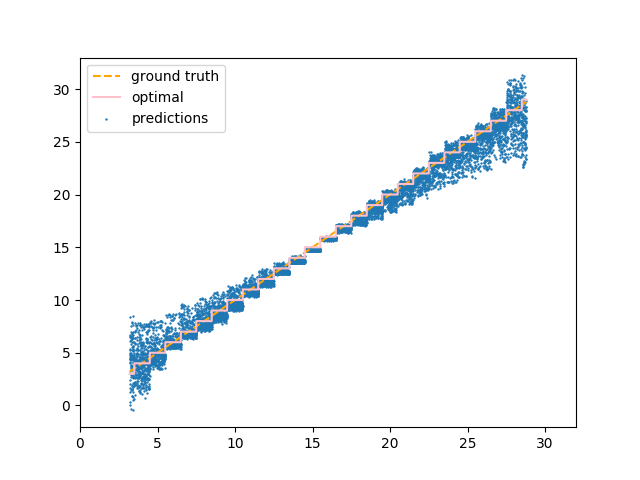}}
    \hfill\\
    \subfloat{
         \includegraphics[width=0.18\textwidth]{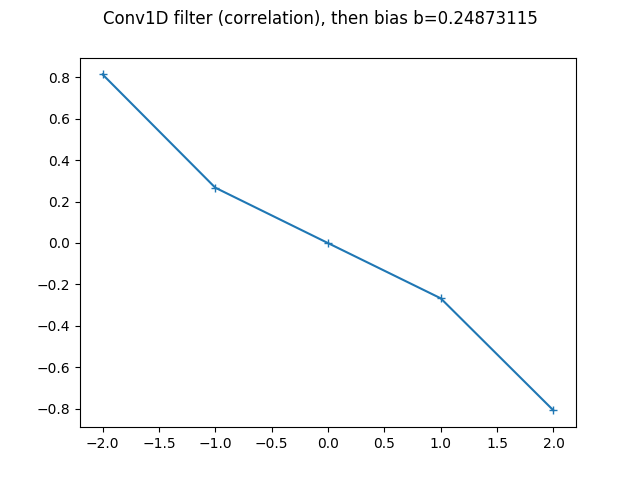}}
    \hfill
    \subfloat{
         \includegraphics[width=0.18\textwidth]{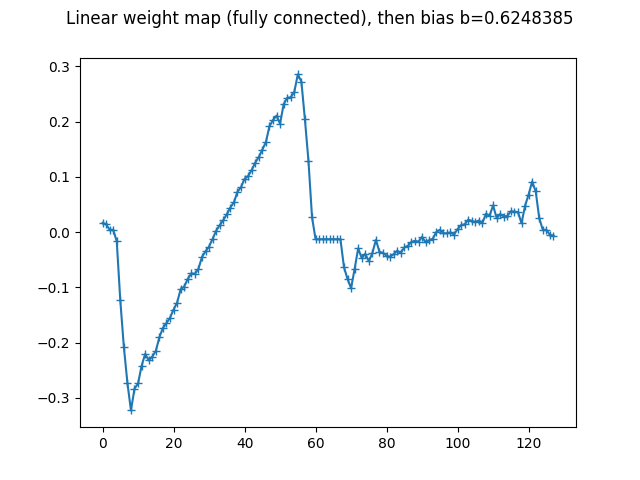}}
    \hfill
    \subfloat{
         \includegraphics[width=0.18\textwidth]{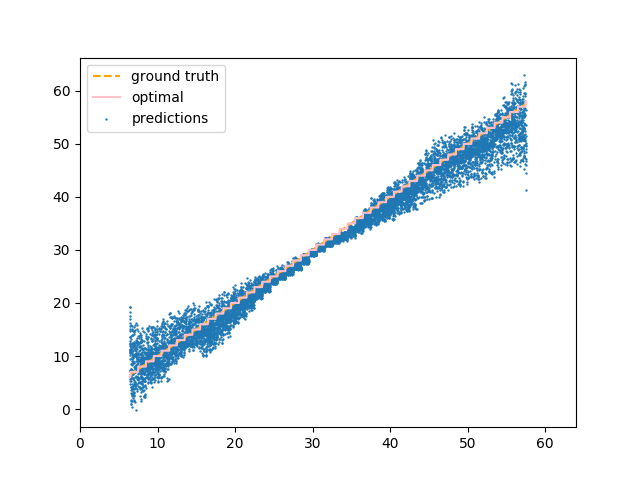}}
    \hfill\\
    \subfloat{
         \includegraphics[width=0.18\textwidth]{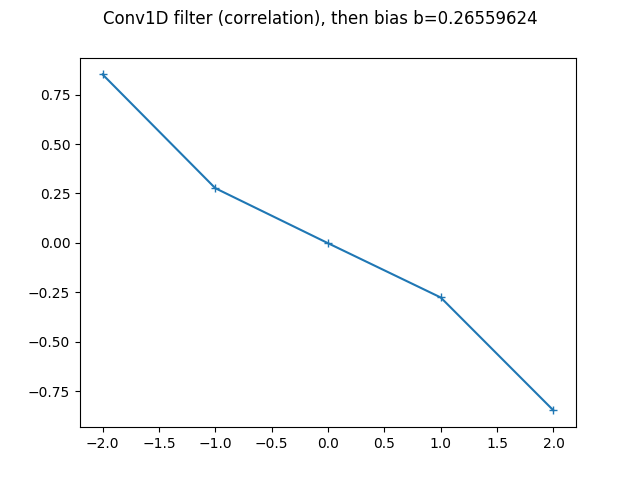}}
    \hfill
    \subfloat{
         \includegraphics[width=0.18\textwidth]{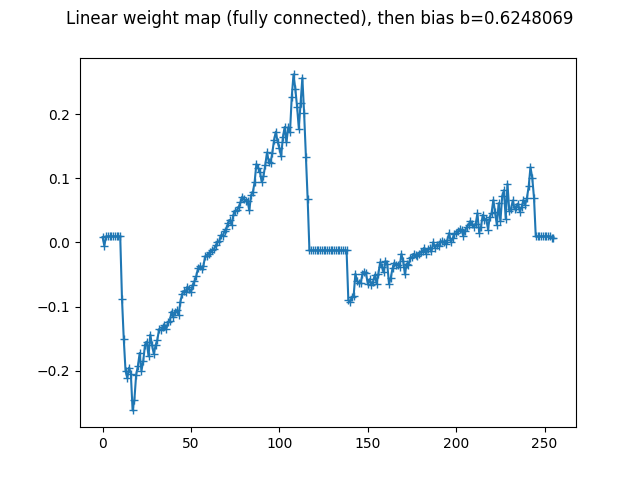}}
    \hfill
    \subfloat{
         \includegraphics[width=0.18\textwidth]{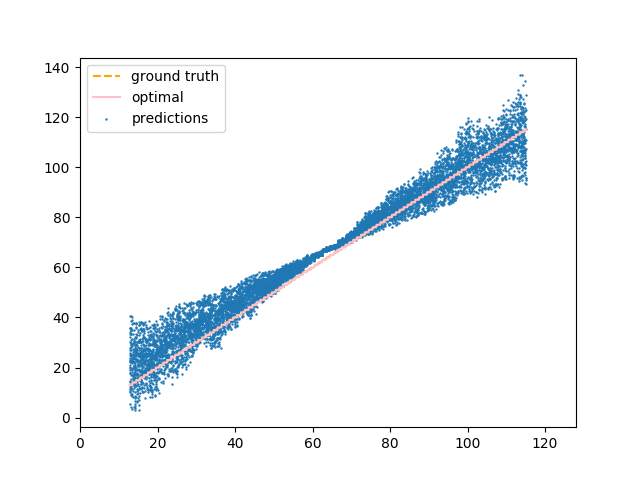}}
    \hfill\\
\caption[]{Learned $\relu$ networks and estimations when varying the resolution of the signals  $D=16$, $D=32$, $D=64$, $D=128$, and $D=256$ from top to bottom, but also keeping the filter's support to have $5$ entries only. The data was taken to be the clean non blurred version with $\sigma_B=\sigma_n=0$. Left: learned correlation filter. Middle: learned weight map for the fully connected layer. Right: estimations of the network on test datasets of $N=10000$ signals.}
\label{fig: step cnn relu evol D}
\end{figure}

We present in Figure \ref{fig: step cnn relu evol D} the results of training networks on positive polarity data with varying resolution $D\in\{13,32,64,128,256\}$ but without any noise or Gaussian blur i.e. $\sigma_g = \sigma_n = 0$, which is useful for comparing with the theoretical derivations in the continuum. Note that we fixed the pixel size of the convolution kernel to be $5$ which implies a shrinking of its support in $[0,1]$ with increasing resolution. This choice comes from the fact that we want our convolutions to be local operations. Indeed, keeping the pixel size fixed of the support increases the locality of the convolution operator with increasing $D$, which gets closer to our assumption in the theoretical case of a mathematically local operator in the continuum. Furthermore, increasing the size of the filter may create more artificial variability in it, allowing it to become an intricate and obscure oscillating kernel function. Such behaviour would unjustly harm the comparison with the reference context of Figure \ref{fig: MAIN step cnn relu weights standard context} where the kernel is limited by the $5$ entries.

\subsubsection{Opening the box: viewing intermediate representations}
\label{subsec: appendix 1d intermediate representations}

\begin{figure}[!ht]
    \centering
    \subfloat[$f_\theta$]{
         \includegraphics[width=0.15\textwidth]{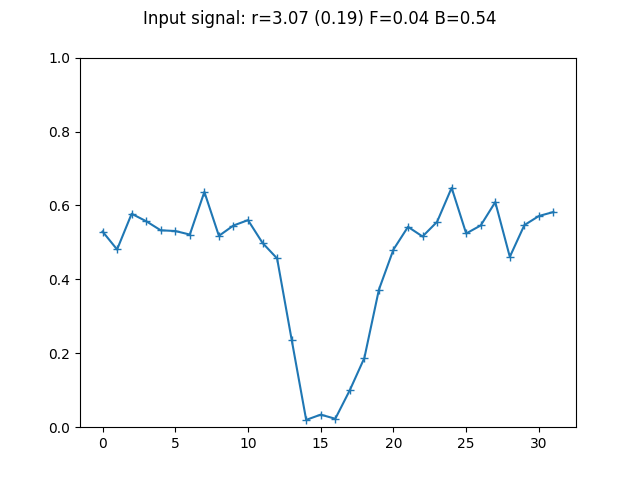}}
    \hfill
    \subfloat[$f_h$]{
         \includegraphics[width=0.2\textwidth]{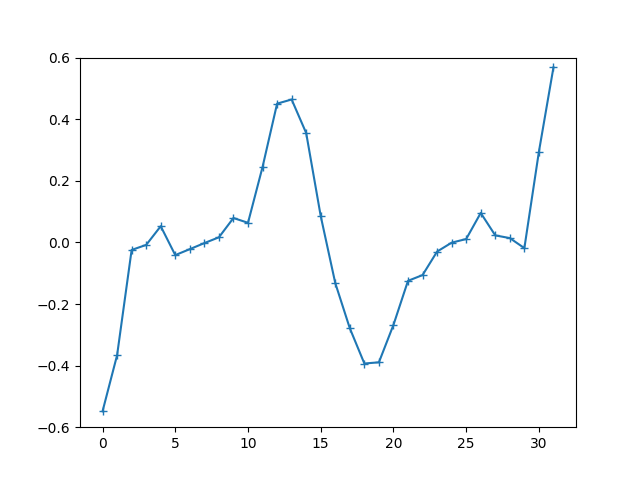}}
    \hfill
    \subfloat[$f_{hb}$]{
         \includegraphics[width=0.2\textwidth]{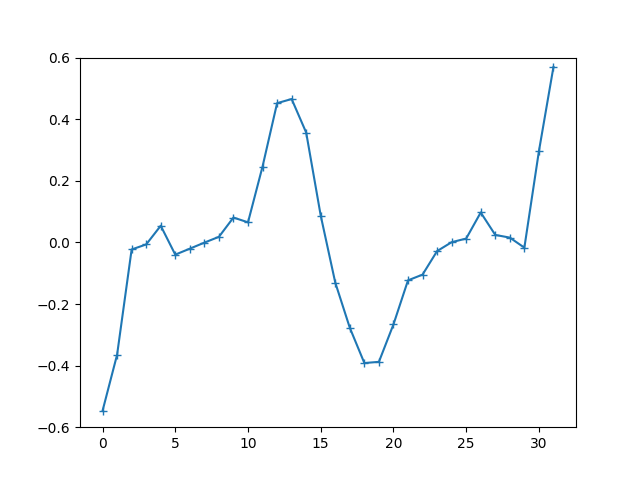}}
    \hfill
    \subfloat[$f_\sigma$]{
         \includegraphics[width=0.2\textwidth]{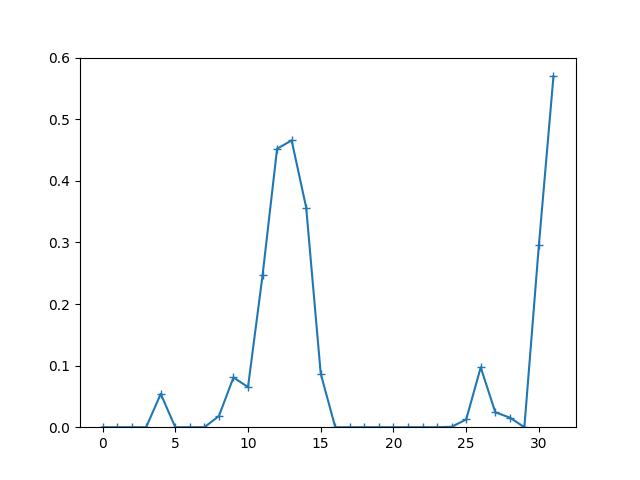}}
    \hfill
    \subfloat[$f_{a\odot}$]{
         \includegraphics[width=0.2\textwidth]{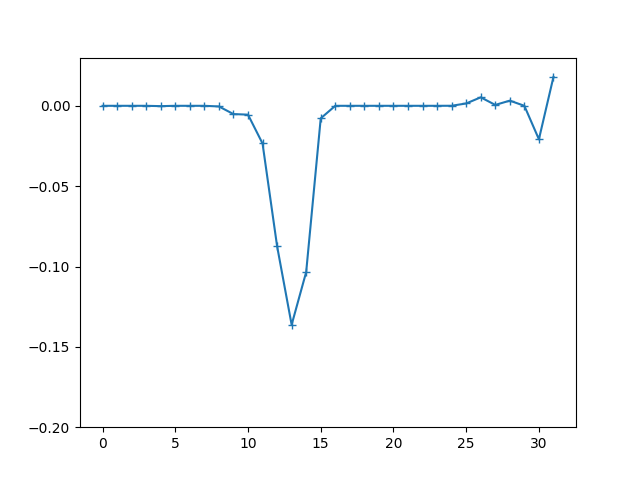}}
    \hfill\\
    \subfloat[$f_\theta$]{
         \includegraphics[width=0.15\textwidth]{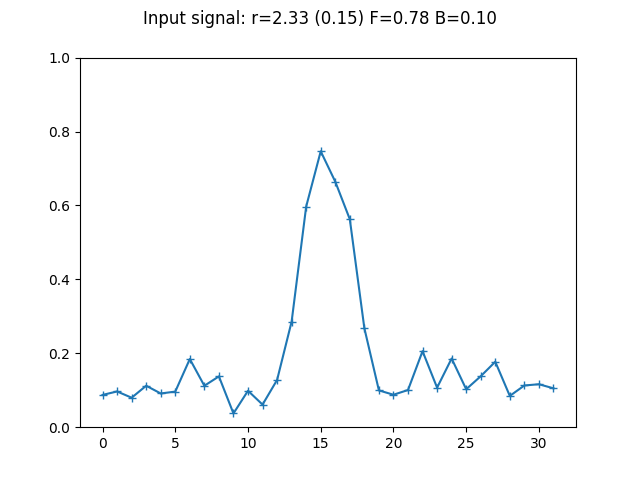}}
    \hfill
    \subfloat[$f_h$]{
         \includegraphics[width=0.2\textwidth]{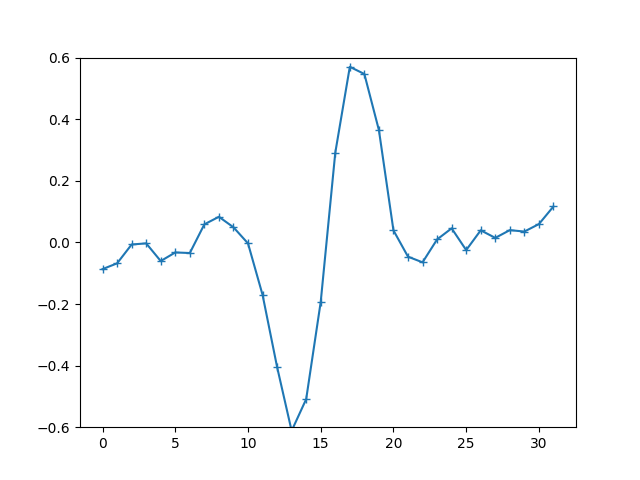}}
    \hfill
    \subfloat[$f_{hb}$]{
         \includegraphics[width=0.2\textwidth]{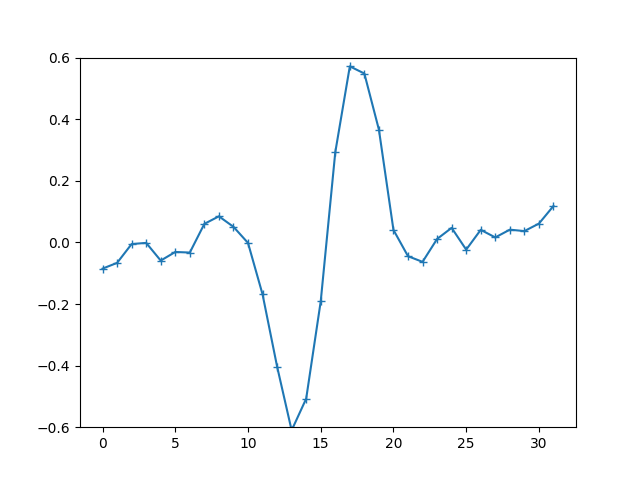}}
    \hfill
    \subfloat[$f_\sigma$]{
         \includegraphics[width=0.2\textwidth]{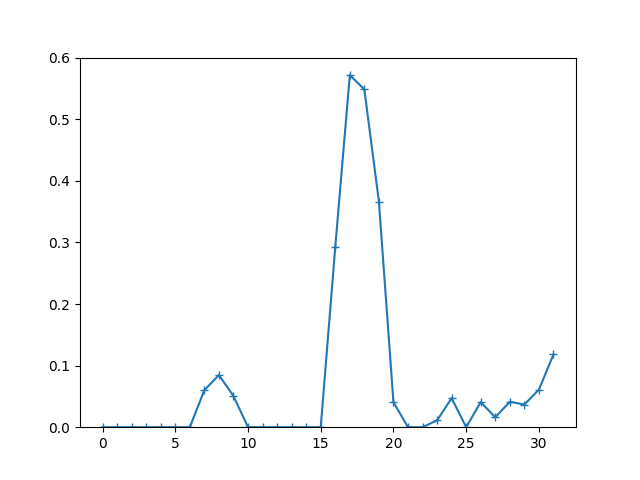}}
    \hfill
    \subfloat[$f_{a\odot}$]{
         \includegraphics[width=0.2\textwidth]{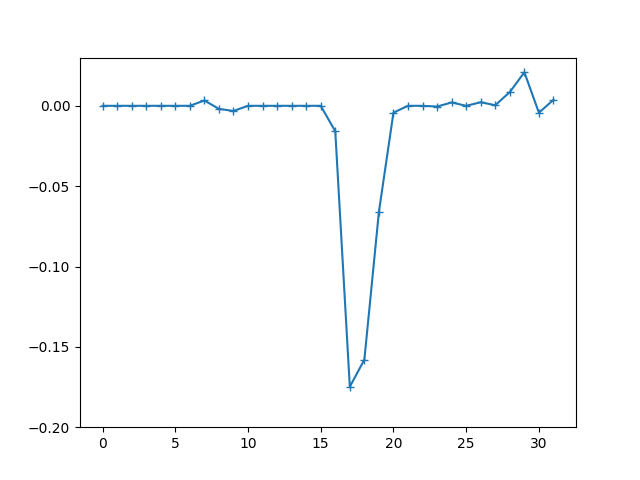}}
    \hfill
\caption[]{Computed features at each level of the neural network on a specific instance (left), with a $\relu$ activation network with $C=1$ convolution filter. The network was trained on both polarity data.}
\label{fig: step cnn relu features 1 conv kernel both pol}
\end{figure}

\begin{figure}[!ht]
    \centering
    \subfloat[$f_\theta$]{
         \includegraphics[width=0.15\textwidth]{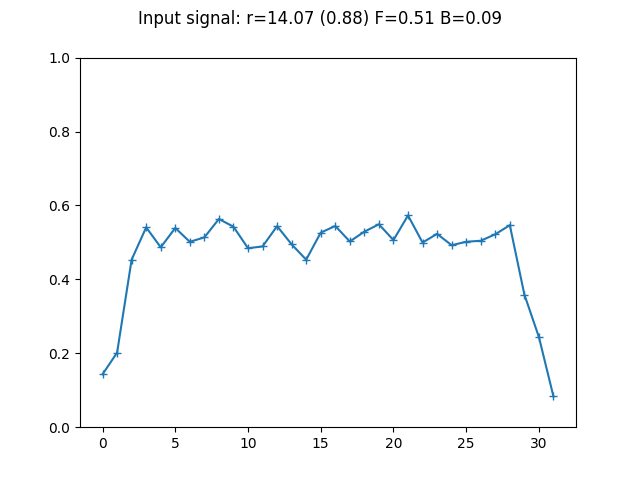}}
    \hfill
    \subfloat[$f_h$]{
         \includegraphics[width=0.2\textwidth]{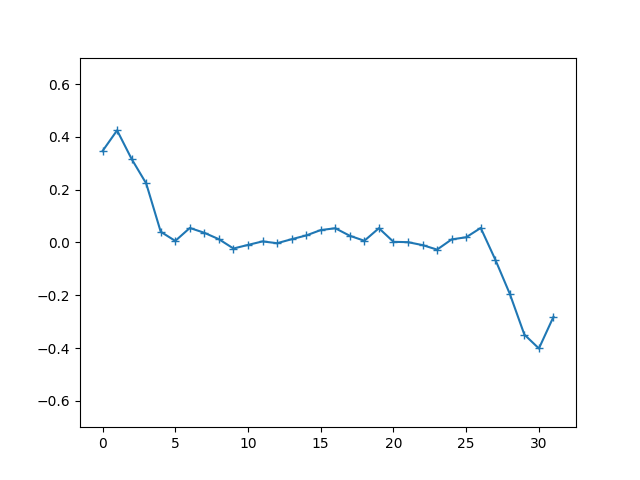}}
    \hfill
    \subfloat[$f_{hb}$]{
         \includegraphics[width=0.2\textwidth]{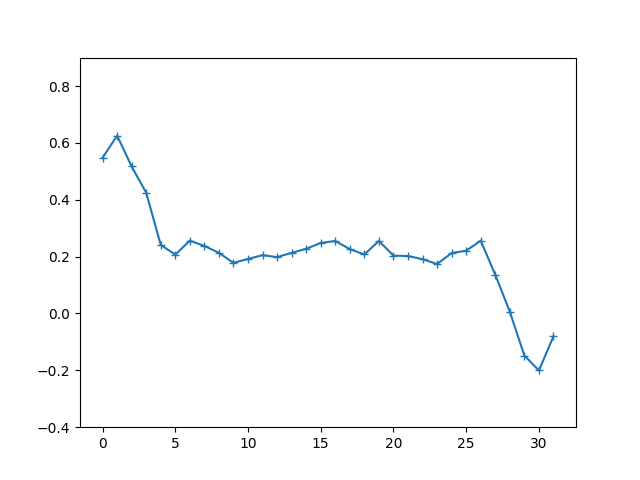}}
    \hfill
    \subfloat[$f_\sigma$]{
         \includegraphics[width=0.2\textwidth]{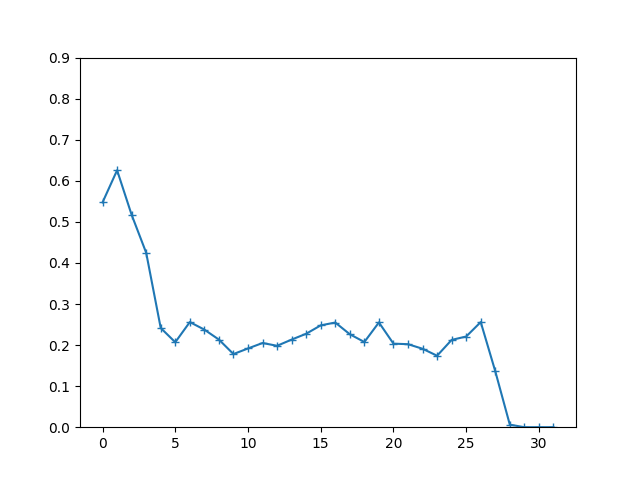}}
    \hfill
    \subfloat[$f_{a\odot}$]{
         \includegraphics[width=0.2\textwidth]{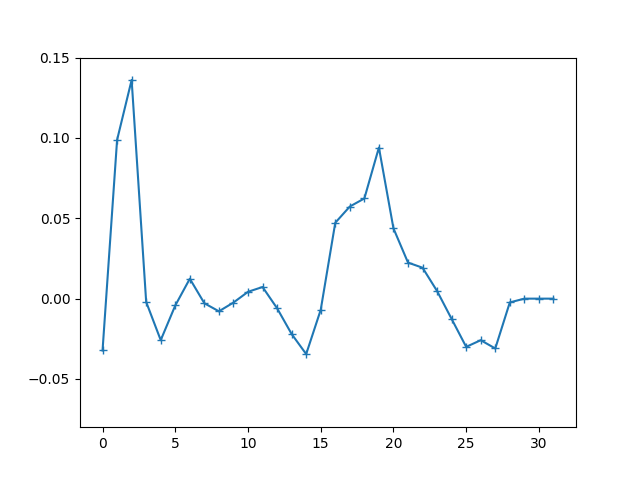}}
    \hfill\\
    \subfloat[$f_\theta$]{
         \includegraphics[width=0.15\textwidth]{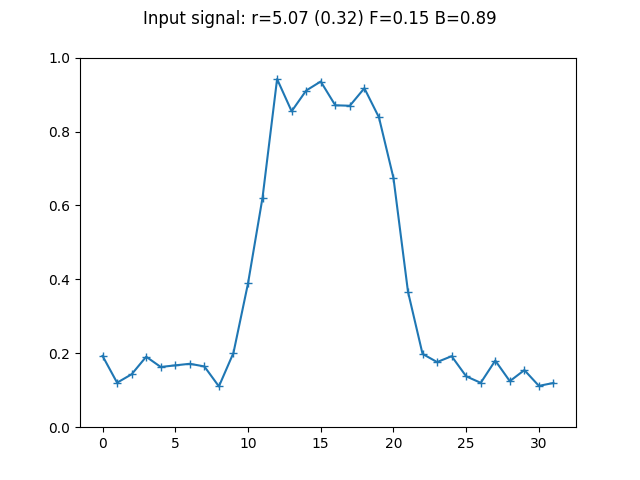}}
    \hfill
    \subfloat[$f_h$]{
         \includegraphics[width=0.2\textwidth]{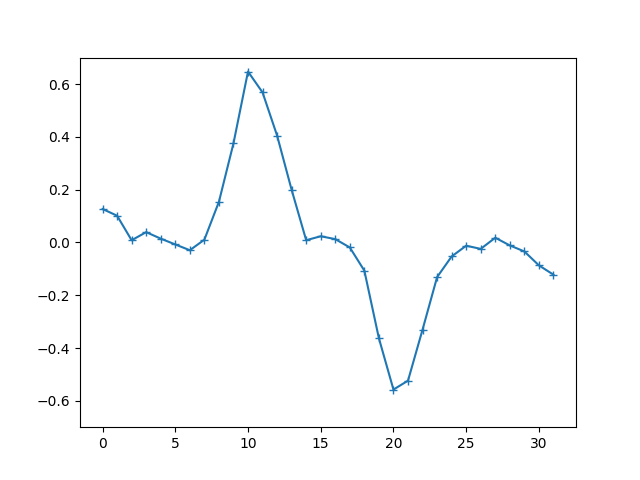}}
    \hfill
    \subfloat[$f_{hb}$]{
         \includegraphics[width=0.2\textwidth]{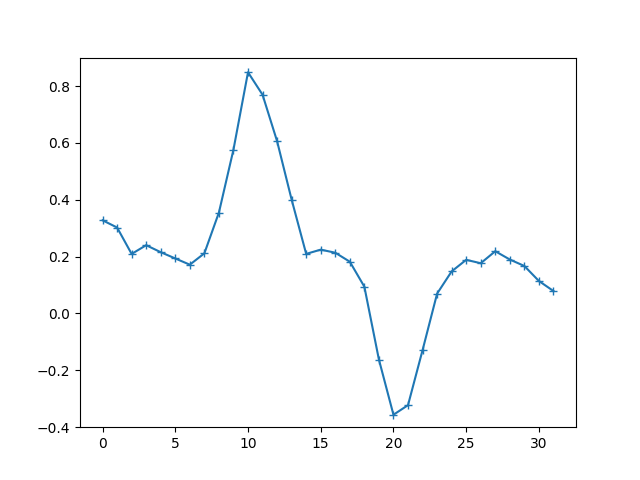}}
    \hfill
    \subfloat[$f_\sigma$]{
         \includegraphics[width=0.2\textwidth]{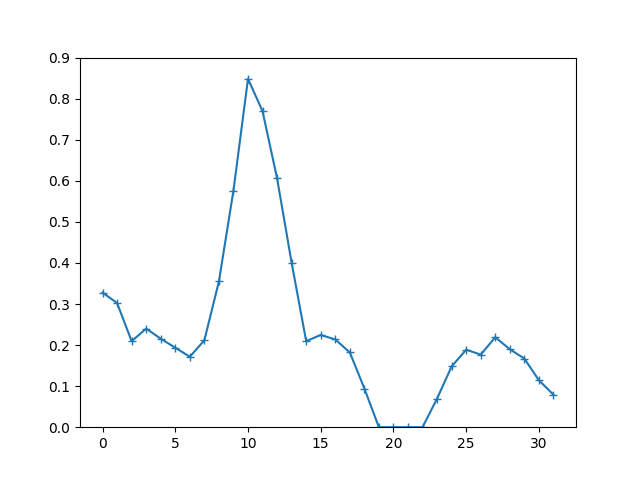}}
    \hfill
    \subfloat[$f_{a\odot}$]{
         \includegraphics[width=0.2\textwidth]{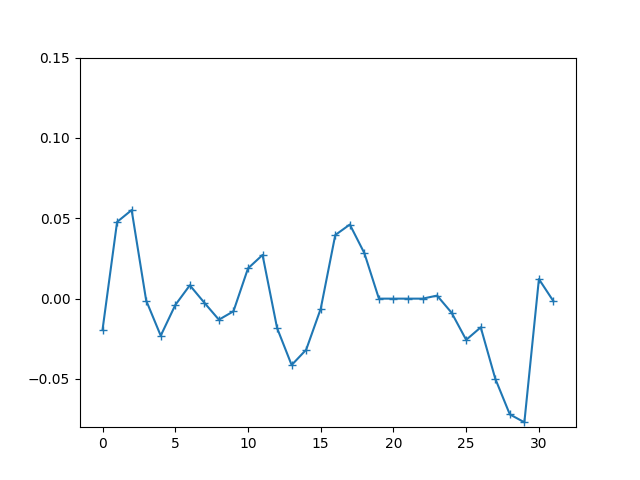}}
    \hfill
\caption[]{Computed features at each level of the neural network on a specific instance (left), with a $\relu$ activation network with $C=1$ convolution filter. The network was trained on positive-only polarity data only.}
\label{fig: step cnn relu features 1 conv kernel pos pol}
\end{figure}

Understanding can also arise from visual presentations. We thus also plot in Figures \ref{fig: step cnn relu features 1 conv kernel both pol} and \ref{fig: step cnn relu features 1 conv kernel pos pol} all the intermediate representations of random data fed to the networks. In the theoretical exercise, we do the same albeit with mathematical tools rather than direct visualisation.

\subsubsection{On the quality of the converged networks}
\label{subsec: appendix 1d on quality of converged networks}

Our quantitative performance indicator for the networks is the Root Means Squared Error (RMSE). In order to better understand this score, we scale the measurements to pixel size. This means that the outputs of the networks, radii estimates in $[0,\tfrac{1}{2}]$, are linearly scaled to $[0,\tfrac{D}{2}]$ by multiplying them by $\tfrac{D}{2}$. The RMSE is thus also scaled to pixel size. Recall that in the reference setting, which is the one studied here, $D=32$ and the noise level is given by $\sigma_n = \tfrac{10}{255}$. The network trained on positive polarity only data performs better with a score of $1.1$ compared to $1.7$ for the one trained on the dataset with both polarities present. Nevertheless, the RMSE in both cases is superior to a pixel which questions the quality of the estimators. 

\begin{figure}[!ht]
    \centering
    \subfloat[]{
         \includegraphics[width=0.4\textwidth]{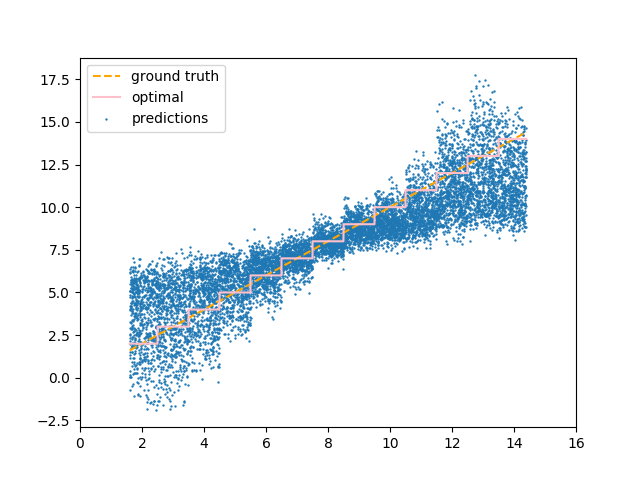}}
    \hfill
    \subfloat[]{
         \includegraphics[width=0.4\textwidth]{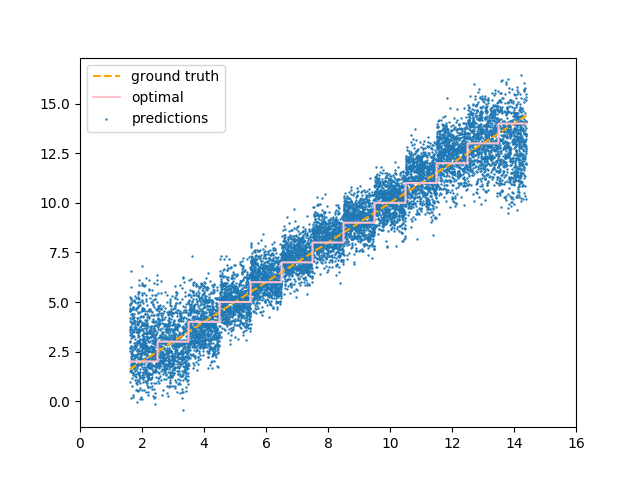}}
\caption[]{Performance of the $\relu$ activation neural networks using $C=1$ convolution filter. On the left (resp. right), the network is trained and tested on data with both (resp. positive-only) polarities. Each dot represents one of the $N=10000$ estimations on the test dataset.}
\label{fig: MAIN step relu estimations}
\end{figure}

To first understand the global behaviour of the estimators, rather than solely focusing on an average score, we look at actual estimations. In Figure \ref{fig: MAIN step relu estimations}, we plot the estimations of both networks on the test data. As suggested by the RMSE, both networks have estimations that significantly differ from the groundtruth label. However, it is clear that the standard deviation of the error is constant for any input radius for the network trained on the data with positive polarity only, unlike in the other network, which in many cases estimates the average dataset radius even for radii far from it. This behaviour suggests that the error in the former network might primarily be due to noise in the data, whereas in the latter it might be due to a failure to correctly grasp the concept of a radius.

 \begin{figure}[!ht]
    \centering
    \subfloat[]{
         \includegraphics[width=0.32\textwidth]{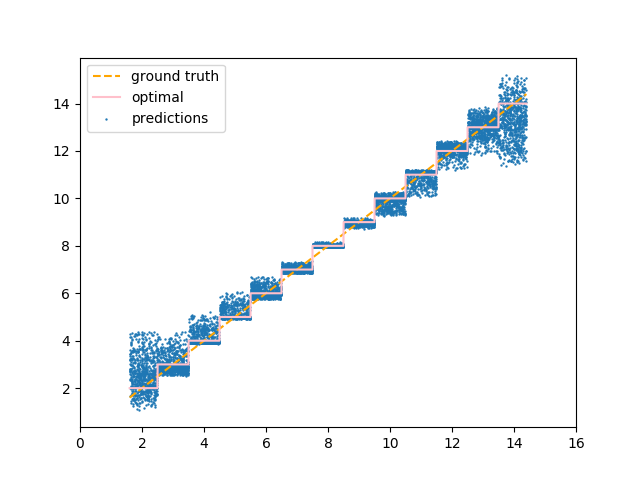}}
    \hfill
    \subfloat[]{
         \includegraphics[width=0.32\textwidth]{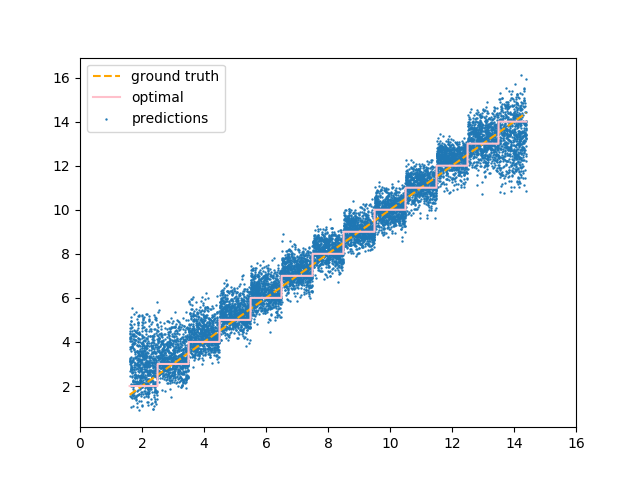}}
    \hfill
    \subfloat[]{
         \includegraphics[width=0.32\textwidth]{Pictures/step_relu_isFBright_True_crudeQuant_True_quantLabels_False_N_10000_D_32_preds_sorted_by_rgt.png}}
\caption[]{estimations on networks trained and tested with data with positive polarity only and noise level $\sigma_n = 0$, $\frac{5}{255}$, $\frac{10}{255}$ from left to right.}
\label{fig: MAIN step cnn relu evol sigma_N}
\end{figure}

This hypothesis is confirmed when analysing the evolution of the estimations with the noise level $\sigma_n$. If we decrease it, we get significant improvement for the network trained on positive polarity data only, unlike the other network. See Table \ref{tab: MAIN step rmse} for the RMSE scores in these cases. This confirms empirically that the network trained on a dataset with both polarities cannot accurately learn the radius concept, which is not the case of the other network. In Figure \ref{fig: MAIN step cnn relu evol sigma_N}, we plot the estimations of data with different noise levels and we can indeed see that the estimations fit better to the actual radius. In fact, we could add that the network also manages to handle quantisation of the radius with close to quantised estimations. Based on the theoretical analysis and understanding from the exemplar theoretical exercise where quantisation disappears, we speculate that the quantisation behaviour is partly due to the noise-like oscillating nature of the fully connected map.

\begin{table}[t]
    \centering
    \resizebox{0.8\textwidth}{!}{%
        \begin{tabular}{lccccccc}
            \toprule
                     & \multicolumn{7}{c}{Convolutional Neural Networks}\\ \cmidrule{2-8}
                     & \multicolumn{3}{c}{Both polarities} & & \multicolumn{3}{c}{Positive polarity}\\ \cmidrule{2-4} \cmidrule{6-8}
                     & $\sigma_n = 0$ & $\sigma_n = \tfrac{5}{255}$ & $\sigma_n = \tfrac{10}{255}$ & & $\sigma_n = 0$ & $\sigma_n = \tfrac{5}{255}$ & $\sigma_n = \tfrac{10}{255}$ \\
            \midrule                                     
            $RMSE$ & 1.50787 & 1.61251 & 1.72602 & & 0.54118 & 0.78567 & 1.07317 \\
            \bottomrule
        \end{tabular}
    }
    \vspace{10pt}
    \caption{Root Mean Squared Error performance for the radius estimation of the neural estimators using $\relu$ activation and $C=1$ convolution filter. The networks were trained and tested on separate datasets of $10000$ random signals each, where either both polarities or just the positive one are present, and having different noise levels $\sigma_n\in\{0,\frac{5}{255},\frac{10}{255}\}$. The RMSE is given in pixel unit and here the resolution is $D=32$.}
    \label{tab: MAIN step rmse}
\end{table}

\subsection{Neural engineer exercise}

\subsubsection{Exemplar exercise}

\paragraph{General unbiased differential convolution operators} 
\label{subsec: 1d extensions unbiased diff conv operators}

We here consider replacing the unbiased derivative convolution filter with a more general monomial differential operator, and without loss of generality give it unit gain. Such a convolution operator simply computes $\tfrac{d^k}{dx^k}f_\theta(x)$ at all locations $x$ of the domain on the input signal $f_\theta$ for some integer $k$. Intuitively, the associated convolution filter $h$ would correspond to a finite difference-like approximation of the derivative. With this understanding, we generalise the $0$-th derivative to any local averaging operation with $h$ of constant sign. Note that for $k\ge 1$, the monomial differential operators are unbiased as they zero out constant signals. Further note that $k$ can be interpreted from $h$ as the number of changes of sign in this filter. We state that for $k\neq 1$, the network cannot accurately tweak its biases and fully connected layer to accurately estimate the radius. We here provide intuitive and constructive arguments, but a tedious formal proof is lacking and left as an exercise.

Consider first the constant sign averaging filters, i.e. $k=0$. Since we naturally exclude the useless $h\equiv0$ filter, the outputs of $f_h$ will all be strictly positive or all strictly negative. As such, the preactivated entries $f_{hb}$ to be fed to the $\relu$ affinely depend on the intensities, implying that all non zeroed entries of the activated signals output of $\relu$ $f_\sigma$ affinely depend on the intensity difference with non zero slope. In turn, since the fully connected layer with its final bias is affine, the radius estimation affinely depends on the intensities in a non zeroed out fashion. This is in constrast to what happens when $k=1$ as in that case, a wise choice of $b_h$ provided some entries of the activated function to take values that are either $0$ or $b_h$, that are both independent of the intensities, in a predetermined portion of the domain.

As an example of the issue, consider a positive averaging filter $h$. Then $b_h$ must be negative otherwise the $\relu$ acts as the identity everywhere which implies a radial estimation that non trivially affinely depends on the intensities. Another failure case is to take $b_h$ negative with too large magnitude as then all preactivated signals are negative and the $\relu$ simply zeroes them out. Thus $b_h$ is negative with ``reasonable'' magnitude. In this hypothetical case, the foreground preactivated signal is non zero whereas the background one is zero, but the height of the non zero foreground affinely depends on the intensities without being constant, which implies that the final estimation also affinely and non trivially depends on the intensities.

Consider now the higher monomial differential operators with $k\ge 2$. Since these filters are unbiased, the signals $f_h$ will be entirely $0$ except at the locations of the border of the pulse $x=\tfrac{1}{2}\pm r$. At each of these locations, instead of a single upward or downward peak as in $k=1$, we will have $k$ peaks due to the $k$ changes of sign in $h$ with upwards and downwards peaks. The peaks do not need to have the same magnitude, but their magnitude all linearly depend on the intensity difference $|f-b|$. Since positive and negative peaks are present around both border locations, using a zeroing out strategy in the fully connected layer as in the case when $k=1$ is not viable. Indeed, this strategy needs to zero out upwards peaks, relying on the thresholding of the negative peaks by the $\relu$. But since positive peaks are present at both the left and right border of the pulse, zeroing them out would also zero all the valuable downwards peaks by locality of the filter\footnote{Schematically, all peaks at a border are squeezed together, thus applying a $0$ weight to that location zeroes out all peaks.}. This reasoning is valid under minor assumptions on  the fully connected layer $a$, such as a non oscillating behaviour on intervals of order of magnitude the support of the convolution filter.  As such, the network cannot estimate accurately what the radius is. 

Finally, note that the analysis for small $k$ is sufficient. Indeed, in the discrete world, the local filter only has a few entries. For instance, should we restrict ourselves to filters with $5$ entries, then there can be at most $2$ changes of sign in $h$ meaning that $k\le 4$ is sufficient.

\paragraph{Going deeper: two convolution layers} 
\label{subsec: 1d extensions going deeper}

We here increase the depth of the network to allow sequentially two convolutions (joined by a $\relu$ activation). We naturally extend our notations to this case by adding the indexing $1$ or $2$ depending on which layer the object corresponds to. Thus, the estimation is given by:
\begin{equation}
    \mathcal{R}(f_\theta) = a\sigma(h_2*\sigma(h_1*f_\theta + b_{h_1}) + b_{h_2}) + b_a = \left[\int_\Omega a_{i} \sigma(h_2*\sigma(h_1*f_\theta + b_{h_1}) + b_{h_2})_{i}\right] + b_a.
\end{equation}

As described in the main text, we choose $h_1$ and $b_{h_1}$ as in the case with only one convolution, namely that $h_1$ is an unbiased derivative filter and $b_{h_1}>0$ but small (as previously described), but also $h_2 = -\delta_0$ a $0$-centred negative Dirac delta function\footnote{In the discrete world, the 0-indicator function $\mathbbm{1}(0 = \cdot)$ would be analogous to the Dirac filter.} and $b_{h_2} = b_{h_1}$. For a signal $f_\theta$ with positive polarity we then have, under the same mild assumptions as previously:
\begin{align}
    \mathcal{R}(f_\theta) - b_a &= \int_{\Omega} \sigma(h_2 * \sigma(h_1 * f_\theta + b_{h_1}) + b_{h_2})(x) a(x)dx\\
    &\approx \int_{\Omega\setminus P_l\cup P_r} \sigma(-\sigma(b_{h_1}) + b_{h_1})a(x)dx \nonumber\\
    &\hspace{2em}+ \int_{P_l}\sigma(-\sigma(h_1*f_\theta + b_{h_1}) + b_{h_1})a(x)dx + \int_{P_r}\sigma(b_{h_1})a(x)dx \\
    &\approx \sigma(0)\int_{\Omega\setminus P_l\cup P_r} a(x)dx + \sigma(-h_1*f_\theta)_{|{\scriptstyle \frac{1}{2}} - r}\int_{P_l}a(x)dx + \sigma(b_{h_1})\int_{P_r}a(x)dx \\
    &\approx b_{h_1}\int_{P_r}a(x)dx  \approx b_{h_1}\cdot 2\Delta \cdot a({\scriptstyle \frac{1}{2}}\!+\!r),
\end{align}
because $h_1*f_\theta\ge0$ on $P_l$ for $f_\theta$ with positive polarity.

On the other hand, if $f_\theta$ has negative polarity, we then have for the same network:
\begin{align}
    \mathcal{R}(f_\theta) - b_a &= \int_{\Omega} \sigma(h_2 * \sigma(h_1 * f_\theta + b_{h_1}) + b_{h_2})(x) a(x)dx\\
    &\approx \int_{\Omega\setminus P_l\cup P_r} \sigma(-\sigma(b_{h_1}) + b_{h_1})a(x)dx \nonumber\\
    &\hspace{2em}+ \int_{P_l}\sigma(b_{h_1})a(x)dx + \int_{P_r}\sigma(-\sigma(h_1*f_\theta + b_{h_1}) + b_{h_1})a(x)dx \\
    &\approx \sigma(0)\int_{\Omega\setminus P_l\cup P_r} a(x)dx + \sigma(b_{h_1})\int_{P_l}a(x)dx + \sigma(-h_1*f_\theta)_{|{\scriptstyle \frac{1}{2}} + r}\int_{P_r}a(x)dx  \\
    &\approx b_{h_1}\int_{P_l}a(x)dx  \approx b_{h_1}\cdot 2\Delta \cdot a({\scriptstyle \frac{1}{2}}\!-\!r),
\end{align}
because $h_1*f_\theta\ge0$ on $P_r$ for $f_\theta$ with negative polarity.

To get the same estimation for both polarities, we must here choose $a$ to be symmetric around the midpoint of the domain $\tfrac{1}{2}$. To correctly estimate the radius with $\mathcal{R}(f_\theta)$, we must then choose $a$ to be a symmetric $V$-shaped piecewise linear function:
\begin{equation}
    a({\scriptstyle \frac{1}{2}}\!+\!r) = a({\scriptstyle \frac{1}{2}}\!-\!r) = \frac{r}{2\Delta b_{h_1}} - \frac{b_A}{2\Delta b_{h_1}}.
\end{equation}

We have thus found that increasing the depth of the network allows to find a solution that can become invariant to the intensities and correctly estimate the radius regardless of the polarity. Note that it may not be unique.

\subsubsection{Modified exercise: from $\relu$ to $\sigmoid$} 
\label{subsec: 1d sigmoid}

The discussion we provide here is similar to that in the $\relu$ case. Working in the continuum, we once again choose $h$ to be an unbiased derivative filter of small support size $\Delta$ and gain $\alpha$ to be used in the single channel and unique convolution layer. Without loss of generality by symmetry of the pulse functions, convolving a monotonically increasing function with $h$ provides positive outputs only\footnote{Thus $h(-\tfrac{\Delta}{2})<0$ and $h(\tfrac{\Delta}{2})>0$ for the correlation filter $h$.}. 

Let $f_\theta$ be a centred pulse signal with arbitrary polarity. When fed to the convolution filter, we get as previously described a function equal to $0$ everywhere except around the borders of the pulse on two $2\Delta$-wide peak intervals $P_l$ and $P_r$, where there is a positive peak on the left ($P_l$) and a negative one on the right ($P_r$) of magnitude $|\alpha\tfrac{f-b}{2}|$ if $f_\theta$ has positive polarity or the opposite otherwise. As we have naturally rediscovered in the $\relu$ case, it is essential to obtain a final estimation that is invariant to the intensity difference. If we do not push the peaks to lie in the flat regions of the $\sigmoid$, where approximate thresholding is performed, then the estimation will inevitably be approximately affine in the intensity difference, and the only way to get an invariant estimation would be to estimate a constant number that does not vary with the radius, which is unacceptable. To do so, we must\footnote{If we restrain ourselves to fixed polarity, e.g. positive-only, thanks to the minimal contrast $\delta$ it is also possible to have a non large gain $\alpha$ but instead choose the bias to have large magnitude $|b_h|\in (\tau - \delta\tfrac{\alpha}{2}<\tau)$ without being too large as to push the $0$ background level beyond the $\tau$ level of the flat regime and thus loosing all information. Unfortunately, this solution does not scale to the case when the dataset has data with arbitrary polarity.} choose a high gain $\alpha$ that will provide $\alpha|\tfrac{f-b}{2}|+b_h>\tau$. Contrary to intuition, choosing $b_h = 0$ and $\alpha> \tau\frac{2}{\delta} \ge\tau |\tfrac{2}{f-b}|$ does not provide a correct estimator although the activated function is non trivially invariant to the intensities everywhere in $[0,1]$. Such a choice would lead to representations $f_\sigma$ that are constant $0$ functions except on the borders of the pulse where there are now upward and downward peaks of same constant magnitude, the upward one being on the left for positive polarity data. Thus the representations are opposites for positive and negative polarity data of same radius. By linearity of the fully connected operator $a$, we have $\mathcal{R}(f_{(r,f,b)}) - b_a = -(\mathcal{R}(f_{r,b,f}) - b_a)$, implying that $\mathcal{R}(f_{(r,f,b)}) = b_a$ is a constant that does not depend on the radius, which is unacceptable\footnote{This reasoning can be generalised to affine transforms. If an affine transform of the input (such as a polarity flip) always leads to the another yet constant affine transform of the activated representations, then the fully connected layer will have no other choice but to remove all the information on the signal and just provide a final estimation that does not depend on the radius.}. We must therefore break the symmetry with $b_h\neq 0$. A solution is to choose a bias with large magnitude, e.g. $b_h<-\tau<0$, to push the background $0$ level to the thresholding regimes of the $\sigmoid$. Without loss of generality, we thus choose:
\begin{align}
    b_h &< -\tau \\
    \alpha &> 2\frac{\tau - b_h}{\delta}>\frac{4\tau}{\delta}.
\end{align}

Thus, the output of the $\sigmoid$ $f_\sigma$ will now always consist in an approximately constant $0$ function, except around one border of the pulse, on $P_l$ in the left half of the domain for positive polarity data and on $P_r$ in the right half of the domain for negative polarity data, where there is there an increasing constant height thresholded peak. The representation is thus non trivially invariant to intensities and closely resembles those obtained in the $\relu$ case when we allowed the network to use two convolution layers. We already saw then that the radius can be accurately recovered by designing a symmetric V-shaped weight map $a$.

More formally, denoting $(P_+, P_-) = (P_l, P_r)$ if $f_\theta$ has positive polarity and $(P_+, P_-) = (P_r, P_l)$ otherwise, we have under our usual assumptions:
\begin{align}
    \mathcal{R}(f_\theta) - b_a &= \int_{\Omega} \sigma(h*f_\theta + b_h)(x)a(x)dx\\
    &\approx \sigma(b_h) \int_{\Omega\setminus (P_+ \cup P_-)} a(x)dx + \int_{P_+} (\sigma(h*f_\theta + b_h)(x)) a(x)dx \nonumber\\
    &\hspace{2em}+ \int_{P_-} (\sigma(h*f_\theta + b_h)(x)) a(x)dx \\
    &\approx \sigma(b_h) \int_{\Omega} a(x)dx + \int_{P_+} \left(\sigma\left(\alpha \frac{f-b}{2} +b_h\right) - \sigma(b_h)\right) a(x)dx \nonumber\\
    &\hspace{2em}+ \int_{P_-} \left(\sigma\left(-\alpha \frac{f-b}{2} +b_h\right) - \sigma(b_h)\right) a(x)dx \\
    &\approx 0 \int_{\Omega} a(x)dx + \int_{P_+} (1 - 0) a(x)dx + \int_{P_-} (0-0) a(x)dx \\
    &\approx \int_{P_+} a(x)dx \approx 
        \begin{cases}
            2\Delta\cdot a\left(\frac{1}{2}-r\right) &\text{if } f>b\\
            2\Delta\cdot a\left(\frac{1}{2}+r\right) &\text{if } f<b\\
        \end{cases}.
\end{align}

To get the correct estimation $\mathcal{R}(f_\theta) = r$, we must thus design $a$ to be a symmetric V-shaped weight map:
\begin{equation}
    a\left(\frac{1}{2} \pm r\right) = \frac{r}{2\Delta} - \frac{b_a}{2\Delta}.
\end{equation}

Note that should we consider $L^2$ regularisation of the weights of $a$, then the optimal final bias would symmetrise each affine half of $a$ at height $0$, i.e. $a(\tfrac{1}{4}) = a(\tfrac{3}{4}) = 0$, which occurs for a choice of $b_a = \tfrac{1}{4}$ equal to the average radius in the dataset.

\section{Details for two-dimensional circle images exercise}

\subsection{Network details}

\subsubsection{Training procedure}
\label{subsec: appendix circle training procedure}

Training was done using the SGD optimiser with learning rate of $\eta=0.001$ (with small less important decay) and batch size $32$. We add $L^2$ regularisation penalties with small Lagrangian coefficients for the convolution and fully connected layers but not for the biases. Training is done using $N=10000$ training samples. The validation and tests sets also have the same size. Since the networks are fairly small, convergence happens rather fast, but we keep on training for a long duration to make sure that no strange phenomenon occurs.

\subsubsection{Opening the box: viewing intermediate representations}
\label{subsec: appendix circle intermediate representations}

\begin{figure}[!ht]
    \centering
    \subfloat[$f_\theta$]{
         \includegraphics[width=0.15\textwidth]{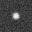}}
    \hfill
    \subfloat[$f_h$]{
         \includegraphics[width=0.2\textwidth]{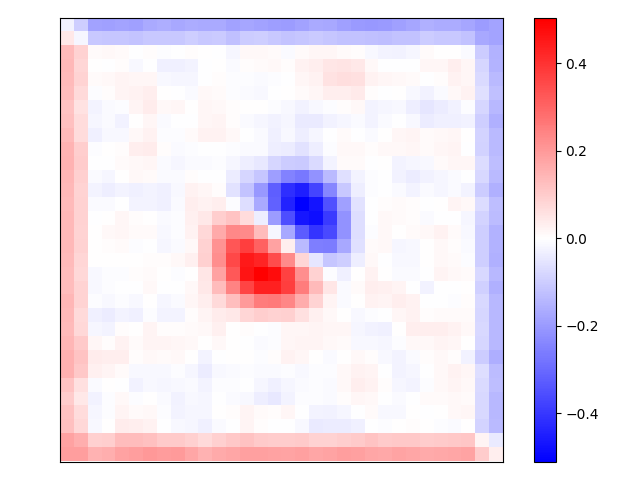}}
    \hfill
    \subfloat[$f_{hb}$]{
         \includegraphics[width=0.2\textwidth]{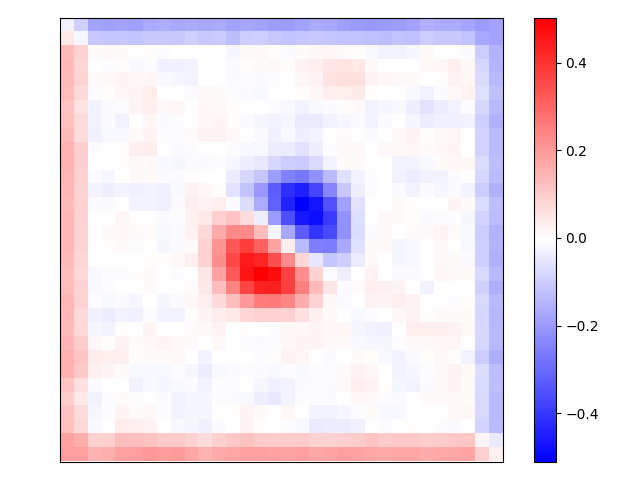}}
    \hfill
    \subfloat[$f_\sigma$]{
         \includegraphics[width=0.2\textwidth]{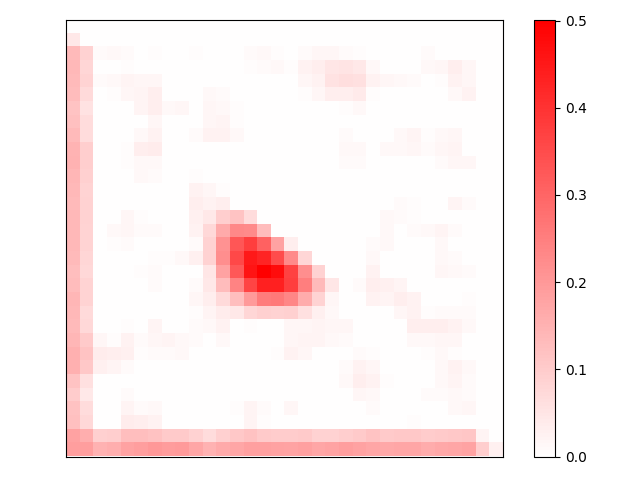}}
    \hfill
    \subfloat[$f_{a\odot}$]{
         \includegraphics[width=0.2\textwidth]{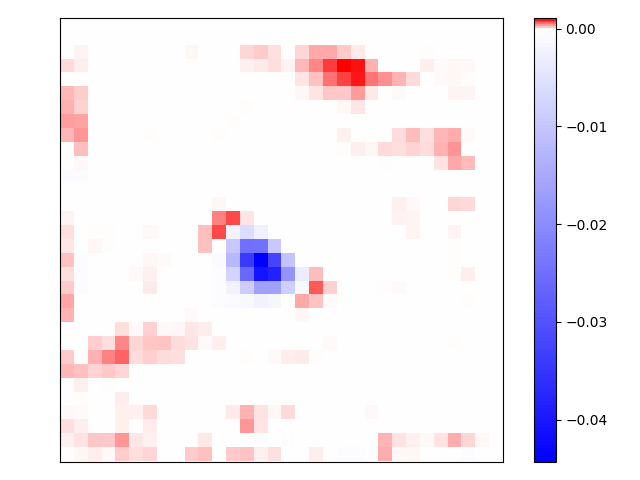}}
    \hfill\\
    \subfloat[$f_\theta$]{
         \includegraphics[width=0.15\textwidth]{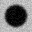}}
    \hfill
    \subfloat[$f_h$]{
         \includegraphics[width=0.2\textwidth]{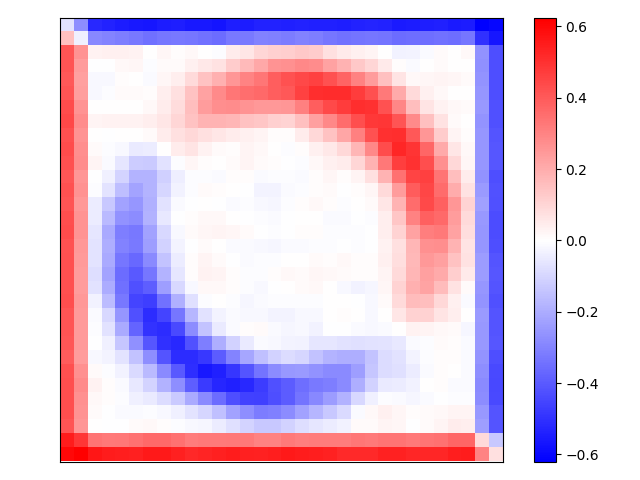}}
    \hfill
    \subfloat[$f_{hb}$]{
         \includegraphics[width=0.2\textwidth]{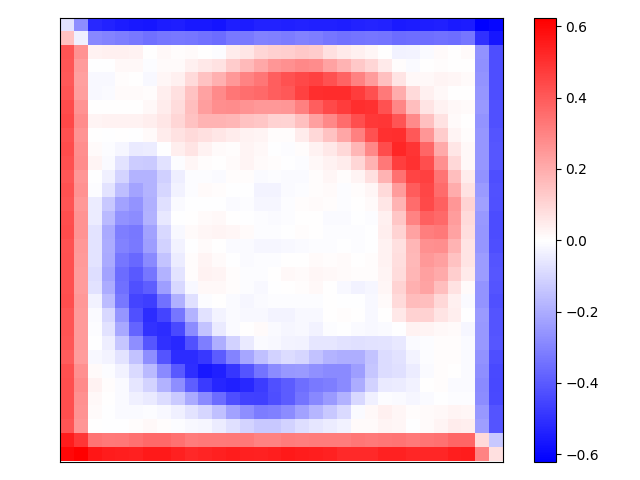}}
    \hfill
    \subfloat[$f_\sigma$]{
         \includegraphics[width=0.2\textwidth]{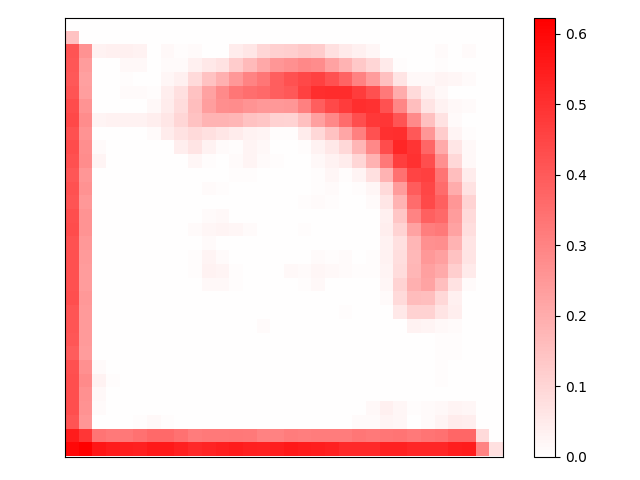}}
    \hfill
    \subfloat[$f_{a\odot}$]{
         \includegraphics[width=0.2\textwidth]{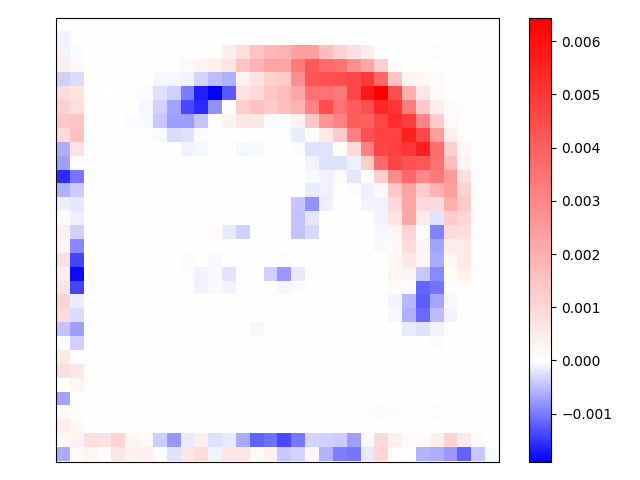}}
    \hfill
\caption[]{Computed features at each level of the neural network on a specific instance (left), with a $\relu$ activation network with $C=1$ convolution filter. The network was trained on data with both polarities.}
\label{fig: circle cnn relu features 1 conv kernel}
\end{figure}

\begin{figure}[!htbp]
    \centering
    \subfloat[$f_\theta$]{
         \includegraphics[width=0.08\textwidth]{Pictures/deg_im_2.png}}
    \subfloat[$f_h$]{
         \includegraphics[width=0.23\textwidth]{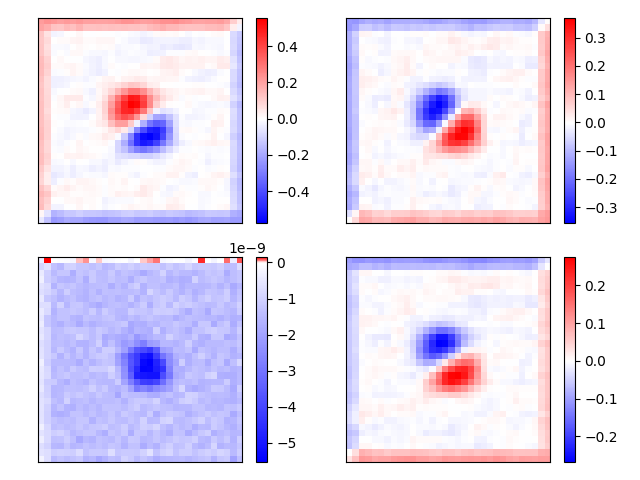}}
    \subfloat[$f_{hb}$]{
         \includegraphics[width=0.23\textwidth]{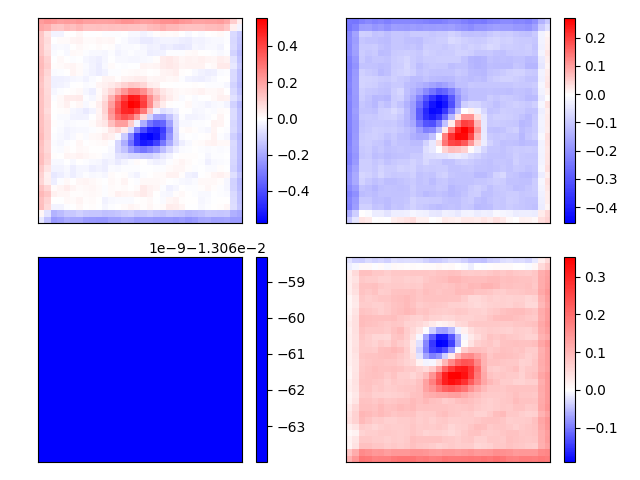}}
    \hfill\\
    \hspace{3.7em}
    \subfloat[$f_\sigma$]{
         \includegraphics[width=0.23\textwidth]{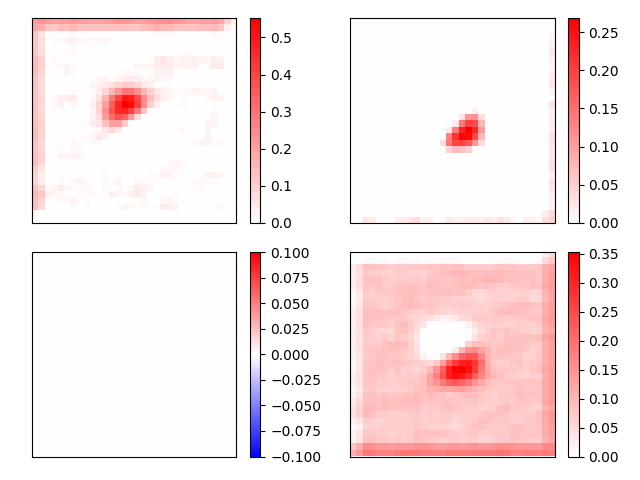}}
    \subfloat[$f_{a\odot}$]{
         \includegraphics[width=0.23\textwidth]{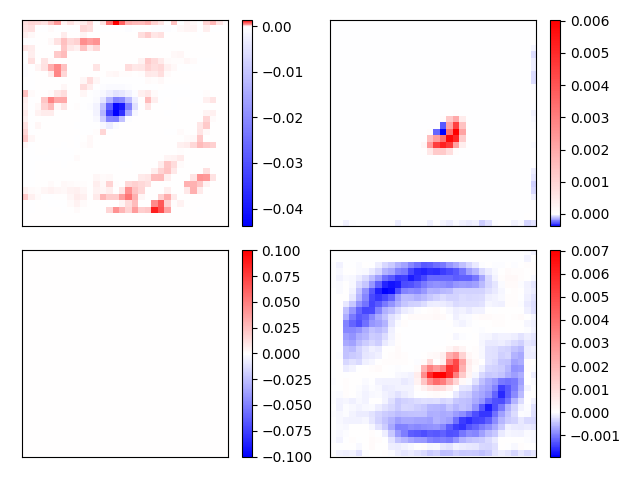}}
    \hfill\\
    \subfloat[$f_\theta$]{
         \includegraphics[width=0.08\textwidth]{Pictures/deg_im_9.png}}
    \subfloat[$f_h$]{
         \includegraphics[width=0.23\textwidth]{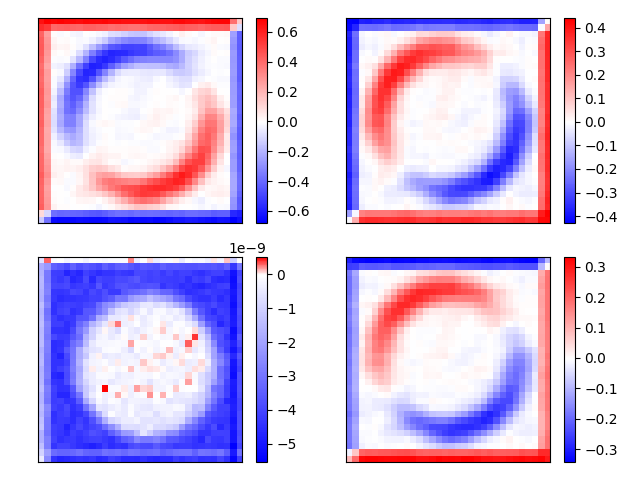}}
    \subfloat[$f_{hb}$]{
         \includegraphics[width=0.23\textwidth]{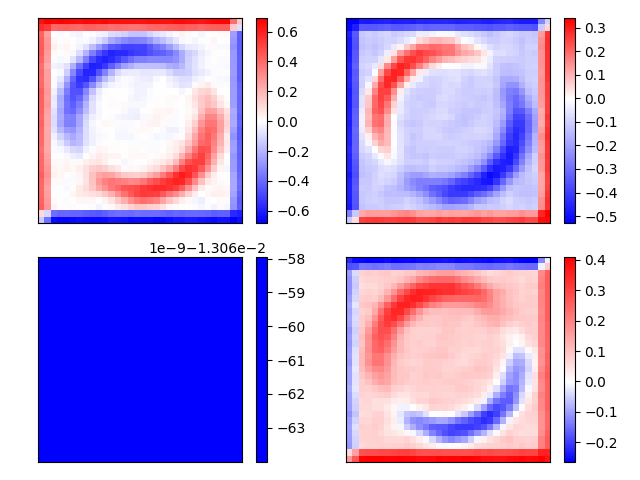}}
    \hfill\\
    \hspace{3.7em}
    \subfloat[$f_\sigma$]{
         \includegraphics[width=0.23\textwidth]{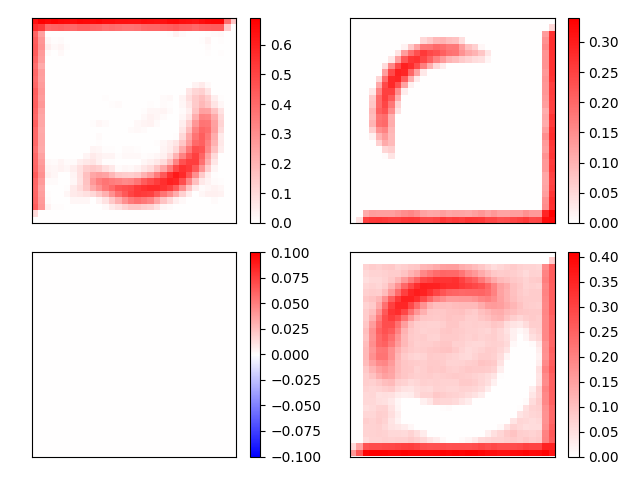}}
    \subfloat[$f_{a\odot}$]{
         \includegraphics[width=0.23\textwidth]{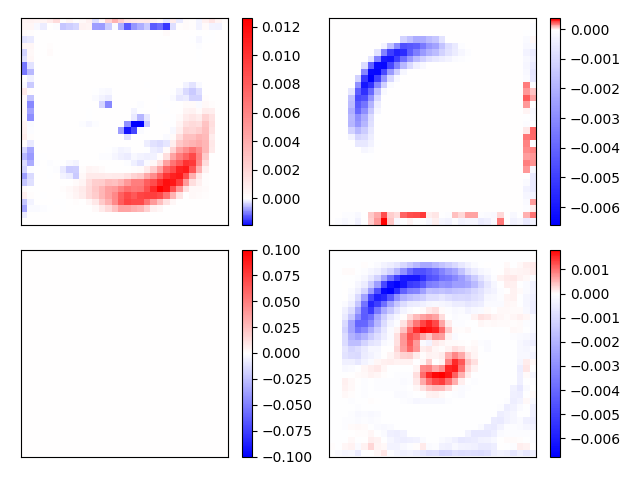}}
    \hfill
\caption[]{Computed features at each level of the neural network on a specific instance (left), with a $\relu$ activation network with $C=4$ convolution filters. The network was trained on data with both polarities.}
\label{fig: circle cnn relu features 4 conv kernel}
\end{figure}

\begin{figure}[!htbp]
    \centering
    \subfloat[$f_\theta$]{
         \includegraphics[width=0.1\textwidth]{Pictures/deg_im_2.png}}
    \hfill\\
    \subfloat[$f_h$]{
         \includegraphics[width=0.45\textwidth]{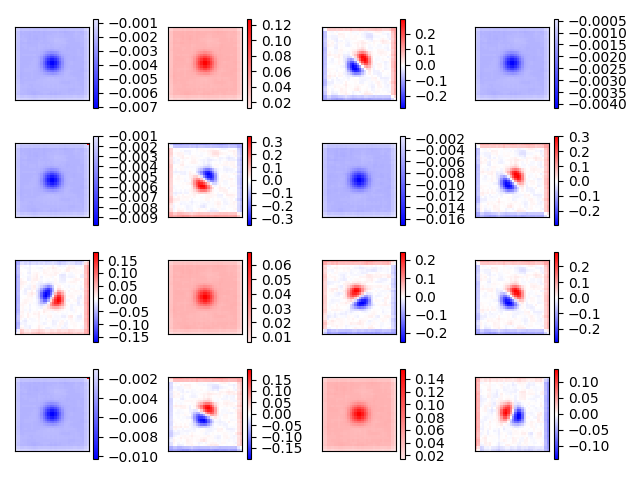}}
    \hfill
    \subfloat[$f_{hb}$]{
         \includegraphics[width=0.45\textwidth]{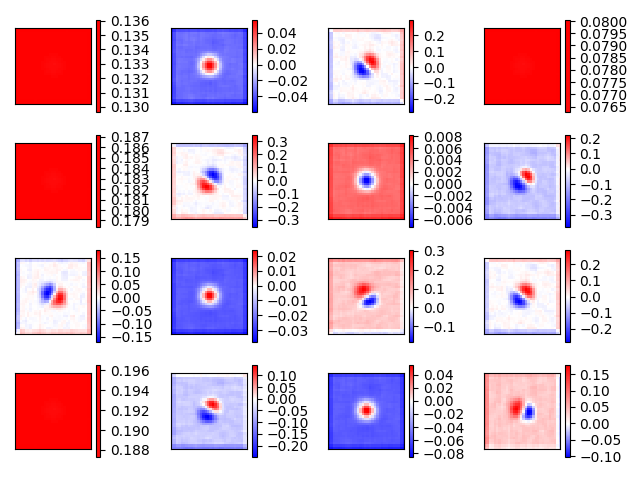}}
    \hfill\\
    \subfloat[$f_\sigma$]{
         \includegraphics[width=0.45\textwidth]{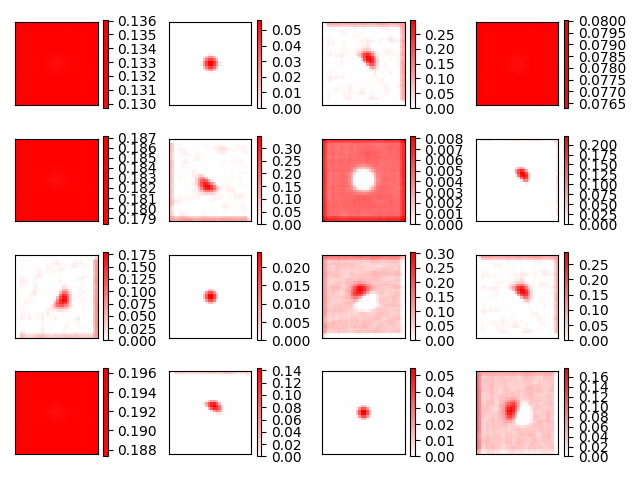}}
    \hfill
    \subfloat[$f_{a\odot}$]{
         \includegraphics[width=0.45\textwidth]{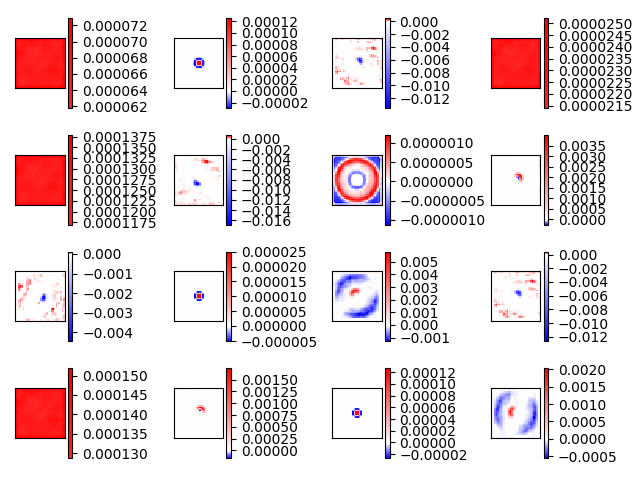}}
    \hfill\\
\caption[]{Computed features at each level of the neural network on the first example specific image, with a $\relu$ activation network with $16$ convolution filters. The network was trained on data with both polarities.}
\label{fig: circle cnn relu features 16 conv kernel image 2}
\end{figure}

\begin{figure}[!htbp]
    \centering
    \subfloat[$f_\theta$]{
         \includegraphics[width=0.1\textwidth]{Pictures/deg_im_9.png}}
    \hfill\\
    \subfloat[$f_h$]{
         \includegraphics[width=0.45\textwidth]{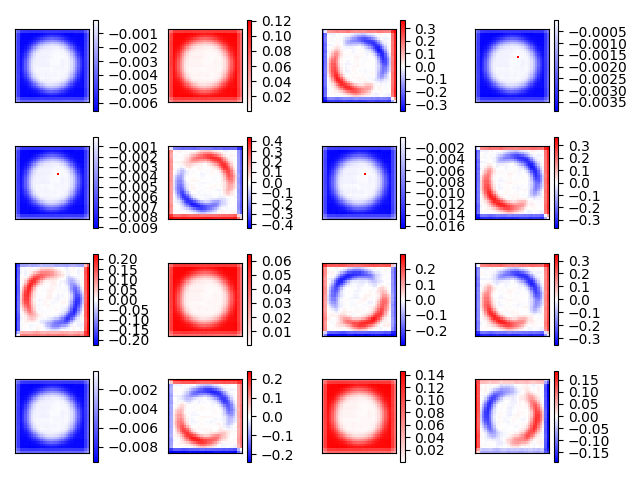}}
    \hfill
    \subfloat[$f_{hb}$]{
         \includegraphics[width=0.45\textwidth]{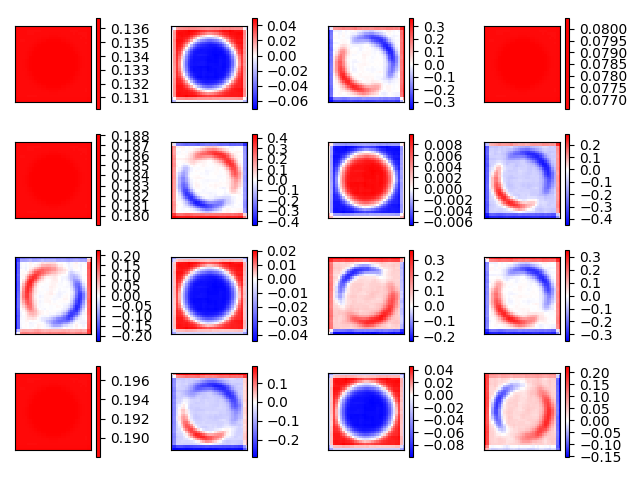}}
    \hfill\\
    \subfloat[$f_\sigma$]{
         \includegraphics[width=0.45\textwidth]{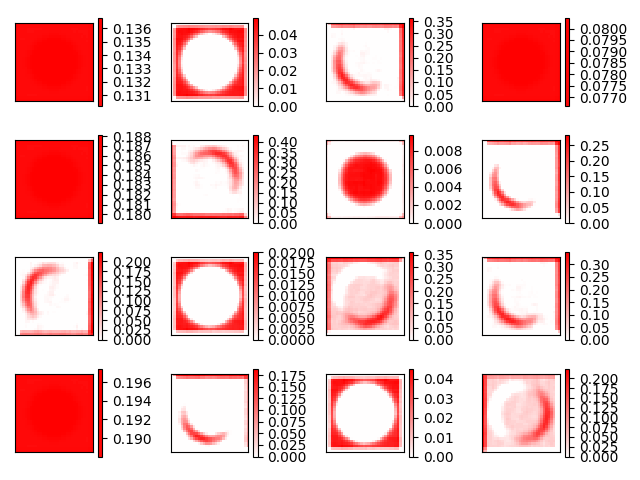}}
    \hfill
    \subfloat[$f_{a\odot}$]{
         \includegraphics[width=0.45\textwidth]{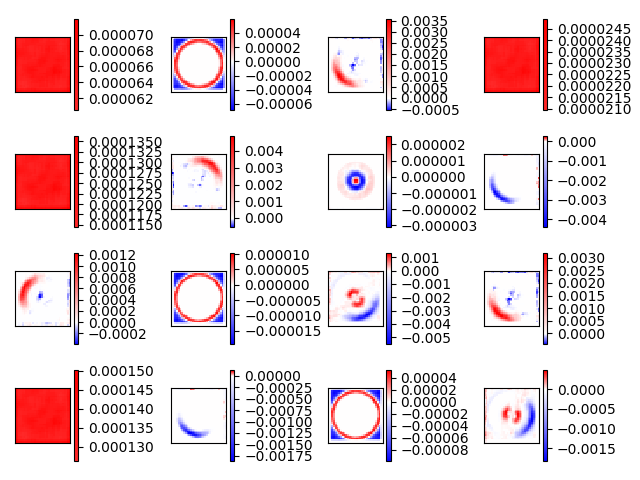}}
    \hfill
\caption[]{Computed features at each level of the neural network on the second example specific image, with a $\relu$ activation network with $16$ convolution filters. The network was trained on data with both polarities.}
\label{fig: circle cnn relu features 16 conv kernel image 9}
\end{figure}

\begin{figure}[!htbp]
    \centering
    \subfloat[$f_\theta$]{
         \includegraphics[width=0.15\textwidth]{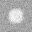}}
    \hfill
    \subfloat[$f_h$]{
         \includegraphics[width=0.2\textwidth]{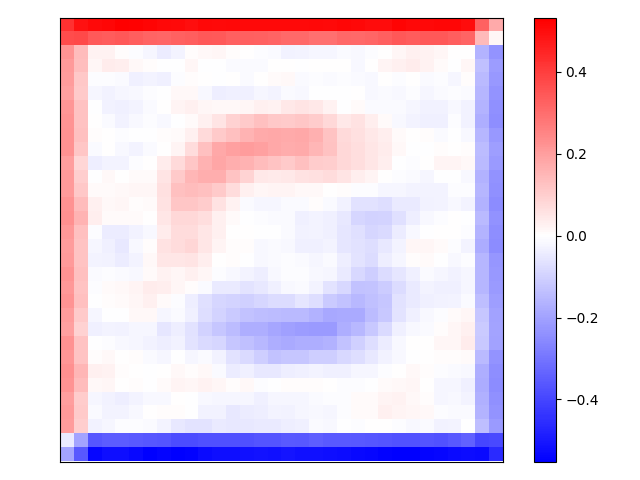}}
    \hfill
    \subfloat[$f_{hb}$]{
         \includegraphics[width=0.2\textwidth]{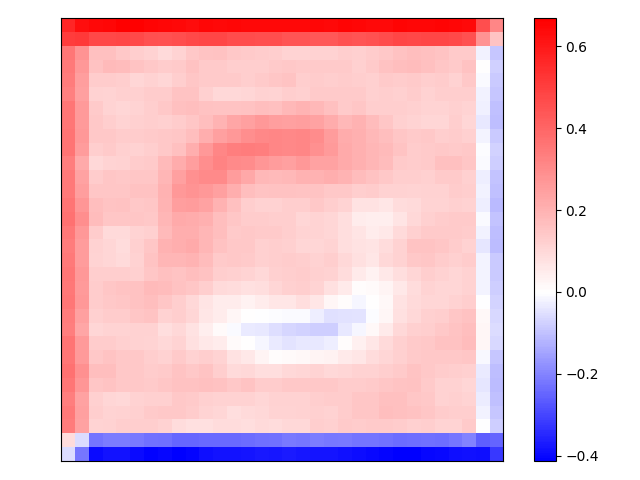}}
    \hfill
    \subfloat[$f_\sigma$]{
         \includegraphics[width=0.2\textwidth]{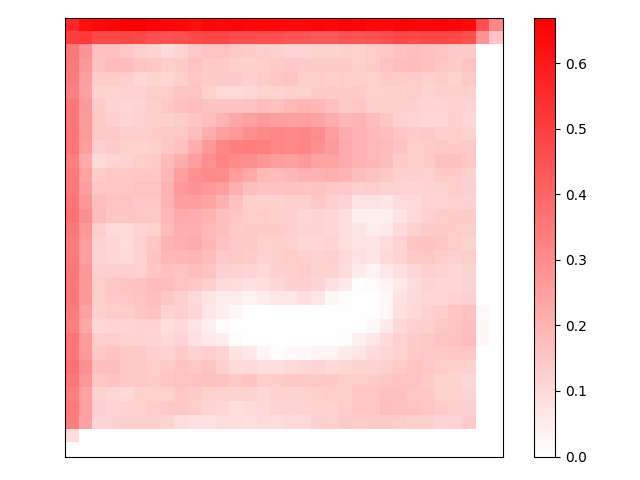}}
    \hfill
    \subfloat[$f_{a\odot}$]{
         \includegraphics[width=0.2\textwidth]{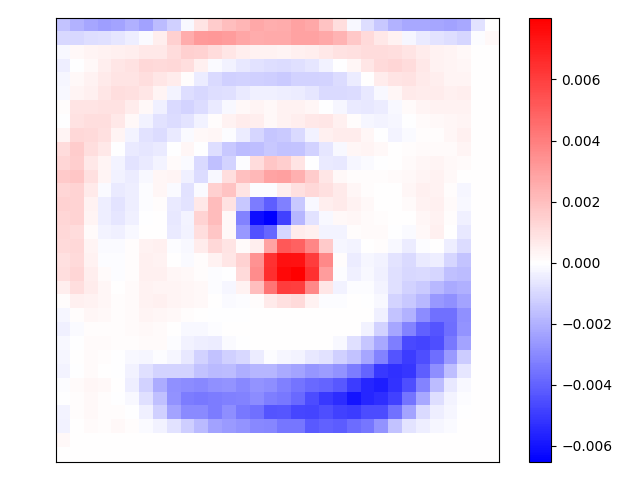}}
    \hfill\\
    \subfloat[$f_\theta$]{
         \includegraphics[width=0.15\textwidth]{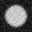}}
    \hfill
    \subfloat[$f_h$]{
         \includegraphics[width=0.2\textwidth]{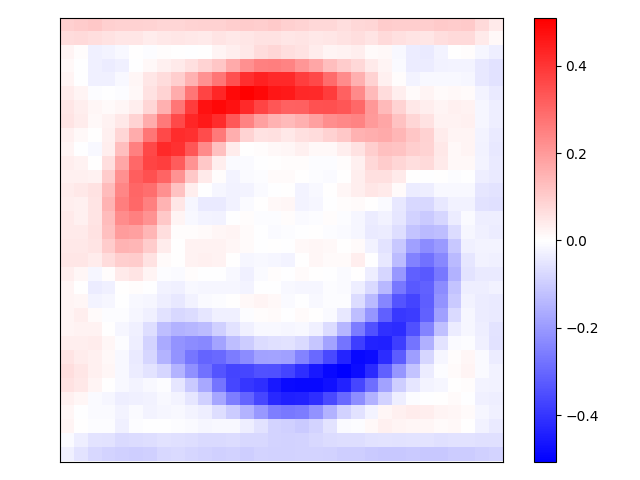}}
    \hfill
    \subfloat[$f_{hb}$]{
         \includegraphics[width=0.2\textwidth]{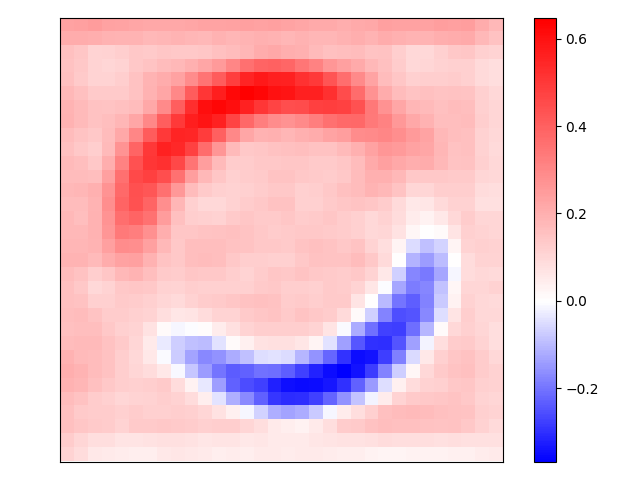}}
    \hfill
    \subfloat[$f_\sigma$]{
         \includegraphics[width=0.2\textwidth]{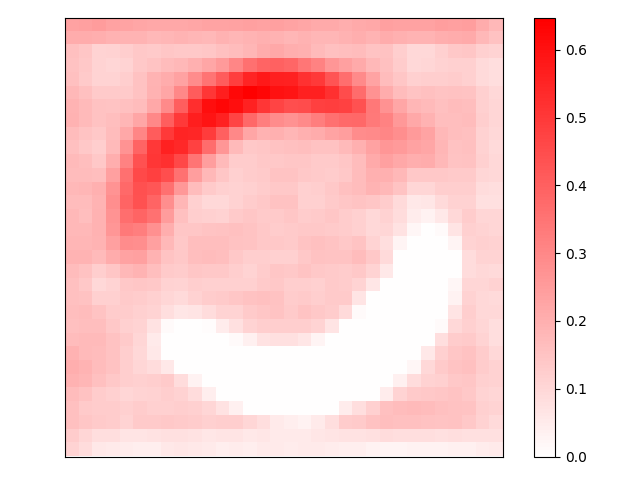}}
    \hfill
    \subfloat[$f_{a\odot}$]{
         \includegraphics[width=0.2\textwidth]{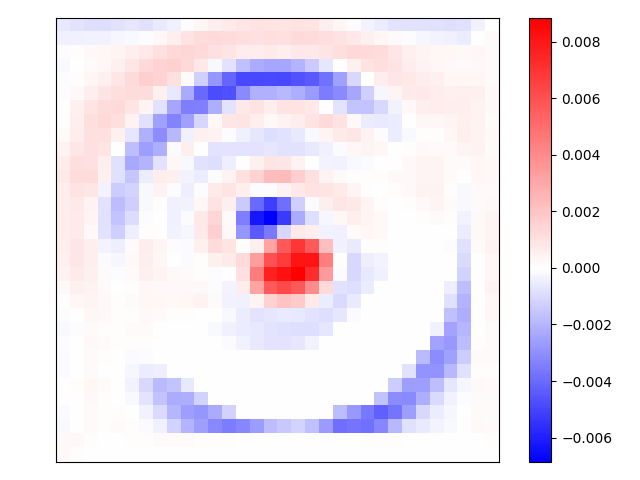}}
    \hfill
\caption[]{Computed features at each level of the neural network on a specific instance (left), with a $\relu$ activation network with $C=1$ convolution filter. The network was trained on positive-only polarity data.}
\label{fig: circle cnn relu features 1 conv kernel isF_bright True}
\end{figure}

\begin{figure}[!htbp]
    \centering
    \subfloat[$f_\theta$]{
         \includegraphics[width=0.08\textwidth]{Pictures/isF_bright_True_deg_im_22.png}}
    \subfloat[$f_h$]{
         \includegraphics[width=0.23\textwidth]{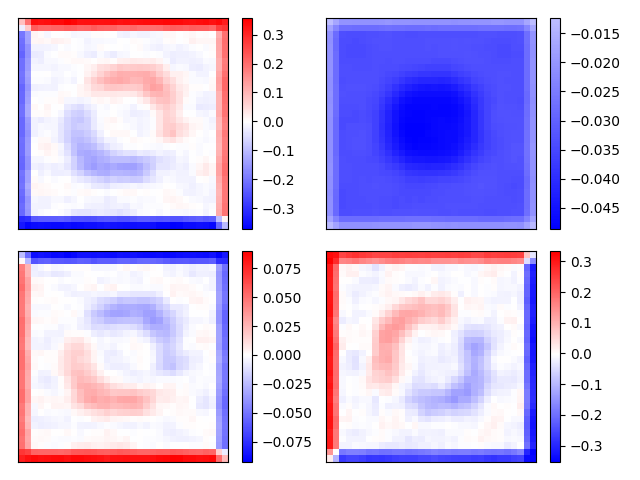}}
    \subfloat[$f_{hb}$]{
         \includegraphics[width=0.23\textwidth]{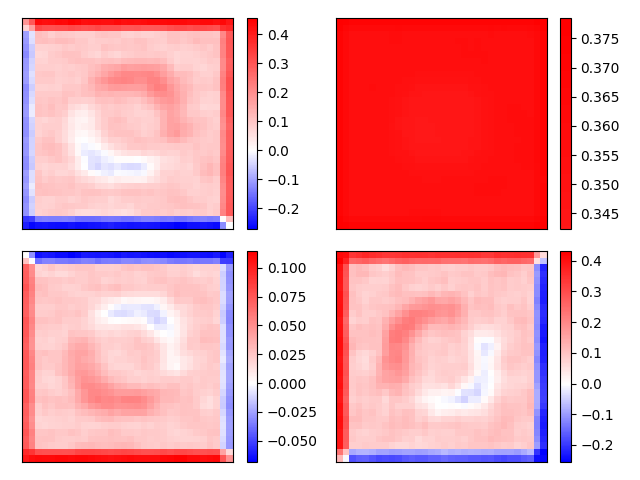}}
    \hfill\\
    \hspace{3.7em}
    \subfloat[$f_\sigma$]{
         \includegraphics[width=0.23\textwidth]{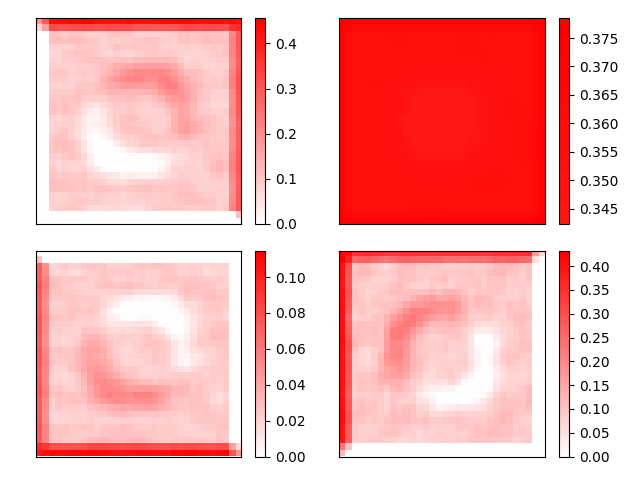}}
    \subfloat[$f_{a\odot}$]{
         \includegraphics[width=0.23\textwidth]{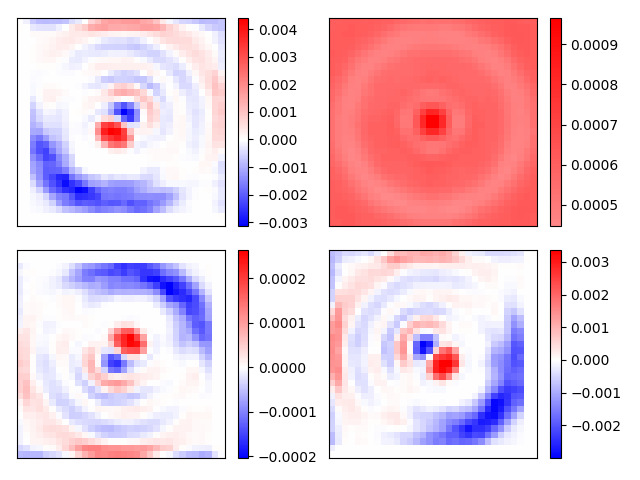}}
    \hfill\\
    \subfloat[$f_\theta$]{
         \includegraphics[width=0.08\textwidth]{Pictures/isF_bright_True_deg_im_9.png}}
    \subfloat[$f_h$]{
         \includegraphics[width=0.23\textwidth]{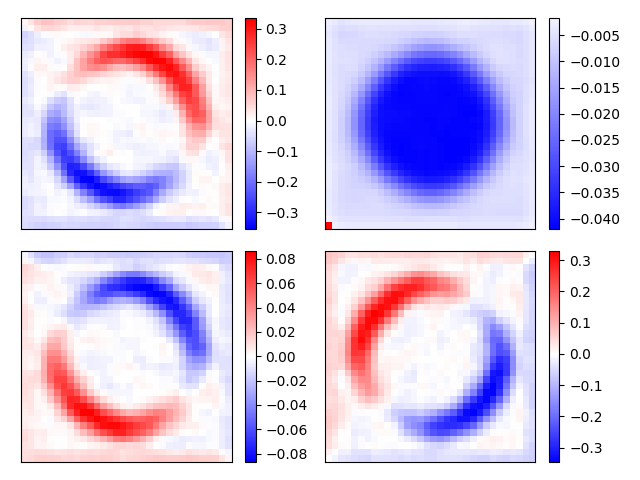}}
    \subfloat[$f_{hb}$]{
         \includegraphics[width=0.23\textwidth]{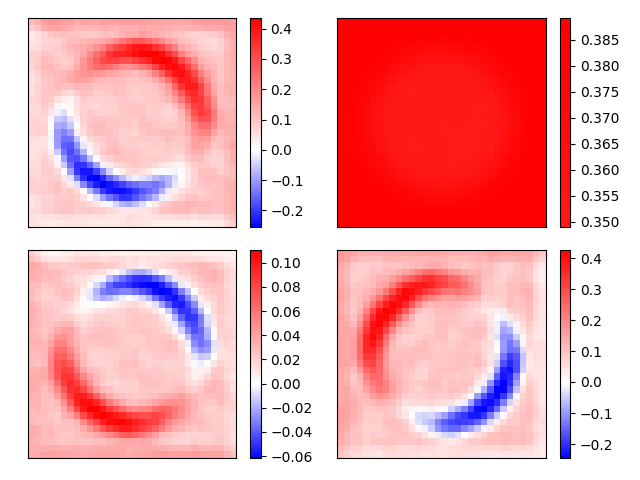}}
    \hfill\\
    \hspace{3.7em}
    \subfloat[$f_\sigma$]{
         \includegraphics[width=0.23\textwidth]{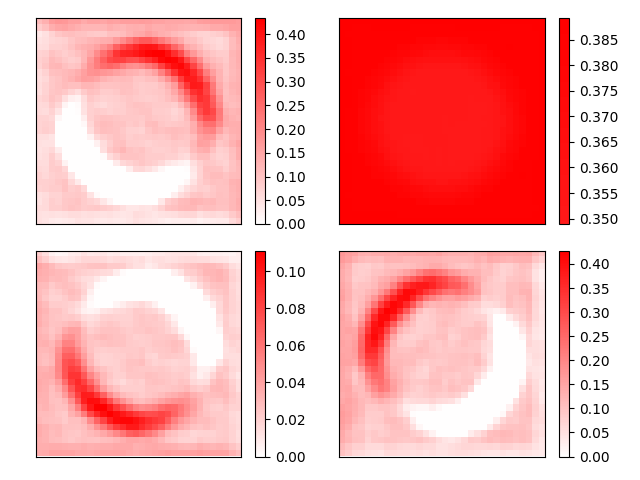}}
    \subfloat[$f_{a\odot}$]{
         \includegraphics[width=0.23\textwidth]{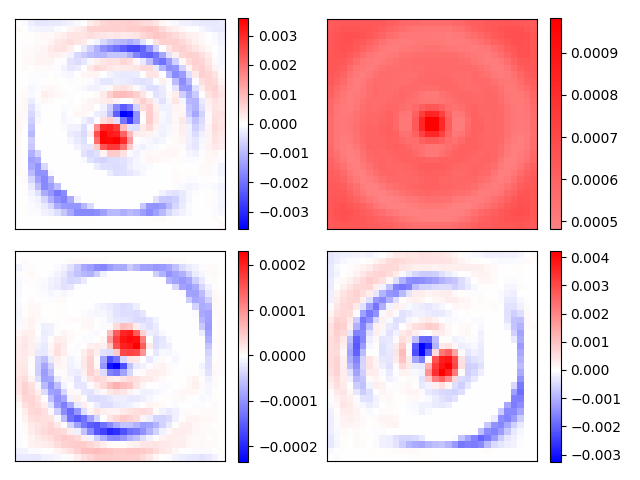}}
    \hfill
\caption[]{Computed features at each level of the neural network on a specific instance (left), with a $\relu$ activation network with $C=4$ convolution filters. The network was trained on positive-only polarity data.}
\label{fig: circle cnn relu features 4 conv kernel isF_bright True}
\end{figure}

\begin{figure}[!htbp]
    \centering
    \subfloat[$f_\theta$]{
         \includegraphics[width=0.1\textwidth]{Pictures/isF_bright_True_deg_im_22.png}}
    \hfill\\
    \subfloat[$f_h$]{
         \includegraphics[width=0.45\textwidth]{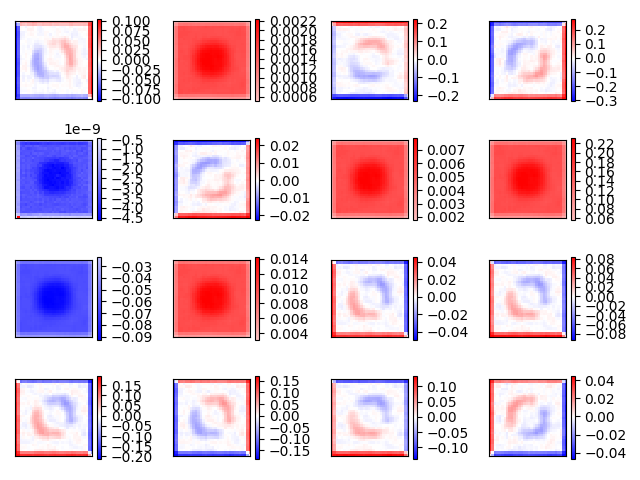}}
    \hfill
    \subfloat[$f_{hb}$]{
         \includegraphics[width=0.45\textwidth]{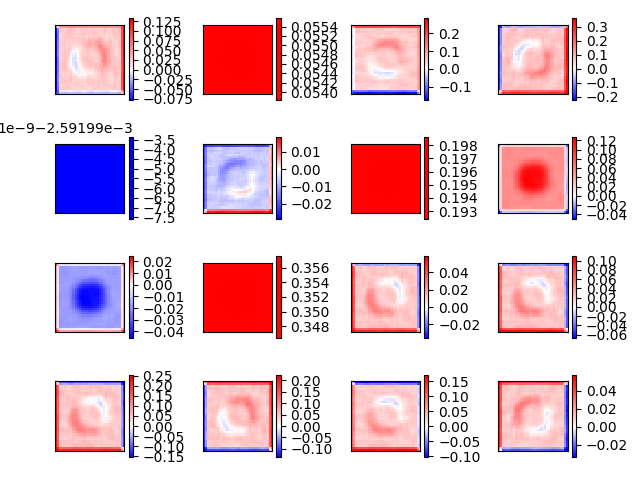}}
    \hfill\\
    \subfloat[$f_\sigma$]{
         \includegraphics[width=0.45\textwidth]{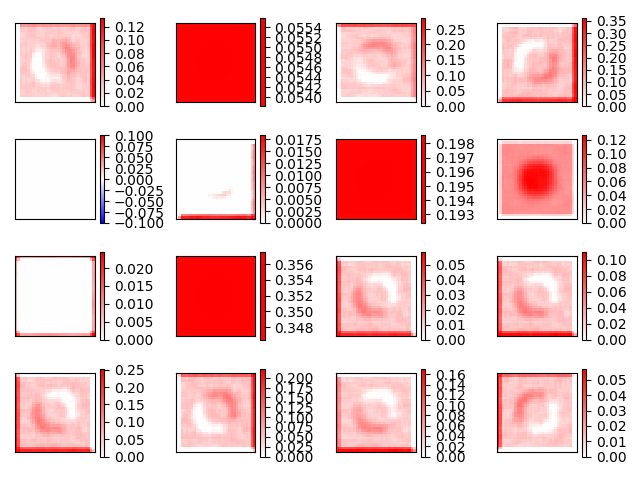}}
    \hfill
    \subfloat[$f_{a\odot}$]{
         \includegraphics[width=0.45\textwidth]{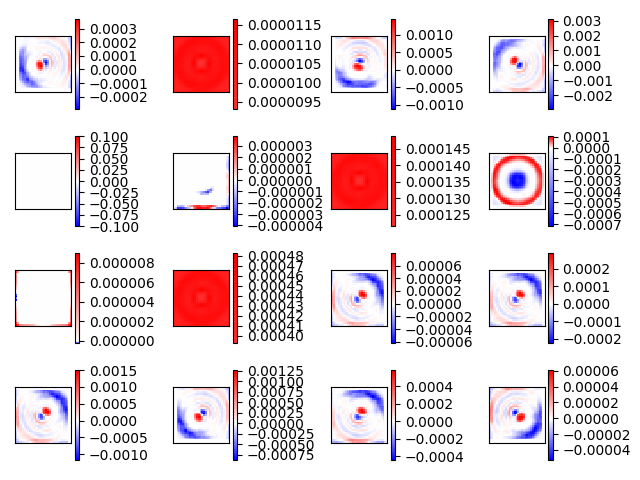}}
    \hfill\\
\caption[]{Computed features at each level of the neural network on the first example specific image, with a $\relu$ activation network with $16$ convolution filters. The network was trained on positive-only polarity data.}
\label{fig: circle cnn relu features 16 conv kernel image 2 isF_bright True}
\end{figure}

\begin{figure}[!htbp]
    \centering
    \subfloat[$f_\theta$]{
         \includegraphics[width=0.1\textwidth]{Pictures/isF_bright_True_deg_im_9.png}}
    \hfill\\
    \subfloat[$f_h$]{
         \includegraphics[width=0.45\textwidth]{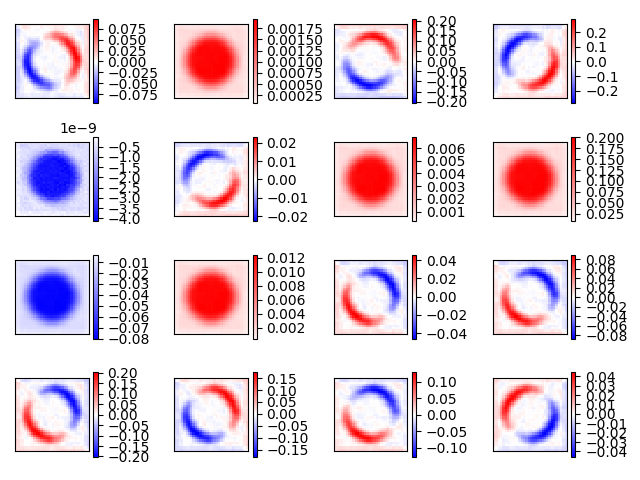}}
    \hfill
    \subfloat[$f_{hb}$]{
         \includegraphics[width=0.45\textwidth]{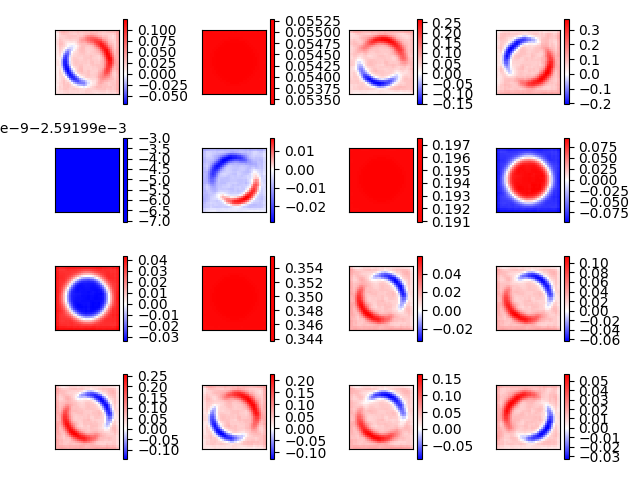}}
    \hfill\\
    \subfloat[$f_\sigma$]{
         \includegraphics[width=0.45\textwidth]{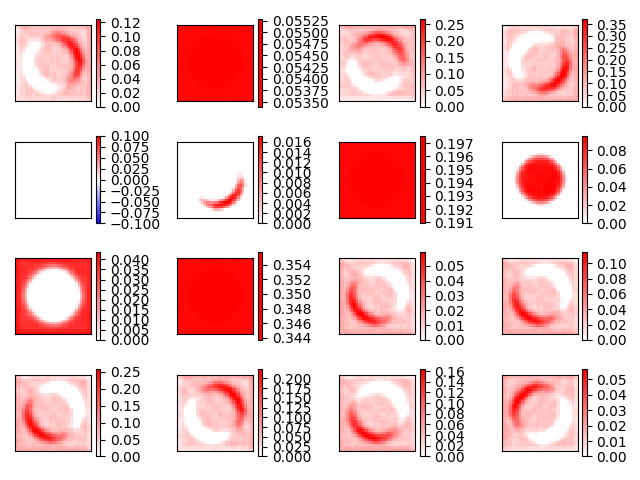}}
    \hfill
    \subfloat[$f_{a\odot}$]{
         \includegraphics[width=0.45\textwidth]{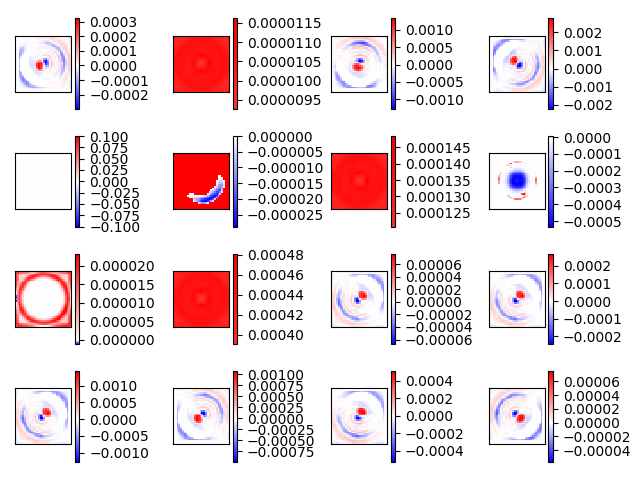}}
    \hfill
\caption[]{Computed features at each level of the neural network on the second example specific image, with a $\relu$ activation network with $16$ convolution filters. The network was trained on positive-only polarity data.}
\label{fig: circle cnn relu features 16 conv kernel image 9 isF_bright True}
\end{figure}

Understanding can also arise from visual presentations. We thus also plot in Figure \ref{fig: circle cnn relu features 1 conv kernel} (resp. Figure \ref{fig: circle cnn relu features 1 conv kernel isF_bright True}) all the intermediate representations of random data fed to the networks trained on both (resp. positive-only) polarity  data. We do the same in Figures \ref{fig: circle cnn relu features 4 conv kernel}, \ref{fig: circle cnn relu features 16 conv kernel image 2}, and \ref{fig: circle cnn relu features 16 conv kernel image 9} and Figures \ref{fig: circle cnn relu features 4 conv kernel isF_bright True}, \ref{fig: circle cnn relu features 16 conv kernel image 2 isF_bright True}, and \ref{fig: circle cnn relu features 16 conv kernel image 9 isF_bright True} where this time the networks have $C\in\{4,6\}$ convolution filters in the same layer, in order to help provide visual help to understanding what is going on in this architecture extension.

\subsubsection{On the quality of the converged networks}
\label{subsec: circle on quality of converged networks}

\begin{figure}[!htbp]
    \centering
    \subfloat[]{
         \includegraphics[width=0.4\textwidth]{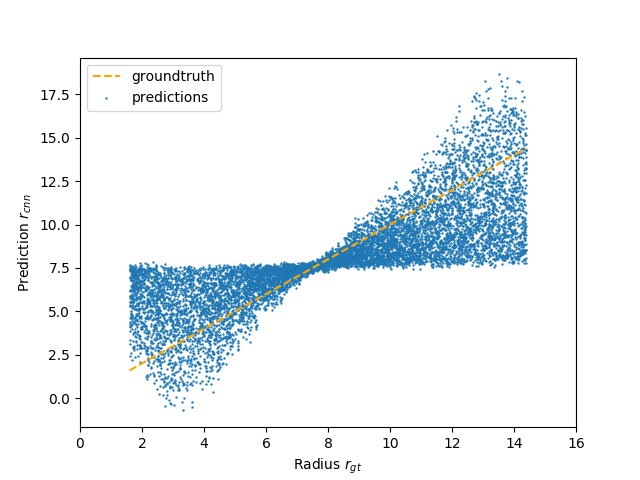}}
    \hfill
    \subfloat[]{
         \includegraphics[width=0.4\textwidth]{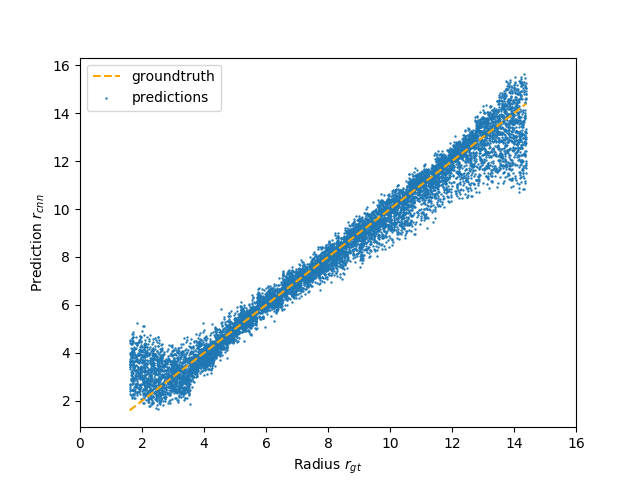}}
\caption[]{Performance of the $\relu$ activation neural networks using $C=1$ convolution filter. On the left (resp. right), the network is trained and tested on data with both (resp. positive-only) polarities. Each dot represents one of the $N=10000$ estimations on the test dataset.}
\label{fig: MAIN circle relu estimations}
\end{figure}

Our quantitative performance indicator for the networks is the Root Means Squared Error (RMSE) scaled once again to pixel size in the same way by multiplying the estimations (or analogously the RMSE) by $\tfrac{D}{2}$ with reference dimensionality $D=32$. The data is noisy with $\sigma_n = \tfrac{10}{255}$. The network trained on positive polarity only data performs significantly better with a score of $0.7$ compared to $2.6$ for the one trained on the dataset with both polarities present (see Table \ref{tab: MAIN circle rmse}). Thanks to the increase in dimensionality, the Gaussian noise is more easily mitigated. We also look at the actual estimations on the test data rather than just the global RMSE score, plotted in Figure \ref{fig: MAIN circle relu estimations}. It is clear that the network trained on data with both polarities learns accurately to measure a radius, albeit with some precision errors, whereas the behaviour of the other network trained on a dataset with both polarities present was not able to grasp the concept of a radius, with many catastrophic failure estimations at all radii levels equal to the average radius of the dataset.

\begin{table}[t]
        \begin{center}
        \begin{tabular}{lccccccc}
            \toprule
                     & \multicolumn{3}{c}{Convolutional Neural Networks}\\ \cmidrule{2-4}
                     & Both polarities & & Positive polarity\\
            \midrule                                     
            $RMSE$ &  2.61833 & & 0.68386  \\
            \bottomrule
        \end{tabular}\\[0.5em]
        \end{center}
    \caption{Root Mean Squared Error performance for the radius estimation of the neural estimators using $\relu$ activation and $C=1$ convolution filter. The networks were trained and tested on separate datasets of $10000$ random images each, consisting in either both polarities or just the positive polarity (with noise level $\sigma_n = \frac{10}{255}$). The RMSE is given in pixel unit and here the resolution is $D=32$.}
    \label{tab: MAIN circle rmse}
\end{table}

\subsubsection{Changing the architecture: adding more channels}
\label{subsec: appendix circle more conv channels}

Here, the convolution layer will have $C\ge 1$ channels. It is important to remember that when computing the final estimation, there is no switch mechanism allowing to choose only one channel to consider for the final computation. All will be processed with equal blind importance, only the weight of the filters and of the fully connected layers differentiate them, but they do not change between two different instances at inference time.

\begin{figure}[!htbp]
    \centering
    \subfloat[$C=1$]{
         \includegraphics[width=0.15\textwidth]{Pictures/circle_relu_model_last_epoch_conv_ker_channels_1.png}}
    \hspace{14em}
    \subfloat[$C=1$]{
         \includegraphics[width=0.15\textwidth]{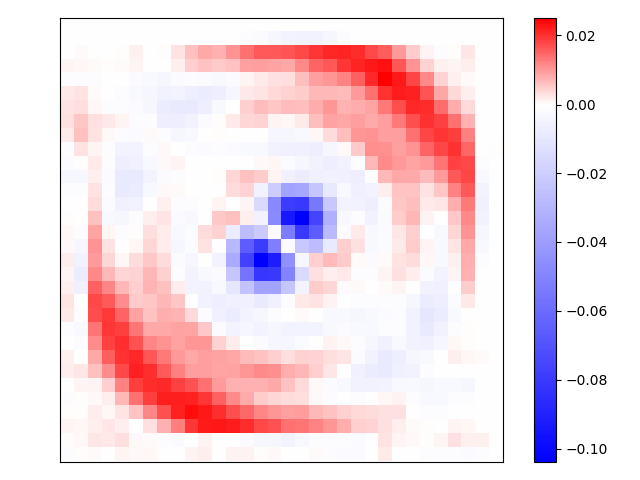}}
    \hfill\\
    \subfloat[$C=4$]{
         \includegraphics[width=0.33\textwidth]{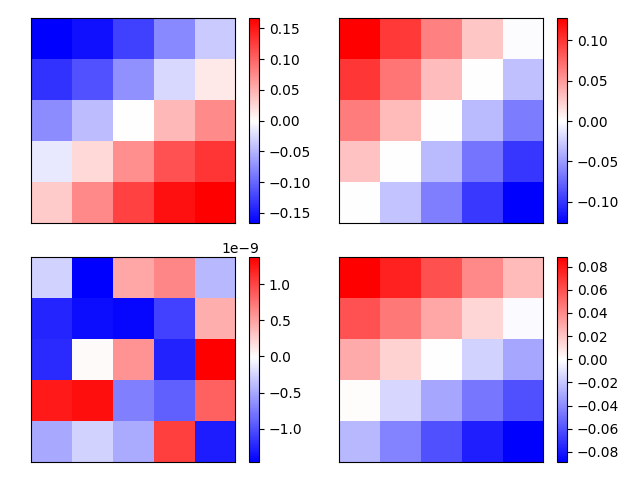}}
    \hspace{7em}
    \subfloat[$C=4$]{
         \includegraphics[width=0.33\textwidth]{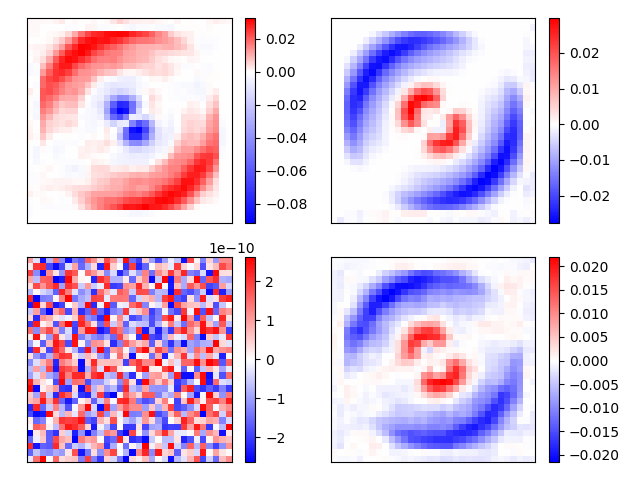}}
    \hfill\\
    \subfloat[$C=16$]{
         \includegraphics[width=0.49\textwidth]{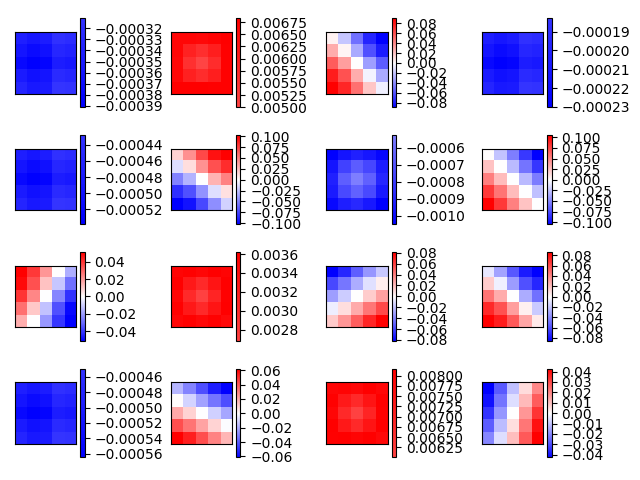}}
    \hfill    
    \subfloat[$C=16$]{
         \includegraphics[width=0.49\textwidth]{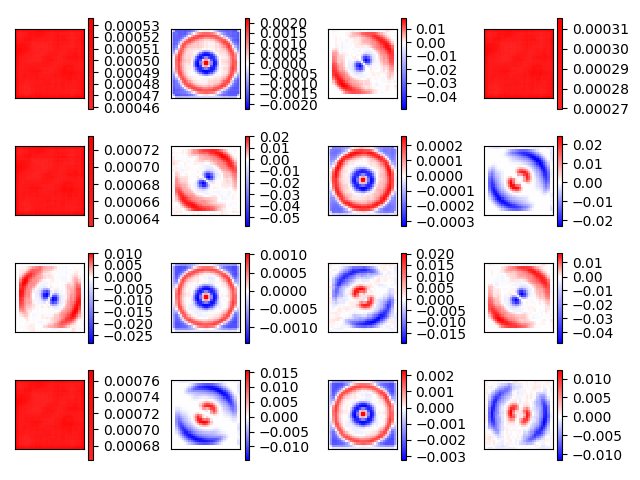}}
    \hfill\\
\caption[]{Learned weights of the convolution networks using $\relu$ activations and $C\in\{1,4,16\}$ convolution filters trained on data with both polarities. Left: the correlation filters. Right: the weights of the fully connected layer. We normalised the colouring scheme so that white corresponds to the value $0$, strong blue to the smallest negative value, and strong red to the largest positive value.}
\label{fig: appendix circle cnn relu weights}
\end{figure}

\begin{figure}[!htbp]
    \centering
    \subfloat[$C=1$]{
         \includegraphics[width=0.15\textwidth]{Pictures/circle_relu_isF_bright_True_model_last_epoch_conv_ker_channels_1.png}}
    \hspace{14em}
    \subfloat[$C=1$]{
         \includegraphics[width=0.15\textwidth]{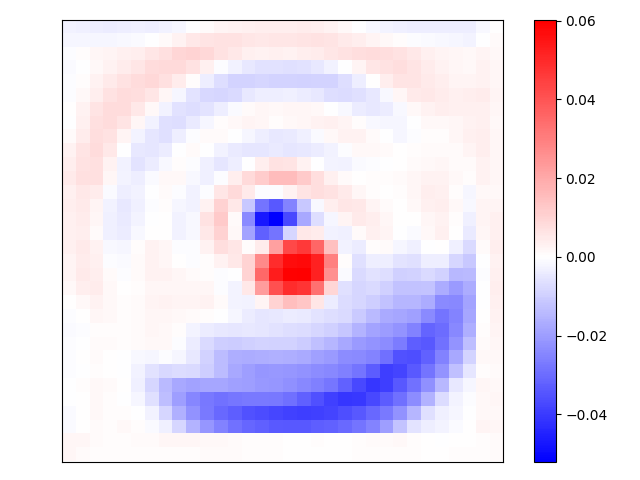}}
    \hfill\\
    \subfloat[$C=4$]{
         \includegraphics[width=0.33\textwidth]{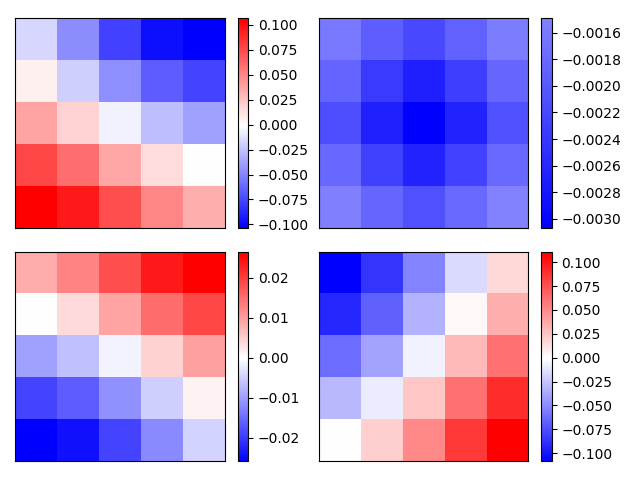}}
    \hspace{7em}
    \subfloat[$C=4$]{
         \includegraphics[width=0.33\textwidth]{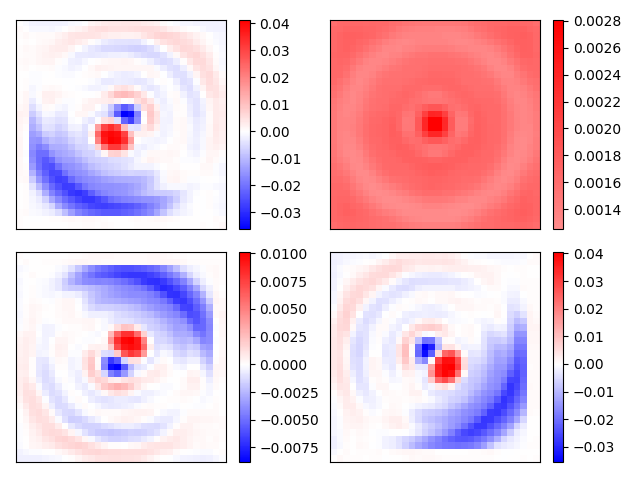}}
    \hfill\\
    \subfloat[$C=16$]{
         \includegraphics[width=0.49\textwidth]{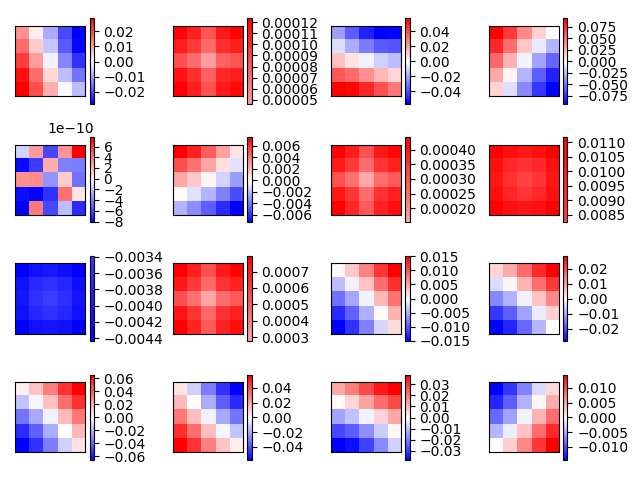}}
    \hfill    
    \subfloat[$C=16$]{
         \includegraphics[width=0.49\textwidth]{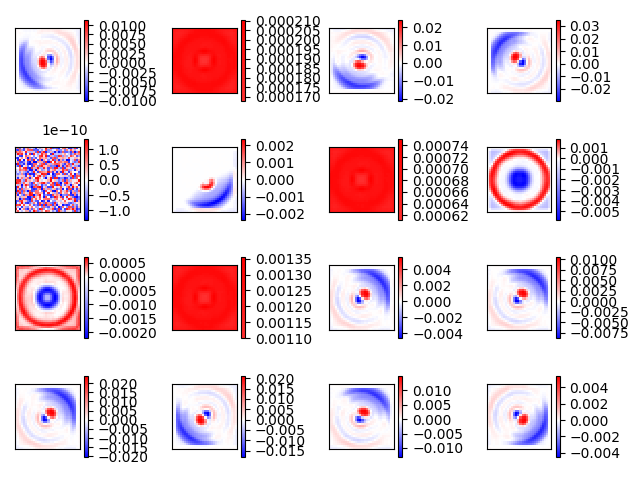}}
    \hfill\\
\caption[]{Learned weights of the convolution networks using $\relu$ activations and $C\in\{1,4,16\}$ convolution filters trained on data with positive-only polarity. Left: the correlation filters. Right: the weights of the fully connected layer. We normalised the colouring scheme so that white corresponds to the value $0$, strong blue to the smallest negative value, and strong red to the largest positive value.}
\label{fig: appendix circle cnn relu weights isF_bright True}
\end{figure}

\paragraph{Training results} As we do not change the depth of the network, each channel has a similar structure to a network with $C=1$ channel as described previously. Thus, it is impossible to provide non trivial and meaningful invariant representations to the intensity difference in all cases regardless of it. However, the fully connected final layer somewhat mixes all channels together allowing complex compensation mechanisms to remove the intensity dependence. Unfortunately, theoretically analysing these phenomena is significantly harder making designing such networks by hand a challenge. We nevertheless trained networks with $C\in\{4,16\}$ channels and display the learned layers in Figures \ref{fig: appendix circle cnn relu weights} and \ref{fig: appendix circle cnn relu weights isF_bright True}. We encourage to also refer to Figures \ref{fig: circle cnn relu features 1 conv kernel}, \ref{fig: circle cnn relu features 4 conv kernel}, \ref{fig: circle cnn relu features 16 conv kernel image 2}, and \ref{fig: circle cnn relu features 16 conv kernel image 9} to visually see what occurs to instances when fed through these networks.

\begin{figure}[!htbp]
    \centering
    \subfloat[$C=1$]{
         \includegraphics[width=0.32\textwidth]{Pictures/circle_relu_nbKerConv_1_preds_sorted_by_rgt.png}}
    \hfill
    \subfloat[$C=4$]{
         \includegraphics[width=0.32\textwidth]{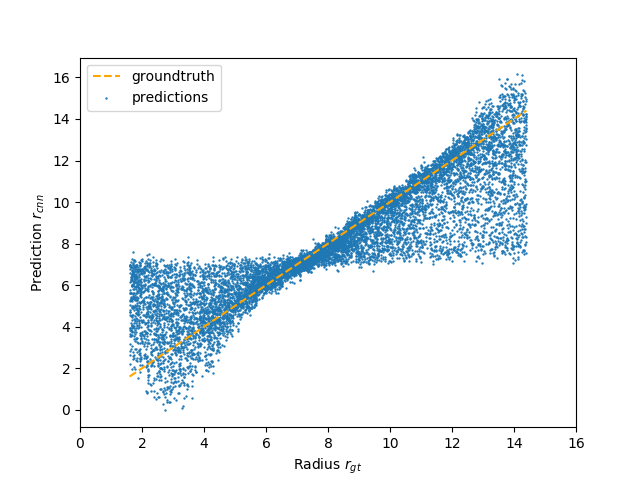}}
    \hfill
    \subfloat[$C=16$]{
         \includegraphics[width=0.32\textwidth]{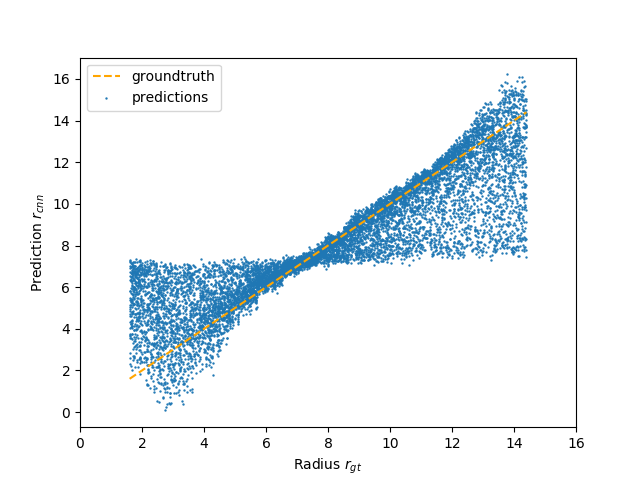}}
    \hfill\\
    \subfloat[$C=1$]{
         \includegraphics[width=0.32\textwidth]{Pictures/circle_relu_nbKerConv_1_isF_bright_True_preds_sorted_by_rgt.png}}
    \hfill
    \subfloat[$C=4$]{
         \includegraphics[width=0.32\textwidth]{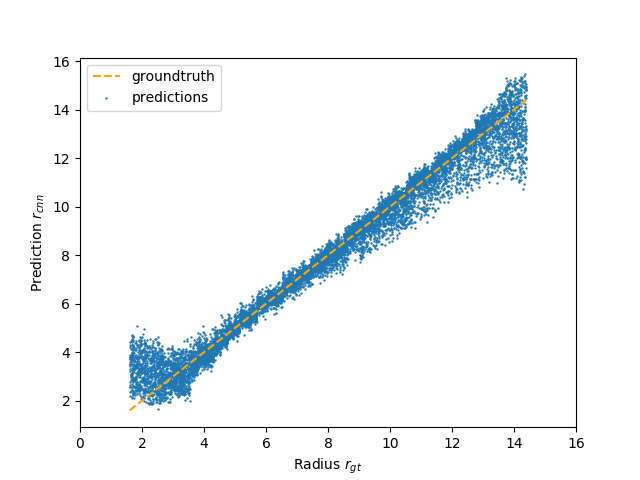}}
    \hfill
    \subfloat[$C=16$]{
         \includegraphics[width=0.32\textwidth]{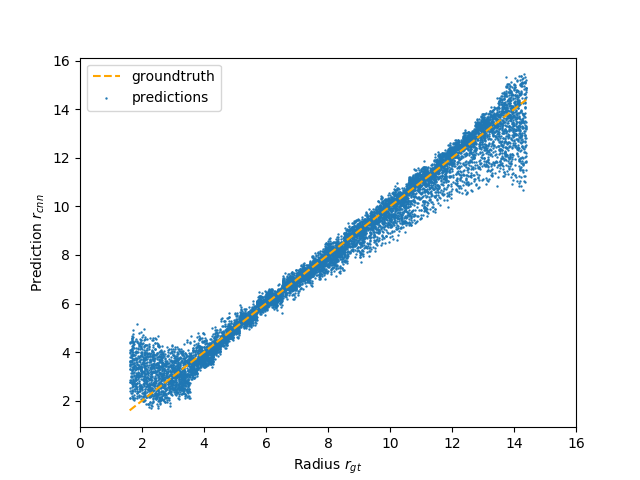}}
    \hfill\\
\caption[]{Performance of the $\relu$ activation neural networks trained and tested on both polarity data (top) and positive-only polarity data (bottom). From left to right, the networks use $C= 1,4,16$ convolution filters. Each dot represents one of the $N=10000$ estimations of the network on the test datasets.}
\label{fig: appendix circle estimations and errors}
\end{figure}

\paragraph{Performance} When looking at performance, increasing the number of filters seems to have a small or negligible impact. Improvement mostly lies in the case when the dataset has both polarities present, but it is not sufficient to learn the radius concept with an RMSE decreasing only to $1.7$ with $C=16$ filters, far from the $0.6$ of the fixed polarity case (see Table \ref{tab: appendix circle rmse}). This is also confirmed when looking at all the estimations (see Figure \ref{fig: appendix circle estimations and errors}).

\begin{table}[t]
        \begin{center}
        \begin{tabular}{lccccccc}
            \toprule
                     & \multicolumn{3}{c}{Both polarities} & & \multicolumn{3}{c}{Positive polarity}  \\ \cmidrule{2-4} \cmidrule{6-8}
                     & $C=1$&  $C=4$ & $C=16$ & & $C=1$&  $C=4$ & $C=16$  \\ 
            \midrule
            $RMSE$ & 2.61833 & 1.76942 &  1.74973 & & 0.68386 & 0.64562 & 0.64027 \\
            \bottomrule
        \end{tabular}
        \end{center}
    \caption{Root Mean Squared Error performance for the radius estimation of the neural networks with $\relu$ activations using $C\in\{1,4,16\}$ convolution filters, trained on $N=10000$ images of either both polarities or positive-only polarity, and tested on the same amount of data with the same polarity property as the training set.}
    \label{tab: appendix circle rmse}
\end{table}

We now discuss the types of filters observed in both cases.

\paragraph{Weight maps analysis - Network trained on both polarities} We see that there are three types of learned convolution filters. The first type consists in the filters with a straight white line passing through the middle of the filter, separating red and blue areas. These filters are clearly edge filters with some random orientation. The second type regroups fully red filters. They correspond to some type of positive averaging filters. The last type of filters are the fully blue ones. They correspond to some negative averaging filters. We could also add a fourth filter, found only once here, corresponding in the $0$ filter: it is the bottom left filter of the $C=4$ case, with all entries of absolute value magnitude less than $10^{-9}$. 

The edge filter was chosen when having only one filter, and is present in at least half the cases in each network, suggesting that it might be of superior importance to the other averaging ones, which could be useful for refinements of the estimation. Furthermore, inside a network, the edge filters do not seem to have random independent orientation. The $4$-convolution network chose one random direction and created edge filters along it with very small orientation perturbation. The $16$-convolutional network chose only two directions, being orthogonal to each other, and then created the filters along those directions with small orientation perturbation. This fact is surprising, as we would expect the edge orientations to be chosen fully randomly due to the orientation invariance of the data. It is possible that sampling along vastly different orientations allows to recover ``low frequency'' information in some sense, whereas sampling along small perturbations of a fixed direction boosts the recovery of ``high frequency'' information or details. Since the data is invariant to orientation, sampling along very different orientations might not provide much detailed information and thus the network preferred the later choice.

The positive averaging filters have values several orders of magnitude smaller than those of the edge filters. This difference also holds for their associated linear weights in the fully connected layer, which confirms that channels of these filters are of lesser importance than those of the edge filters and will only contribute to a refinement of the estimation.

The negative averaging filters all have very small entries. Due to the positivity of the data, it is not so clear if these filters have created a dying $\relu$ phenomenon, when the entry to the $\relu$ is entirely negative for all data point implying that the $\relu$ zeros out the entire data and learning is no longer possible along this channel, or whether the bias after the convolution brings some positive information through the $\relu$. For dying $\relu$s, the $L^2$ regularisation pushes all the weights to $0$.

For the approximately $0$ convolution filter, which may be the result of a dying $\relu$ combined with $L^2$ regularisation, all information has been approximately zeroed-out by the filter, thus the $L^2$ regularisation also brings the associated weight map to $0$.

Concerning the learned fully connected weight maps, there are essentially as many types of maps than there are of filters, with a one-to-one mapping between a class of convolution filter and of weight map.

The edge filters are associated with symmetric maps oriented along the edge filter's direction. We can classify these maps into two sub-categories. The first category groups the maps with large red circular arc blobs in the periphery, followed by a white circle in the middle, and with blue blobs close to the centre of the image. The other one groups the maps with large blue circular arc blobs in the periphery, followed by a white circle, red circular arc blobs close to the centre, and a small white disk in the centre. It is highly non trivial why these two categories exist as it is not so clear what they represent. Their presence is due to the existence of both polarities in the training data and are probably used together as a compensation mechanism of some sort. One may think that they measure the perimeter of the circle, by measuring it in the top right corner and bottom left corner for the map with $C=1$. Indeed, if the radius is large, then the positive outputs of the filter will mostly lie in either the top-right or bottom-left corner, depending on the polarity of the image, and they then will be multiplied by positive numbers suggesting pushing the radius estimation up. On the other hand, if the radius is small, the positive outputs of the filter lie much closer to the centre, and will be located on the blue weight blobs. This suggests that the linear map tries to take the value of the estimation down (negatively). Finally, if the radius is in the mid range, approximately $8$ pixels, then the positive outputs of the filter will be superimposed with the white weights of the fully connected layer, suggesting that the network wants to estimate 0 for them. Thus, we could believe that the network is performing some kind of perimeter estimation by measuring the length of a quarter of the circle. The estimation is unbiased in the sense that $0$ corresponds to the average radius on the training set, and that the final bias of the network corrects this. While we may convince ourselves with handwaiving that this is what is happening in the networks, at least with $C\in\{1,4\}$, it is not so obvious that it is really true. We insist that we do not entirely believe in this explanation, and this scepticism is reinforced by the analysis of what happens when using data with only one polarity, or when looking at the bad performance on the entire test set.

The positive averaging filters are associated with dartboard-like maps. The corners are blue, and the circle never passes there, thus they contribute to a negative bias in the estimation. Likewise, since all circles have radii bigger than a small non-zero minimum, the centre of the dartboard contributes to a positive bias. Between the centre. It is not so clear as to how these channels refine the estimation via the edge filter channels in order to remove the intensity dependence and what their meaning is.

\paragraph{Weight maps analysis - Network trained on positive polarity only} As in the previous case, the weights do not seem completely random and converge to nice structures. While similar, these structures differ. Let us try to understand what these weights mean, without giving much focus on the biases for simplicity.

As before, there are three kinds of learned convolution filters: the edge filters, the positive (red) averaging filters, and the negative (blue) averaging filters. We could also add a fourth category consisting in the approximately $0$ filter, where each entry has absolute value or order of magnitude less than $10^{-10}$. 

Once again, the edge filter is dominantly present, and is especially the chosen filter when the network is constrained to use only one filter. The edge filter could thus again be of primordial importance, with the averaging operators contributing only to refinements of the estimation. As previously, the edge orientations are not completely random inside each network. It seems that the $4$ and $16$ convolution networks chose a random edge orientation and sampled edge filters along this orientation and its orthogonal one, with small perturbations of the direction around them. This behaviour is similar to what happened in the previous case, and could once again be due to the fact that this sampling approach might be able to recover finer details than methods sampling along very different orientations. 

The positive averaging filters seem to be at least an order of magnitude smaller than those of the edge filters, and this difference also holds for their associated linear weights in the fully connected layer. This observation once again confirms that these filters are of lesser importance than the edge filters and will only contribute to a refinement of the estimation.

As in the previous case, the negative averaging filters are of small importance due to their small magnitude and that of their associated weight maps. It is possible that they are dead neurons. The approximately $0$ filters and their associated approximately $0$ weight maps are most likely due to a dead neuron and then pushed to $0$ by $L^2$ regularisation.

Concerning the fully connected layers and its weights per convolution channel, each type of convolution filter is once again associated to a unique type of weight map. However, they fundamentally differ to what we had previously obtained when training on data with both polarities.

The fully connected weight maps are no longer symmetric along the direction orthogonal to the edge orientation, but only along the direction of the edge. To describe them without confusion, we will focus on the case $C=1$. The bottom right part of the weight map resembles what we had previously obtained: a large outer blue arc, followed by a (thin) white circle and then a small red blob. This is only a similarity, not a perfect copy, as for instance the white circle is particularly thin here, whereas it was a thicker ring in the previous case, or the inner blob was sometimes an inner circle arc previously. However, the top left part completely differs to what we had previously obtained. First, the network learned to compute there only entries with very small absolute value compared to the large weights obtained in the bottom right part of the weight map. This suggests that the network is mostly focusing on what is occurring in the bottom right part of the image and that it uses the feature maps in the top left for a refinement of the estimation. Second, the top left part has a higher frequency ``texture'' than that of the lower right part. Indeed, there, when looking gradually from close to the centre of the weight map to its outer parts, we find first a small red ring, followed by a larger white ring, then a larger blue one, then a larger and thicker white one, and then the opposite alternation: a thin blue ring, followed by a thin white one, and finally a thin red one. Thus, we have twice the weight radial negative to positive ``texture'', and it is inverted with respect to the mid radius. We do not fully understand the meaning of these weights and textures in the top-left corner. We conjecture as in the one-dimensional case that these oscillations may be due to quantisation and discretisation of the domain.

The positive averaging filters are no longer necessarily associated with the mysterious dartboard-like maps. In one case we do find it to occur, but otherwise we see that the network simply attributed uniform positive small weights in the fully connected layer for these filters. Their meaning is not so clear, but they do not contribute much to the final estimation.

We point out that here the weight maps of the fully connected layer associated to the negative averaging filters are dartboards (with small entries). Once again, we do not fully understand what this structure means and how it helps for the estimation.

\begin{figure}[!htbp]
    \centering
    \subfloat{
         \includegraphics[width=0.23\textwidth]{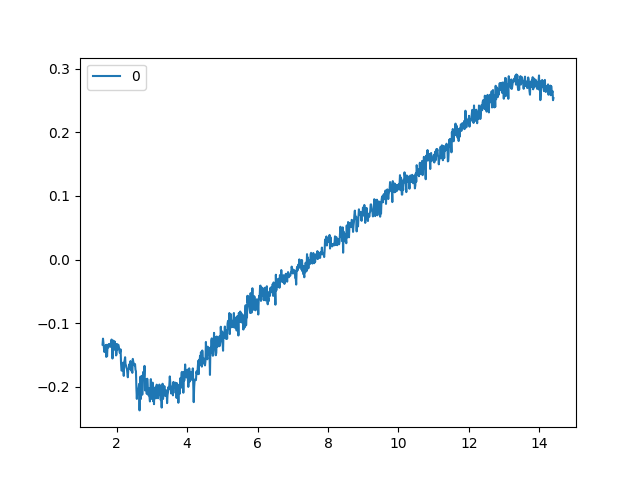}}
    \hfill
    \subfloat{
         \includegraphics[width=0.23\textwidth]{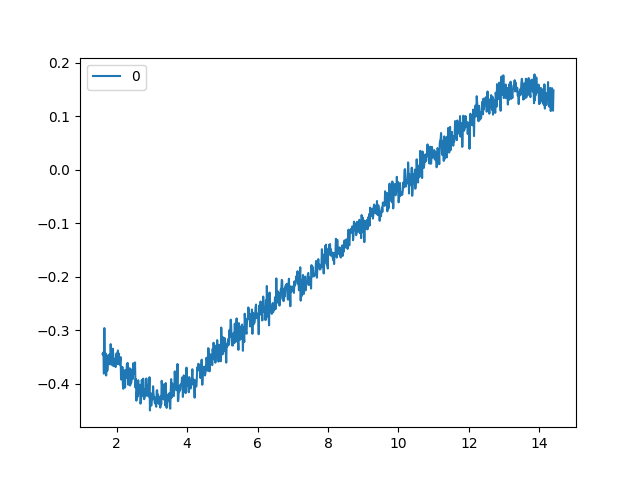}}
    \hfill
    \subfloat{
         \includegraphics[width=0.23\textwidth]{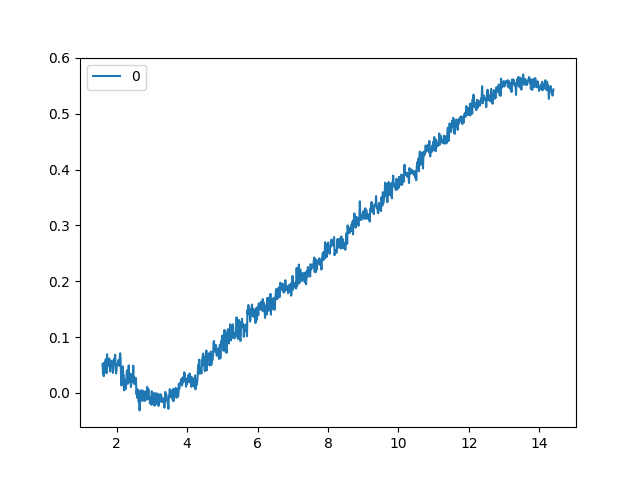}}
    \hfill\\
    \subfloat{
         \includegraphics[width=0.23\textwidth]{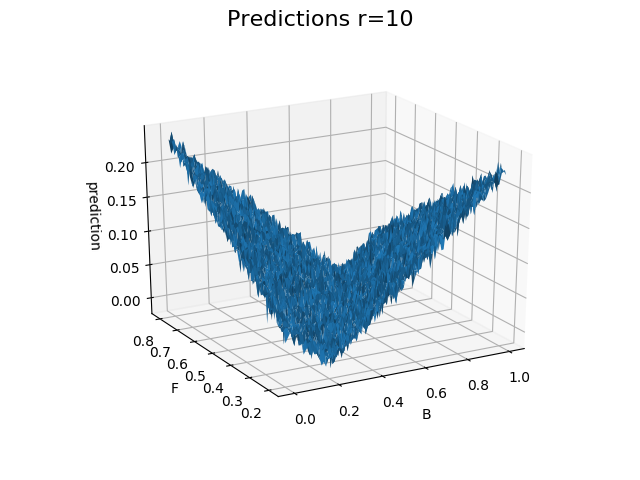}}
    \hfill
    \subfloat{
         \includegraphics[width=0.23\textwidth]{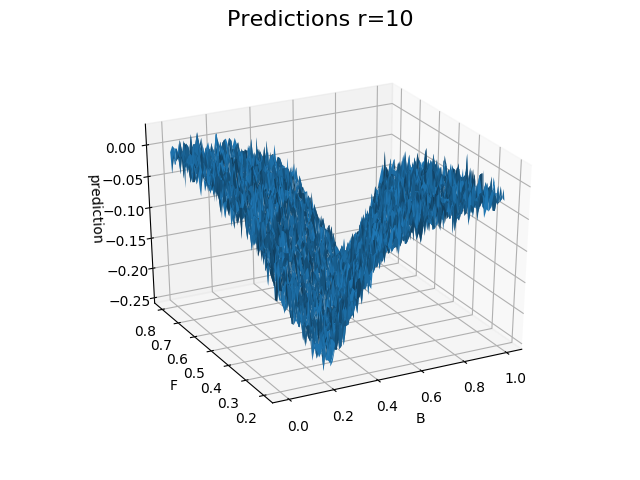}}
    \hfill
    \subfloat{
         \includegraphics[width=0.23\textwidth]{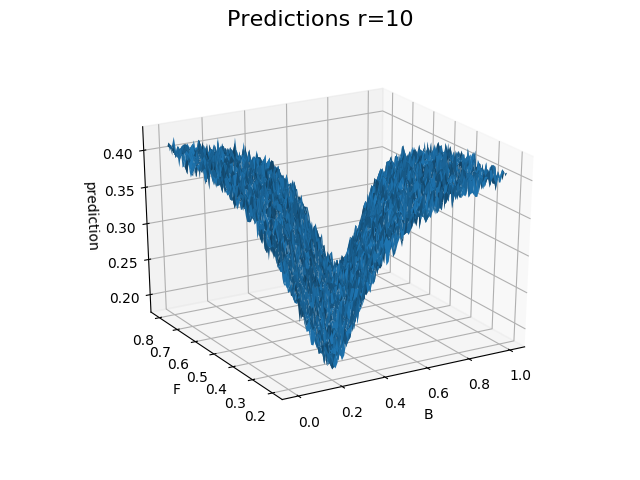}}
    \hfill\\
    \subfloat{
         \includegraphics[width=0.23\textwidth]{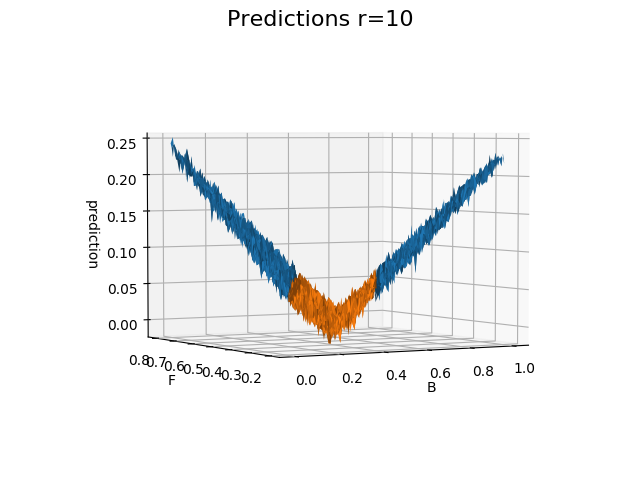}}
    \hfill
    \subfloat{
         \includegraphics[width=0.23\textwidth]{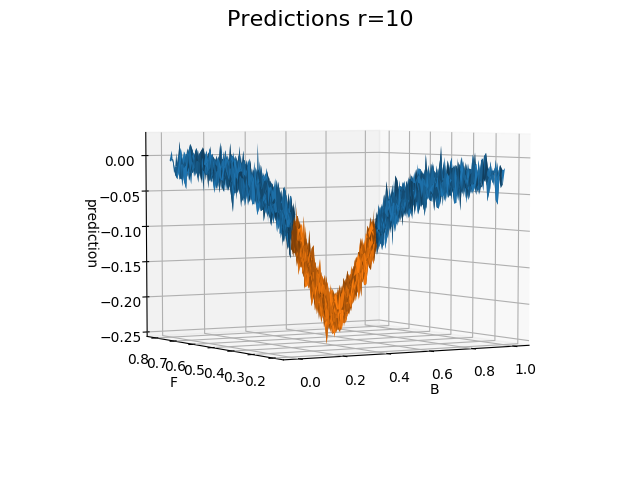}}
    \hfill
    \subfloat{
         \includegraphics[width=0.23\textwidth]{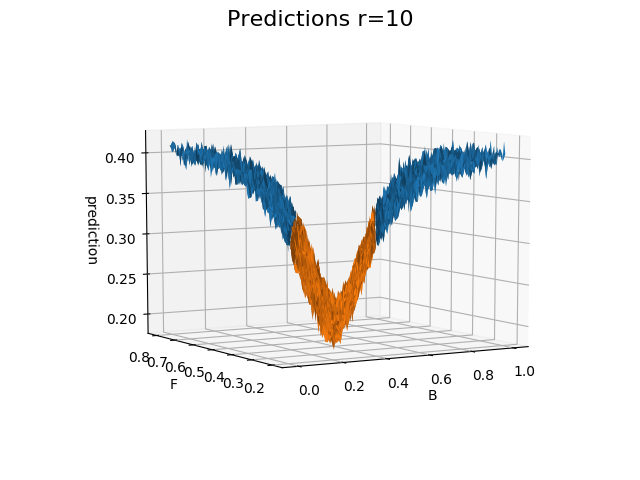}}
    \hfill\\
\caption[]{Unbiased network estimations with fixed intrinsic intensities $(f,b) = (0.6,0.2)$ (top) or with fixed radius $r=10$ (middle and bottom). In the bottom figures, we colour in blue the estimation manifold for the domain of intensities covered by the training data, i.e. $|f-b|\ge\delta$, and in orange the domain not covered by the training data, i.e. $|f-b|< \delta$. If the networks correctly estimate the radius then the top curves should be linear and the middle and bottom ones should be flat constant surfaces in the blue areas. However, this is not a sufficient condition. Focusing on the top row, it is clear that for given intensities the radius estimation is approximately linear. However, the slope depends on them, thus when superimposing all estimations on the entire dataset as in Figure \ref{fig: appendix circle estimations and errors}, we get poor global behaviour. The networks use $\relu$ activations with $C=1$, $4$, or $16$ convolutions kernels in parallel from left to right and trained on both polarity data. }
\label{fig: appendix circle cnn res fixed fb or r all}
\end{figure}

\begin{figure}[!htbp]
    \centering
    \subfloat{
         \includegraphics[width=0.23\textwidth]{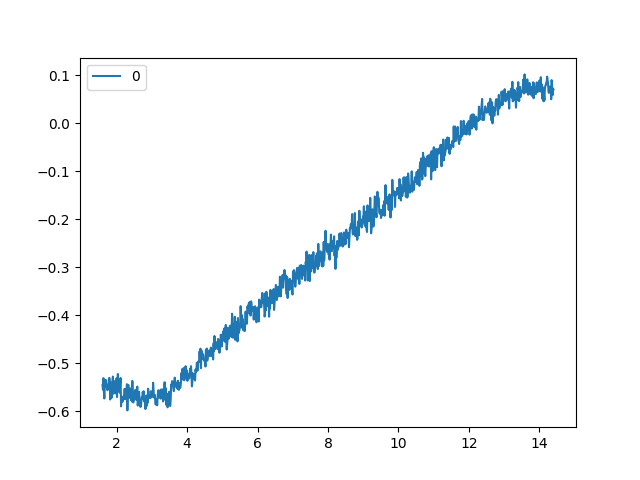}}
    \hfill
    \subfloat{
         \includegraphics[width=0.23\textwidth]{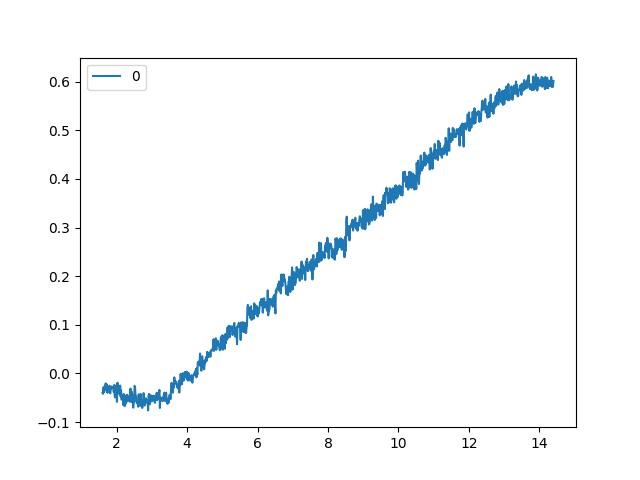}}
    \hfill
    \subfloat{
         \includegraphics[width=0.23\textwidth]{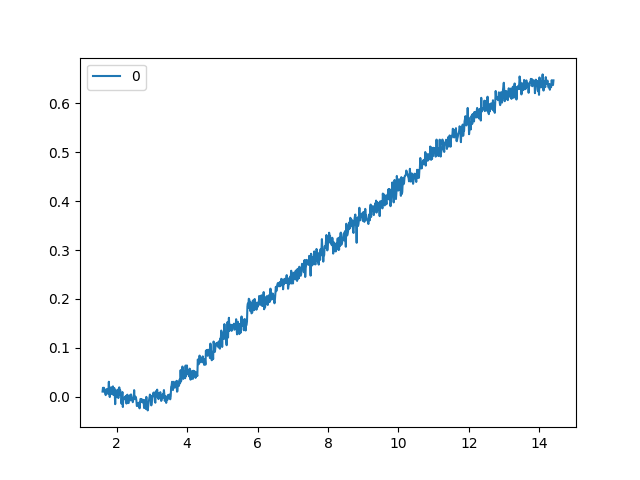}}
    \hfill\\
    \subfloat{
         \includegraphics[width=0.23\textwidth]{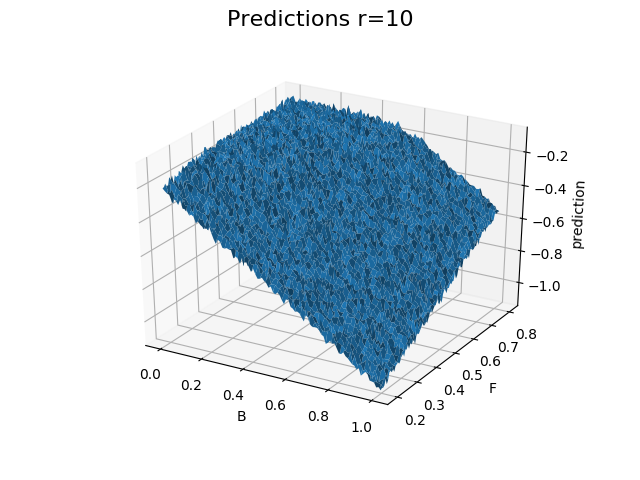}}
    \hfill
    \subfloat{
         \includegraphics[width=0.23\textwidth]{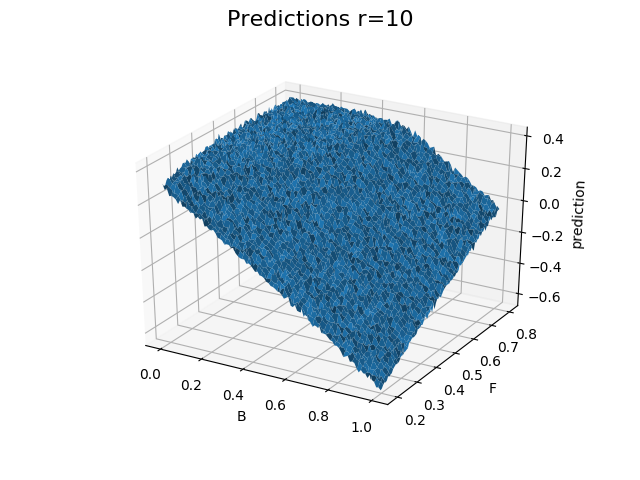}}
    \hfill
    \subfloat{
         \includegraphics[width=0.23\textwidth]{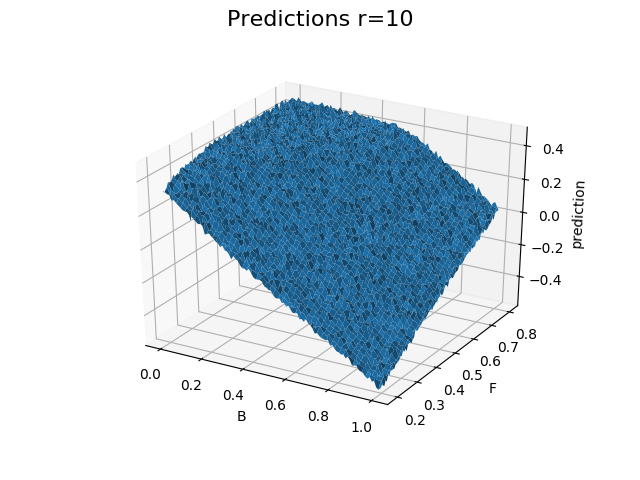}}
    \hfill\\
    \subfloat{
         \includegraphics[width=0.23\textwidth]{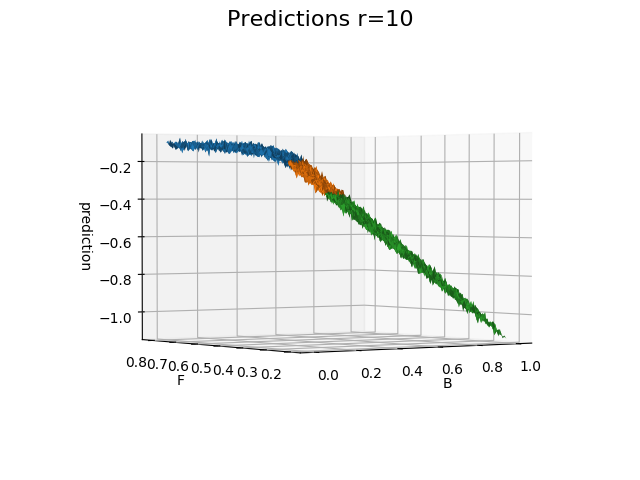}}
    \hfill
    \subfloat{
         \includegraphics[width=0.23\textwidth]{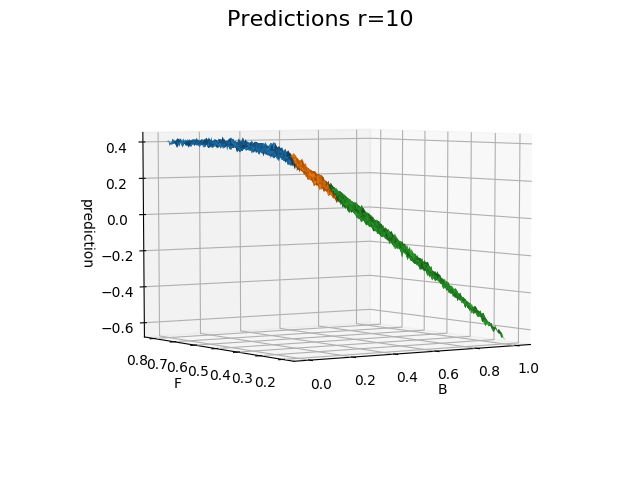}}
    \hfill
    \subfloat{
         \includegraphics[width=0.23\textwidth]{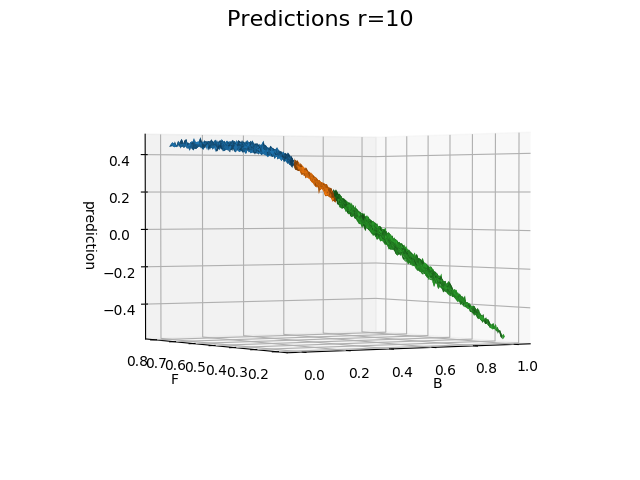}}
    \hfill\\
\caption[]{Same plots as in Figure \ref{fig: appendix circle cnn res fixed fb or r all} but on networks trained with positive polarity data $f>b+\delta$. The orange surface corresponds to estimations on positive polarity data with intensity difference smaller than the margin $\delta$, i.e. with $0<f<b+\delta$, whereas the green surface corresponds to estimations on data with negative polarity $f<b$.}
\label{fig: appendix circle cnn res fixed fb or r all isFbright True}
\end{figure}

\begin{figure}[!htbp]
    \centering
    \subfloat{
         \includegraphics[width=0.4\textwidth]{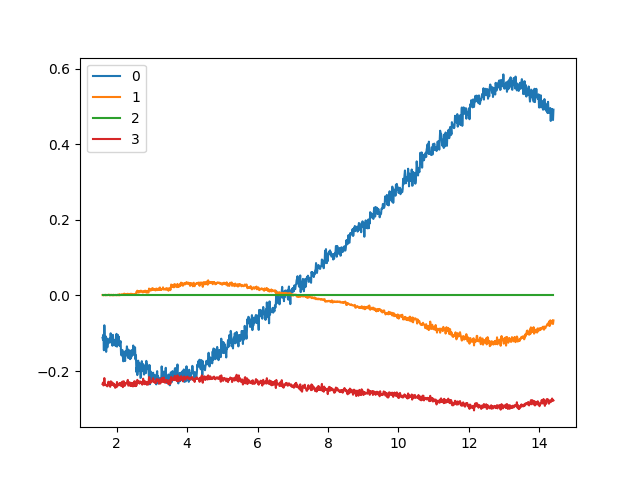}}
    \hfill
    \subfloat{
         \includegraphics[width=0.4\textwidth]{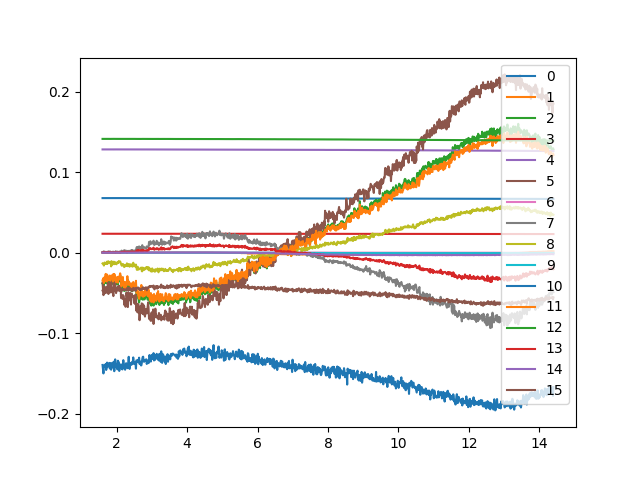}}
    \hfill
    \subfloat{
         \includegraphics[width=0.4\textwidth]{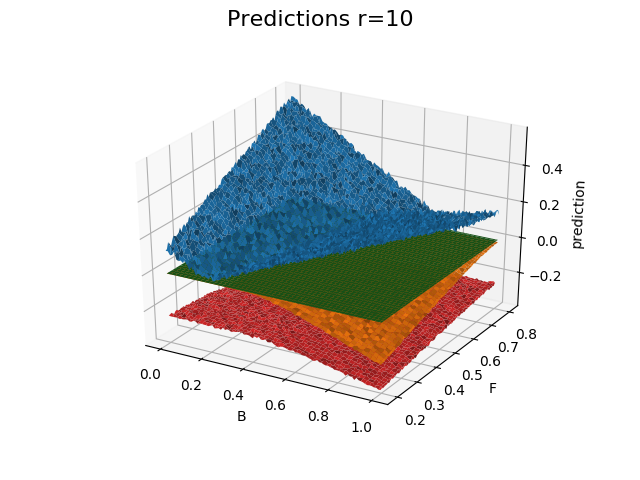}}
    \hfill
    \subfloat{
         \includegraphics[width=0.4\textwidth]{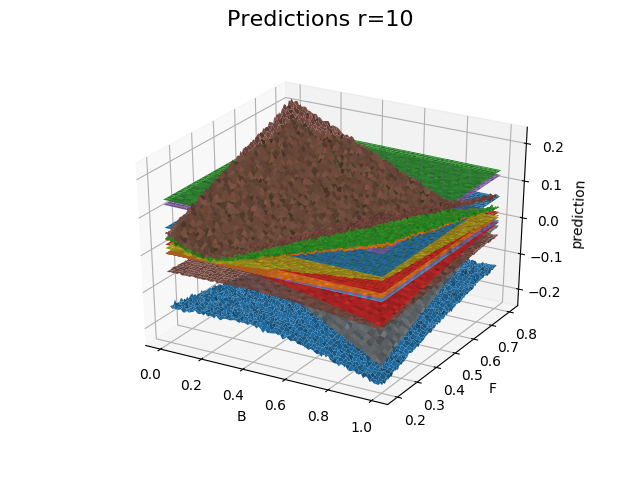}}
    \hfill\\
\caption[]{Unbiased network estimations with fixed intrinsic intensities $(f,b) = (0.6,0.2)$ (top) or with fixed radius $r=10$ (bottom) for a network trained on a dataset with both polarities present. In each figure, we plot the contribution to the final unbiased estimation of each convolution channel. The final estimation given in Figure \ref{fig: appendix circle cnn res fixed fb or r all} is the sum of each curve (top) or surface (bottom).}
\label{fig: circle cnn res fixed fb or r all per conv channel}
\end{figure}

\begin{figure}[!htbp]
    \centering
    \subfloat{
         \includegraphics[width=0.4\textwidth]{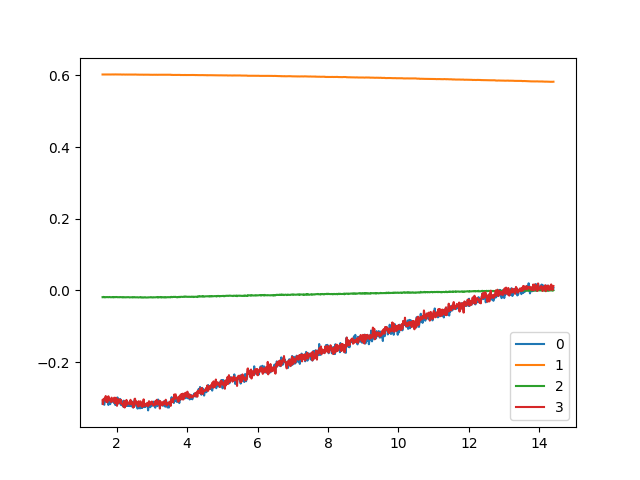}}
    \hfill
    \subfloat{
         \includegraphics[width=0.4\textwidth]{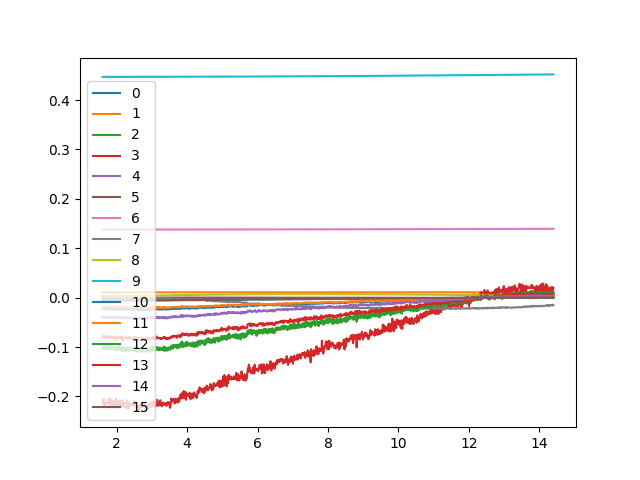}}
    \hfill
    \subfloat{
         \includegraphics[width=0.4\textwidth]{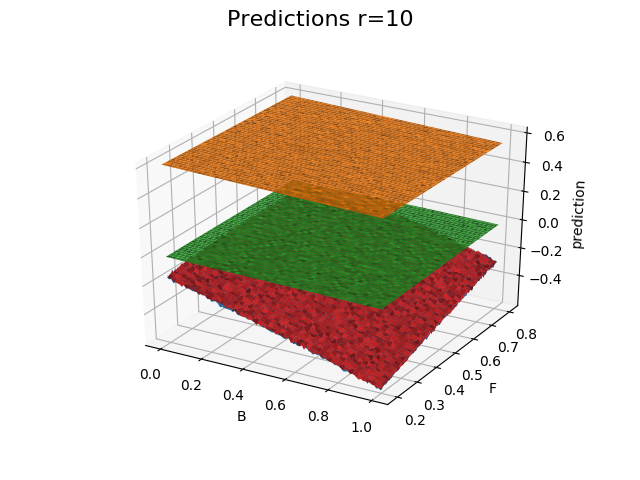}}
    \hfill
    \subfloat{
         \includegraphics[width=0.4\textwidth]{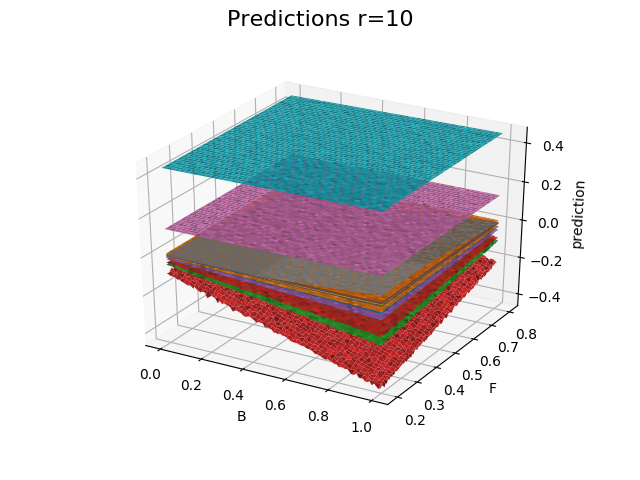}}
    \hfill\\
\caption[]{Unbiased network estimations with fixed intrinsic intensities $(f,b) = (0.6,0.2)$ (top) or with fixed radius $r=10$ (bottom) for a network trained on a dataset with positive polarity only. In each figure, we plot the contribution to the final unbiased estimation of each convolution channel. The final estimation given in Figure \ref{fig: appendix circle cnn res fixed fb or r all isFbright True} is the sum of each curve (top) or surface (bottom).}
\label{fig: circle cnn res fixed fb or r all per conv channel isFbright True}
\end{figure}

\paragraph{Viewing the estimation manifold} Nevertheless, we can witness complex compensation mechanisms encouraging intensity invariance with the increase of the number of channels. In Figures \ref{fig: appendix circle cnn res fixed fb or r all} and \ref{fig: appendix circle cnn res fixed fb or r all isFbright True}, we plot the evolution of the estimations when either the radius or the intensities are fixed. Clearly, increasing the number of channels helps to flatten the estimations, but it does not seem sufficient. In Figures \ref{fig: circle cnn res fixed fb or r all per conv channel} and \ref{fig: circle cnn res fixed fb or r all per conv channel isFbright True}, we plot the same curves and surfaces but for each channel, essentially analysing the evolution in the input space of the contribution of each channel to the final estimation. While it is clear that some channels look uninformative, not all the rest has converged to a unique sloped of flat surface, with some channels having slopes contrary to intuition. These results illustrate the complexity of the multi-channel approach.

\end{document}